\documentclass[lettersize,journal]{IEEEtran}
\usepackage{amsmath,amsfonts}
\usepackage{algorithmic}
\usepackage{array}
\usepackage[caption=false,font=normalsize,labelfont=sf,textfont=sf]{subfig}
\usepackage{textcomp}
\usepackage{stfloats}
\usepackage{url}
\usepackage{verbatim}
\usepackage{graphicx}
\usepackage[numbers]{natbib}
\def\BibTeX{{\rm B\kern-.05em{\sc i\kern-.025em b}\kern-.08em
    T\kern-.1667em\lower.7ex\hbox{E}\kern-.125emX}}
\usepackage{balance}
\usepackage{dsfont}
\usepackage{multirow}
\usepackage{multicol}
\usepackage{graphicx}
\usepackage{booktabs}
\usepackage{mathtools}
\usepackage{ragged2e}
\usepackage{tabularx}
\newcolumntype{L}{>{\RaggedRight}X}
\newcolumntype{R}{>{\RaggedLeft}X}

\begin{document}

\title{Causal Effect Estimation with Global Probabilistic Forecasting: A Case Study of the Impact of Covid-19 Lockdowns on Energy Demand}

\author{
\IEEEauthorblockN{
Ankitha~Nandipura~Prasanna\IEEEauthorrefmark{1},
Priscila~Grecov\IEEEauthorrefmark{1},
Angela~Dieyu~Weng\IEEEauthorrefmark{2},
and Christoph Bergmeir\IEEEauthorrefmark{1}
}

\vspace{3mm}

\IEEEauthorblockA{\IEEEauthorrefmark{1}Department of Data Science and Artificial Intelligence, Monash University, Melbourne, Australia.}

\IEEEauthorblockA{\IEEEauthorrefmark{2}Lauriston Girls' School, Melbourne, Australia
}

}

\vspace{3mm}


\maketitle

\begin{abstract}
The electricity industry is heavily implementing intelligent control systems to improve reliability, availability, security, and efficiency. This implementation needs technological advancements, the development of standards and regulations, as well as testing and planning. Load forecasting and management are critical for reducing demand volatility and improving the market mechanism that connects generators, distributors, and retailers. During policy implementations or external interventions, it is necessary to analyse the uncertainty of their impact on the electricity demand to enable a more accurate response of the system to fluctuating demand. This paper analyses the uncertainties of external intervention impacts on electricity demand. It implements a framework that combines probabilistic and global forecasting models using a deep learning approach to estimate the causal impact distribution of an intervention. The causal effect is assessed by predicting the counterfactual distribution outcome for the affected instances and then contrasting it to the real outcomes. We consider the impact of Covid-19 lockdowns on energy usage as a case study to evaluate the non-uniform effect of this intervention on the electricity demand distribution. We could show that during the initial lockdowns in Australia and some European countries, there was often a more significant decrease in the troughs than in the peaks, while the mean remained almost unaffected.
\end{abstract}


\section{Introduction}

The rise in global energy scarcity, energy efficiency and grid balancing issues have prompted many countries around the world to use intelligent energy management systems. In these systems, typically energy flow is bidirectional and distributed across multiple energy resources. A major challenge in such systems is to balance energy generation and consumption, due to the uncertainty caused by seasonality, natural disasters, technological advances, weather, and temperature \cite{arize2019comparison}, in both the generation and the consumption side.  On the supply side, energy generated by renewable energy resources such as rooftop photovoltaics, wind power, and solar power can exacerbate network issues such as voltage control, loss and congestion in the network. On the demand side, flexible demand for electric vehicles, energy storage and commercial and domestic consumption needs to be met regularly by the electric power provision, which demands communication and planning between the two sides of the electric system, i.e.\@ generation and consumption. In particular, the supply-demand balance must be maintained to avoid system instability, power interruptions, and frequency deviations. Thus, an intelligent energy control system must quantify the potential uncertainties linked to future energy generation and user consumption \cite{mazhari2018quantile,wen2019performance}.

Forecasting methods, in particular probabilistic forecasting algorithms are commonly the tools of choice in this space. Research in this space has been driven by the GEFcom competition series (2012, 2014, and 2017) \cite{hong2019global,hong2016probabilistic}, and there is now a rich literature in forecasting the energy supply and demand; especially also probabilistic approaches of forecasting the energy supply and demand have been proposed \cite{salinas2020deepar,9316921,NAGY20161087}.

There are many external factors that influence both energy supply and demand, such as weather characteristics (wind speed and direction, solar irradiation, temperature), results of climate change such as natural disasters, or external shocks such as the Covid-19 pandemic.
While forecasting methods can try to incorporate such external factors to achieve higher forecasting accuracy, policy makers and market participants often need to assess the impact of such external shocks to the energy market, need to consider different scenarios, and need to make decisions and assess the impact of these decisions.

This is the related domain of causal effect analysis methods.
Causal effect analysis is the process of determining the effect of interventions, from which it enables to directly determine how, e.g., the changes in policy create changes in real world outcomes. This is accomplished by forecasting the counterfactual outcomes for affected instances and comparing them with the real outcomes.
As effects of these interventions in the real world can be complex, it may often not be enough to look at their average effects.
Instead, we want to assess how interventions or policy implementations affect the distribution of energy profiles in quantile levels. For example, an intervention may lead to higher minimal or maximal energy production/consumption, without altering the mean production/consumption.

To the best of our knowledge, very little research has been done in this area, so that our work is the first to take advantage of the recent and emerging literature on the use of machine learning (ML) tools and probabilistic global forecasting methods (GFM), to construct counterfactual energy consumption paths and distributions after interventions. More recently, probabilistic ML and GFM approaches have provided important new techniques to refine causal effect estimation when dealing with observational data in higher dimensional settings. Therefore, its application for causal effect analysis in the energy market enables flexible control of the usual high-frequency time series data in electricity markets. Hourly electricity demand data permits us to embody a large set of controls and fixed effects to detach the causal impact of a specific intervention from other confounders. Applying this ML-GFM modelling on pre-treatment data to forecast counterfactual energy demand paths and distributions under no intervention allows us to create moldable and data-driven energy use models with less risk of overfitting that captures non-uniform effects over the demand distribution. Consequently, we can achieve more reliable treatment effect estimations. As far as we are aware, this paper is the first to empirically implement a neural network (NN) and probabilistic GFM framework for causal effect analysis in electricity markets.

An obvious case study to showcase our methodology is the Covid-19 pandemic and its associated lockdowns, which caused changes in lifestyle, closures of commercial buildings, schools, and whole business sectors.
The International Energy Agency (IEA) indicated that the change in energy demand provoked by the pandemic is the biggest in the last 70 years \cite{IEA}.
The change in lifestyle and work from home caused changes in daily demand distributions, e.g., in the form of a reduced shift in distribution, change in peak and troughs of the distribution. Those are crucial information for understanding the uncertainties related to generation and transmission systems~\cite{mazhari2018quantile,wan2016direct}.

The global rate of reduction in energy demand due to the pandemic ranges from 2\% (mild impact) to 26\% (extreme impact) \cite{buechler2020power}. Due to surplus energy generation, the reduction in energy demand causes voltage increase issues in some distribution systems, particularly in areas with many factories and commercial buildings that were fully or partially closed during Covid-19 lockdowns. Uncertainties in demand and generation further threaten the power balance.

In particular, we choose Australia and Europe as the regions to study the lockdown impact on energy consumption in our case study. This is because, firstly, they imposed restrictive lockdown measures at the very beginning of the pandemic at similar starting dates. Secondly, there are distinctions in how they established lockdown restrictions across the countries/states, with some regions not imposing them, which is essential for a causal impact analysis of policy interventions.

In detail, each state government in Australia imposed Covid-19 restrictions in their respective timelines, causing a nationwide lockdown to begin on March 23, 2020; by this date, all non-essential businesses and schools in Australia were closed.  Whereas in Europe, Italy was the first country that was hit severely and quickly by Covid-19. Spain and Poland went into lockdown on March 14, 2020, and Switzerland on March 16, 2020 (see Figure 1 in Section D of Supplementary Material for more details). The pandemic resulted in the closure of all schools, commercial activities, and businesses, decreasing total country energy consumption. Because of the global decline in electricity demand, some electricity companies have been forced to reduce working hours in both the mining and generation segments. 

We calculate the effect of the lockdowns as an intervention on each state in the Australian national energy market, and the effect of an intervention on countries across Europe per quantile to determine which quantile has the highest concentration of intervention impacts. We find that overall there is (unsurprisingly) a reduced energy consumption in Australian states and European countries. However, while some states or countries experienced a uniform consumption reduction, others experienced a non-uniform reduction or even some increases for specific parts of the energy demand distribution.

The rest of this paper is structured as follows. Section II discusses previous work, and Section III presents the methodology of our proposed approach. The details of the data context and an experiment using a concrete dataset are described in Section IV, and their results, along with the main findings, are discussed in Section V. Finally, Section VI offers conclusions.

\section{Related Work}

\noindent In this section, we review pertinent background literature from the fields of probabilistic forecasting and causal inference, along with the relevant studies that examine the impact of Covid-19 on energy consumption.

\subsection{Global probabilistic forecasting and causal inference}

Early works that train models across series are \citet{BANDARA2020112896,salinas2020deepar,wen2017multi}. However, such models have gained wider attention after such a cross-series model won the M4 forecasting competition~\cite{smyl2020hybrid}, and have recently in the M5 competition~\cite{makridakis2022m5} again shown that they are in many situations superior to models that are trained on each series separately.
\citet{Januschowski2020} introduced the terms and concepts of global models for forecasting. Those authors defined local modelling as the process when the parameters are estimated independently for each time series, whereas global modelling jointly learns across a full set of time series. Both global and local models can be univariate or multivariate, where we focus on univariate versions that operate on one series at a time. Since GFMs benefit from training across multiple time series fitting a single model, they can often learn better the key patterns across a set of time series. Consequently, they often deliver more accurate forecasts \cite{hewamalage2022global, MONTEROMANSO20211632}.

Probabilistic forecasting is the prediction of future values in the form of probabilistic forecast distributions which is based on a large set of historical data. It provides insights into the forecast's uncertainty and aims to maximise the forecast distribution's acuity. 
The DeepAR model proposed by \citet{salinas2020deepar} is arguably the most prominent model that enables global probabilistic forecasting.
It performs probabilistic forecasting by predicting the parameters of a parametric forecasting distribution, such as negative binomial or Gaussian distributions. It is based on learning an autoregressive Recurrent Neural Network (RNN) model on a vast set of related time series, with special treatments when magnitudes in the series vary. RNNs process the data with temporal dependencies and carry that information (model parameters) across multiple time steps. The unfolded RNN can observe the model parameters from each time step and provide them to successive time steps. 
This model learns the temporal dependencies, seasonal patterns and trends to forecast the full probability distribution of the forecast.

On the other hand, causal inference refers to concluding whether a specific treatment or intervention was the ``cause" or not of an observed effect. The main goal of the causal inference literature is to find the impact of an intervention which is vital during policy making. Causal effect estimation is the difference between the counterfactual outcome (prediction in the absence of intervention) and actual post-intervention observation. 
The transition from local to global models (considering all the available time series) led to considerable advances in the forecasting domain that yet have to be fully reflected in the causal inference literature. The only works we are aware of in this space are the works of \citet{grecov2021causal,9833461}. \citet{grecov2021causal}  introduced the DeepCPNet, using a globally trained RNN for counterfactual prediction, overcoming the limitations of structural causal models. 
Our previous work in \citet{9833461} introduced a non-parametric probabilistic version of the former method, which we extend in this paper in various directions, as follows. 

We observe that the benefit of aforementioned methods of probabilistic forecasting, global forecasting and synthetic control methods has not been unlocked yet for causal inference in the electricity domain. Motivated by this gap, we propose a practical framework that combines global and probabilistic forecasting methods by employing DeepAR to generate counterfactual predictions to estimate the impact caused by Covid-19 on energy consumption.
The new framework is more accessible by employing a simpler parametric approach to probabilistic forecasting offered by built-in NN models available in Python packages highly disseminated among GFM users. Therewith, the methods are faster and easier to apply, like DeepAR. We furthermore introduce a different type of sampling, i.e., a sequential sampling, and we show that the proposed methodology works well even in scenarios where we have few time series and only one control unit (as in our study case), because of the use of GFMs. Under these conditions, structural causal models, for example, do not function properly.

\subsection{ Covid-19 impact analysis on energy demand using machine learning models }


Many works in the literature to analysing the impact of Covid-19 on energy demand are completed through exploratory data analysis and the use of several analytical tools to present the energy usage, peak demands, and demand profiles visually, and compare the various countries around the world with their energy consumption. There are various machine learning and statistical models that are implemented in the literature to evaluate the impact of Covid-19 on electricity markets, such as, linear regression \cite{buechler2020power}, probabilistic forecasting models \cite{van2018probabilistic}, clustering \cite{buechler2020power} and counterfactual prediction \cite{ abrell2022effective, graf2021machine} to analyse the impact. The unified modelling framework proposed by \Citet{buechler2020power} explores the relationship between change in electricity demand, mobility and government restrictions. This method estimates the region-specific regression models to predict electricity consumption without the pandemic accounting for weather, seasonal and temporal effects and further K-Means clustering is applied to investigate the similarity in the response between different countries. 

In addition, the K-Means algorithm has been implemented by \citet{garcia2021retrospective} to perform automatic customer clustering to split the residential and non-residential customers based on the behaviour during the pandemic. \citet{burlig2020machine} generated the counterfactual outcome with the Lasso regression model for analysing the energy efficiency, then comparing the predicted outcome to the realised outcome to calculate the treatment effect. \citet{graf2021machine} proposed a counterfactual neural network model to predict business-as-usual re-dispatch costs in the electricity market and compare them to actual re-dispatch costs for the pre-covid and Covid-19 lockdown periods. \citet{abrell2022effective} estimated the impact intervention using the prediction-based estimator. This treatment effect estimator is based on counterfactual results and considers the influence of observed and unobserved data.

In comparison, probabilistic forecasting is used in only a few papers in the electricity literature to assess the impact of the pandemic. \citet{van2018probabilistic} used Gaussian processes (GPs) as a non-parametric model to generate probabilistic forecasts of electricity consumption, photovoltaic (PV) power generation and net demand. This framework includes both a direct and an indirect method of evaluating net demand using static and dynamic GPs, with the dynamic GPs producing sharper prediction intervals while requiring less computational effort.

Finally, the previously described cutting-edge methods for analysing the impact of Covid-19 on energy sectors in the electricity literature are primarily focused on visualising and analysing the change in demand using simple statistical tools. Furthermore, the use of ML methods is limited to predicting post-pandemic data, which is done by using local models and comparing them to actual data. Even when GFMs are used, the average energy consumption is examined \cite{van2018probabilistic}, but not the distribution's quantile levels. The state-of-the-art methods in the power system literature like, Quasi-Monte Carlo \cite{xie2017quasi,krishna2022uniform}, and quantile regression \cite{haque2014hybrid,wan2016direct,mazhari2018quantile,wen2019performance} were implemented for efficient probabilistic power flow in distributed energy resources and probabilistic prediction for generators in power systems. But the quantile level analysis of the energy demand distribution under external shocks is critical for understanding the distribution's peaks and troughs and must be accurate to provide an efficient decision support in transmission systems. The increase or decrease in demand can be observed by examining the behaviour of the impact using the various forecasted quantiles. To fill this gap, this study employs a probabilistic forecasting approach to measure the pandemic impact by each quantile of the electricity demand.

\section{Methodology}

In the following, we first define the basic setup of counterfactual and causal effect estimation problems for counterfactual and synthetic control methods. Then, we present the NN-GFM framework modelling employed to produce the counterfactual point predictions. We conclude this section by showing how the quantile probabilistic approach is integrated into the prior method for forecasting the full counterfactual distribution.

\subsection{The Basic Setup of the Counterfactual and Causal Inference Problem}

We assume a set of $n$ time series instances indexed by $j=1,\ldots,n$, which might be, e.g., different products, regions, or stores, coming from the same data generation process. For each instance and time period $t=1,\ldots,T$, we notice the occurrence of a variable $Y_{jt}$. Furthermore, we assume an intervention (treatment) took place at $T_{0}+1$, where $1<T_0<T$, over only a subset $i$ of \emph{treated} units, where $i < n$. The remaining $c=n-i$ units, which we call \emph{control} units, are never affected by this intervention for all $t$.

In addition, consider $\mathcal{D}_t \in (0,1)$ to be a binary variable that indicates when the intervention happens (binary treatment). That means $\mathcal{D}_t = 1$ when $t > T_0$, and $\mathcal{D}_t = 0$ otherwise. Therefore, following the potential outcome notation, we can express $Y_{i,t}$ such that:
\begin{equation}
\label{eq:1}
    Y_{i,t} = \mathcal{D}_{t} Y_{i,t}^{(1)} + (1-\mathcal{D}_{t}) Y_{i,t}^{(0)},
\end{equation}

\noindent where $Y_{i,t}^{(1)}$ is the potential outcome for the treated unit $i$ when being affected by the intervention, whilst $Y_{i,t}^{(0)}$ is the potential outcome when $i$ is not under the intervention effect (the counterfactual outcome). Moreover, under the synthetic control method (SCM) approach, we determine $Y_{c,t}$ as the control units and assume they are unaffected by the treatment $\mathcal{D}_t$ at any time, which means their observed outcomes are the same as their outcomes with no intervention, that is, \(Y_{c,t}^{(1)}=Y_{c,t}^{(0)}\) for all \(1\leq t \leq T\).

As a result, the treatment effect is assessed by employing the forecast model $\mathcal{M}$ to derive mean-unbiased proxies for the counterfactual of the $i$ treated units, $P_{i,t}^{(0)}:=\mathcal{M}(Y_{j,t \le T_{0}}^{(0)};\theta)$, where $\mathcal{M}:\mathcal{Z} \times \Theta \rightarrow \mathcal{Y}$ is a function which considers all the true values of the pre-intervention period $Y_{j,t \le T_{0}}^{(0)} \in \mathcal{Z} \subseteq \mathbb{R}^d \; (d>0)$ as covariates, and model parameters $\theta \in \Theta$ of a finite-dimensional parametric space as follows: 
\begin{equation}
\label{eq:2}
    Y_{i,t>T_{0}}^{(0)} = P_{i,t>T_0}^{(0)} + u_{t} , \;\;
\end{equation}

\noindent where  $u_t$ is an independent and identically distributed stationary stochastic process (with  $E(u_{t})=0$). This leads us to identify the treatment effect $\delta$, by assuming the stationarity on $Y_{j,t \le T_{0}}^{(0)}$ and knowledge of the exact start timing of a well-defined intervention, as $\widehat{\delta}_t := Y_{i,t>T_{0}}^{(1)} - Y_{i,t>T_{0}}^{(0)}$. Consequently, the estimation of the \emph{average treatment effect} (ATE) on the treated units over the post-treatment period is delineated as

\begin{equation}
\label{eq:4}
    \widehat{\Delta}_T=\frac{1}{T-T_0}\sum_{t = T_0 +1}^{T}\widehat{\delta}_{t} \;.
\end{equation}  

Overall, SCM necessitates the model $\mathcal{M}(\cdot,\cdot)$ selection that maps the information from  pre-intervention data of both the treated and control units, implying that the counterfactual estimation is performed using a model which is trained in the absence of any treatment.
With the exception of the strong assumptions requested by the structural causal models, we propose a flexible model using RNNs for autoregressive time series forecasting, and employing the GFM in conjunction with the quantile probabilistic forecasting approach. Instead of estimating counterfactual point forecasting, we estimate the full counterfactual distribution, as explained in detail in the following section.

\subsection{Global Probabilistic Forecast Framework}
\subsubsection{Neural Network Global Forecasting of Counterfactual Point Estimation}
Following the potential outcome framework \cite{RN50}, we aim to infer counterfactual outcomes $Y_{i,t}^{(0)}$ based on pre-treatment outcomes and covariates to estimate $\widehat{\Delta}_t$ as defined in (\ref{eq:4}). 
The framework we propose for that follows the idea introduced by \cite{grecov2021causal,9833461}, where the non-linear mapping $\mathcal{M}$ is a GFM-RNN model to estimate counterfactual mean unbiased proxies, \(P_{i,t}^{(0)}\). The idea is to jointly fit a forecasting method that employs autoregressive RNNs and, consequently, learns a \emph{global} model from the past data of the entire pool of time series, therefore improving the basis for learning. Hence, the \emph{global} modelling approach allows the $P_{i,t}^{(0)}$ to be assessed taking pre-treatment data of both control, and treated series into account , $Y_{i,t \leq T_0}$ and $Y_{c,t \leq T_0}$, as the covariates to learn the model. Hence, Equation (\ref{eq:2}) can be re-written as $Y_{i,t>T_0}^{(0)} = \mathcal{M}(Y_{j,t\leq T_0},\mathbf{\theta}) + u_t$.


Whereas with the post-treatment prediction for the treated units $Y_{i,t}^{(0)}$ we reach the counterfactual outcomes; on the contrary, the forecast quality of the framework is determined by the accuracy of the post-treatment forecast of the control series $Y_{c,t}$ by the assumption of null treatment effect over controls $Y_{c,t}$ combined with \emph{global} modelling with shared parameters $\theta$ throughout all series $j$. The shared parameters $\theta$ learnt in the pre-treatment period are used for forecasting both treated and control units. As a result, when we verify and test the null treatment effect over the control units, we ultimately test how effectively the model parameters perform in the post-treatment period.

The choice of the GFM-RNN modelling in our work is based on the framework proposed in the work of \citet{salinas2020deepar} by using an autoregressive RNN global time series model called DeepAR
. DeepAR is a global sequence-to-sequence (Seq2Seq) neural net and parametric probabilistic forecasting model mainly used for non-linear and non-stationary time series forecasting. In our implementation, DeepAR involves an RNN (using LSTM cells) that considers the previous time points of related time series and covariate time series as inputs, learning a global model from the history of all sets of time series jointly. Thus, given the $J$ treated and control univariate time series $\{y_{j,1:T_0}\}_{j=1}^{J}$, the aim of our neural net is to forecast the future trajectories after the intervention for each time series $\{y_{j,T_{0}+1:T}\}_{j=1}^{J}$ given its historical pre-intervention data (with no intervention), which are the input variables to our global deep neural model as follows:
\begin{equation}
\label{eq:6}
    \{y_{j,T_{0}+1:T}\}_{j=1}^{J} = m_G(y_{:j,1:T_0},\Theta),
\end{equation}

\noindent where the model's parameter set is $\Theta$, $m_G$ is our GFM-DeepAR model, and the value of the time series $j$ at time $t$ is $y_{j,t} \in \mathbb{R}$.

The proposed architecture of autoregressive LSTM follows the architecture implemented by \citet{salinas2020deepar}. The used LSTM is both autoregressive and recurrent; hence it takes as an input for the current model’s time step the values of the previous time step ($y_{:j,t-1}$) and hidden state (${h}_{j,t-1}$) of the network. Therefore, the LSTM of $m_G(\cdot)$ model is expressed as follows:
\begin{equation}
\label{eq:7}
    \mathbf{h}_{j,t} = r(\mathbf{h}_{j,t-1},y_{:j,t-1},\Theta),
\end{equation}

\noindent where $r(\cdot)$ is a multilayer RNN with LSTM cells, and $\Theta$ are its learnable parameters.

The hidden state activation function in Equation (\ref{eq:7}) is a non-linear function such as the rectified linear unit (ReLU) function. We train the LSTM on pre-treatment outcomes using multiple training instances, i.e., sliding windows with different starting points and a fixed size of 1.25 times the prediction length. This moving window procedure allows to augment the data used and to map the ``new'' time series behaviour by considering all other available predictors. Since the proposed framework follows the GFM-RNN approach, it adds more information to our modelling by learning the parameters $\Theta$ globally across all the time series $J$ and sharing them among each value of $t$. This allows consequent gain of accuracy. Another benefit of incorporating the GFM unlike the current structural causal models like SCM, is that it is not required to consider the relationship between control and treated units in the post-treatment period which is the same as the mapping done in the pre-treatment period to estimate the parameters over the post-treatment time series and later the estimation of the effect is determined. Here it is not necessary to consider these strict dependencies between the pre- and post-treatment data since our learning process only considers data before the intervention. Our GFM-DeepAR modelling employs the ADAM optimiser with two-layer LSTM cells with 25 RNN cells for each layer.

In the following section, we explain how we combine the probabilistic forecasting techniques to this GFM-DeepAR model to estimate the distribution of the counterfactual as an outcome.

\subsubsection{Probabilistic Global Forecasting of Counterfactual Distribution}
\noindent DeepAR is a methodology that performs accurate probabilistic forecasting for a set of related time series by employing autoregressive models such as RNNs. It is trained using maximum likelihood estimation to produce distribution parameters and probabilistic forecasts via sequential sampling. When we suggest applying this methodology to the counterfactual forecasting for causal effect estimation, we are able to generate the counterfactual distribution for the treated units in the absence of the intervention effects and, therefore, estimate the distribution of the treatment effect across the period of intervention. Such probabilistic estimations are crucial in many causal effect studies. They allow optimal decision-making under interventions that affect the distribution non-uniformly or when facing skewed distributions, something not achievable by performing only point forecasts.

The default distribution used by DeepAR to evaluate the historical observations and generate the sample predictions is the Student’s t distribution, which is the one considered for the experiments in this paper. In our application, this probabilistic part of DeepAR has as its goal to model the conditional distribution of the post-intervention time series, i.e., $P(y_{j,T_{0}+1:T}|y_{:j,1:T_0})$. As the model uses an autoregressive RNN architecture (as expressed by Equation (\ref{eq:7})), the model distribution $L_{\Theta}(y_{j,T_{0}+1:T}|y_{:j,1:T_0})$ consists of a product of likelihood factors such as


\begin{equation}
\label{eq:8}
L_{\Theta}(y_{j,T_{0}+1:T}|y_{j,1:T_0}) = \prod_{t=T_{0}+1}^{T}p(y_{j,t}|\theta(\mathbf{h}_{j,t},\Theta)),
\end{equation}

\noindent parameterised by the output $\mathbf{h}_{j,t}$ of the autoregressive RNN. The likelihood $p(y_{j,t}|\theta(\mathbf{h}_{j,t}))$ follows a Student’s t distribution where the parameters are estimated by the function $\theta(\mathbf{h}_{j,t},\Theta)$ of the LSTM output $\mathbf{h}_{j,t}$ (as defined in (\ref{eq:7})).

From the final probabilistic forecasting of $\{y_{j,T_{0}+1:T}\}_{j=1}^{J}$, we extract $Q$ defined quantiles $\tau$ for all time series $j$. The quantiles obtained from quantile forecasting are accumulated to build the estimated quantile Student's t distribution, as follows

\begin{equation}
\label{eq:9}
    \begin{cases}
      (\hat{y}_{:j,t+1}^{(\tau_1)},...,\hat{y}_{:j,t+K}^{(\tau_1)}) &= m_G(h_t,y_{:j,:t};\Theta)[\tau_1]\\
      \cdots\\\
      (\hat{y}_{:j,t+1}^{(\tau_Q)},...,\hat{y}_{:j,t+K}^{(\tau_Q)}) &= m_G(h_t,y_{:j,:t};\Theta)[\tau_Q]
    \end{cases}\,,
\end{equation}

\noindent where $y_{:j,\cdot}$ are the time series $j$ to be forecasted by the \emph{global} LSTM model $m_G$ which learned the parameters globally and shared across with all the forecasting points $k \in \{1,\cdots,K\}$ and time series $j \in \{1,\cdots,J\}$, along with the encoded historical data's information as hidden state values $h_t$, for each of the $q \in \{1,\cdots,Q\}$ quantiles denoted by $\tau_{(\cdot)}$, and $y_{:j,:t}$ are the temporal covariates obtained from the pre-intervention data (the $Y_{j,t\leq T_0}$).


To build and plot the estimated quantile distribution function of the post-intervention for each time series $j$, we pass the series $\{\hat{y}_{j,t+k}^{(\tau_1)},...,\hat{y}_{j,t+k}^{(\tau_Q)}\}$ to plot the probabilistic forecast quantile distribution with the predefined quantiles within the domain $\tau_{(\cdot)}$. Finally, considering these quantile forecasts, we can now contrast the quantile distribution of the factual post-intervention by passing them through the quantile function \texttt{qt()} of the Student's t distribution in R (this is done to analyse the causal effect of intervention) and comparing against the estimated quantile distribution of the counterfactual, and the causal effect identification now might be expressed for each quantile $\tau$ as $\widehat{\delta}_t^{(\tau)} := \hat{Y}_{i,\tau}^{(1)} - \hat{Y}_{i,\tau}^{(0)}$, and the ATE for a specific quantile $\tau$, 
as

\begin{equation}
\label{eq:12}
    \widehat{\Delta}_T^{(\tau)}=\frac{1}{T-T_0}\sum_{t = T_0 +1}^{T}\widehat{\delta}_{t}^{(\tau)} \;,
\end{equation}


\section{Experiments}

In this section, we first describe the context of the data collection process as the characteristics of the datasets to be analysed. Subsequently, we present an exploratory analysis of the data to select the time series that will be considered as control and treated units across the study. Later, for the comparison of our method we outline the baseline models and the models' performance assessment. Lastly, we explain the placebo and hypothesis tests that we conduct. 

\begin{figure}[!t]
\includegraphics[width=.15\textwidth]{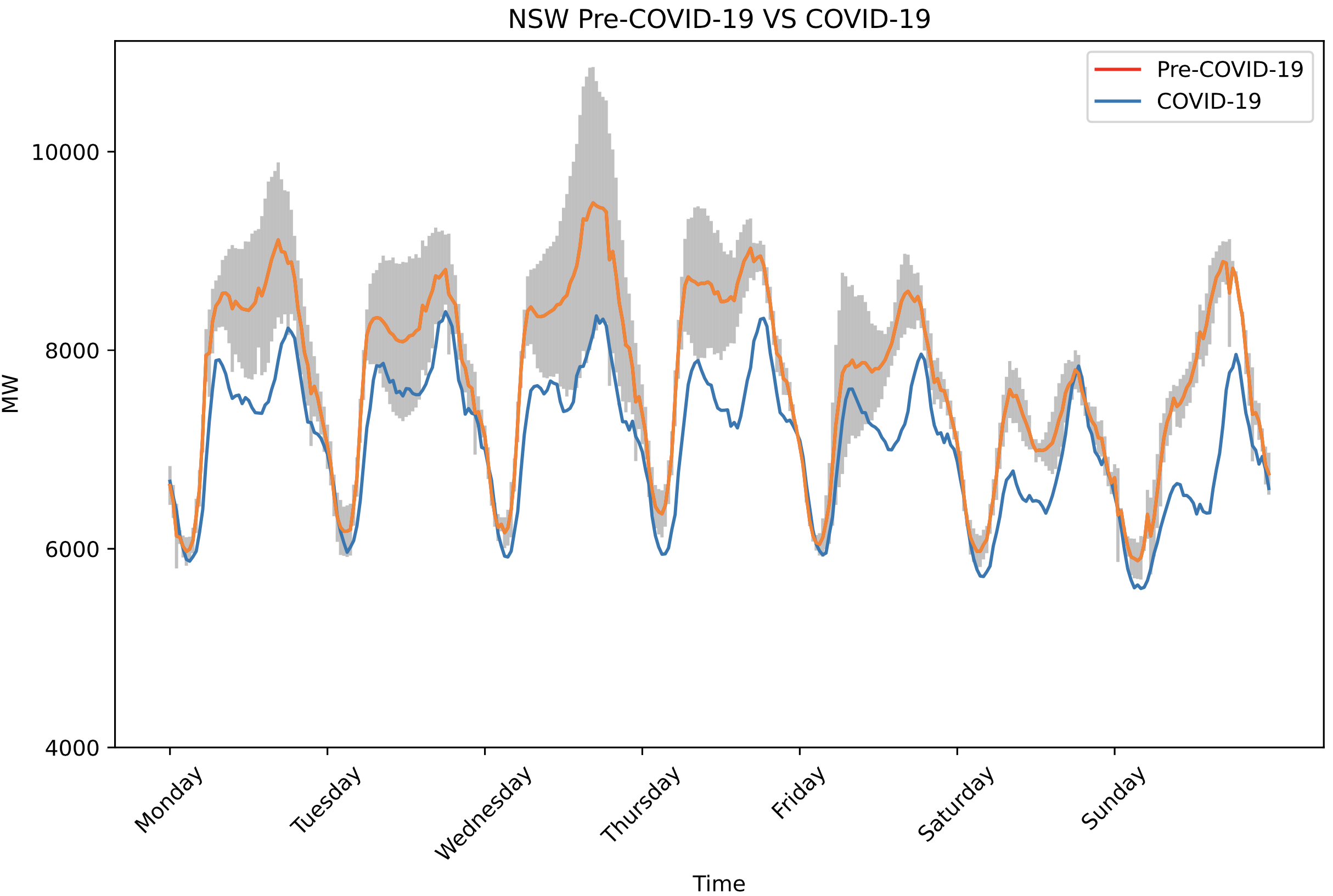}\quad
            \includegraphics[width=.15\textwidth]{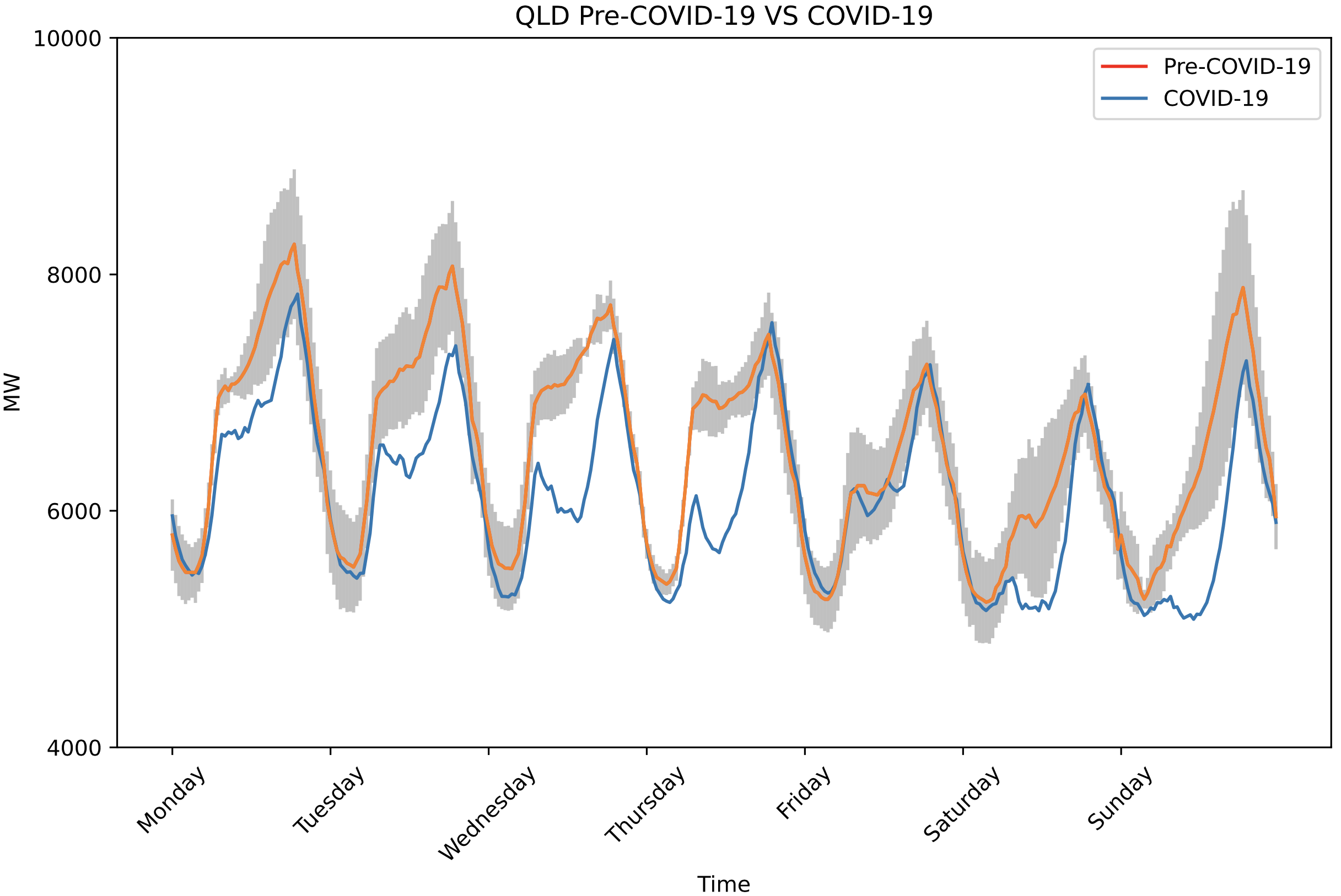}
            \includegraphics[width=.15\textwidth]{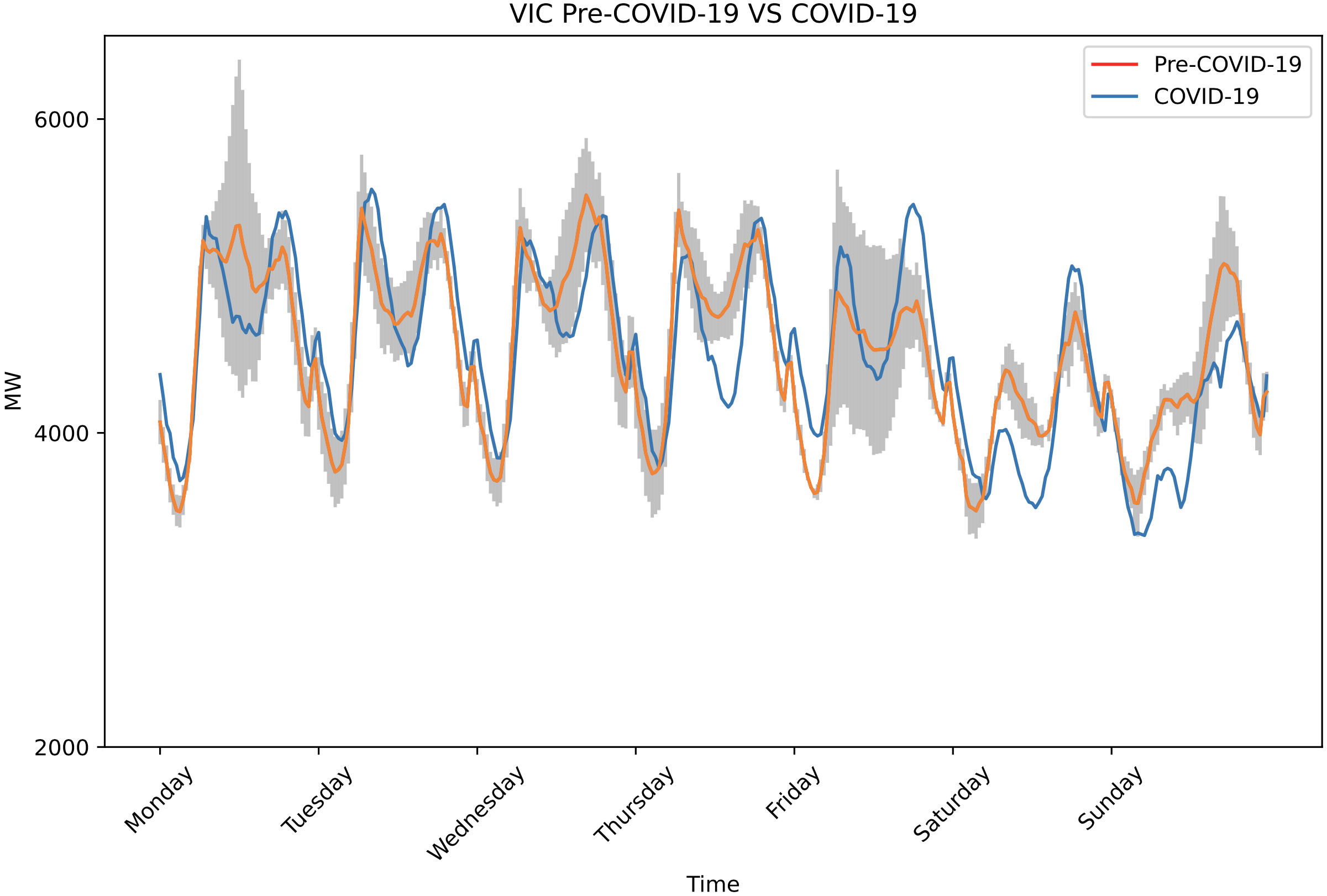}
            \qquad
            \includegraphics[width=.15\textwidth]{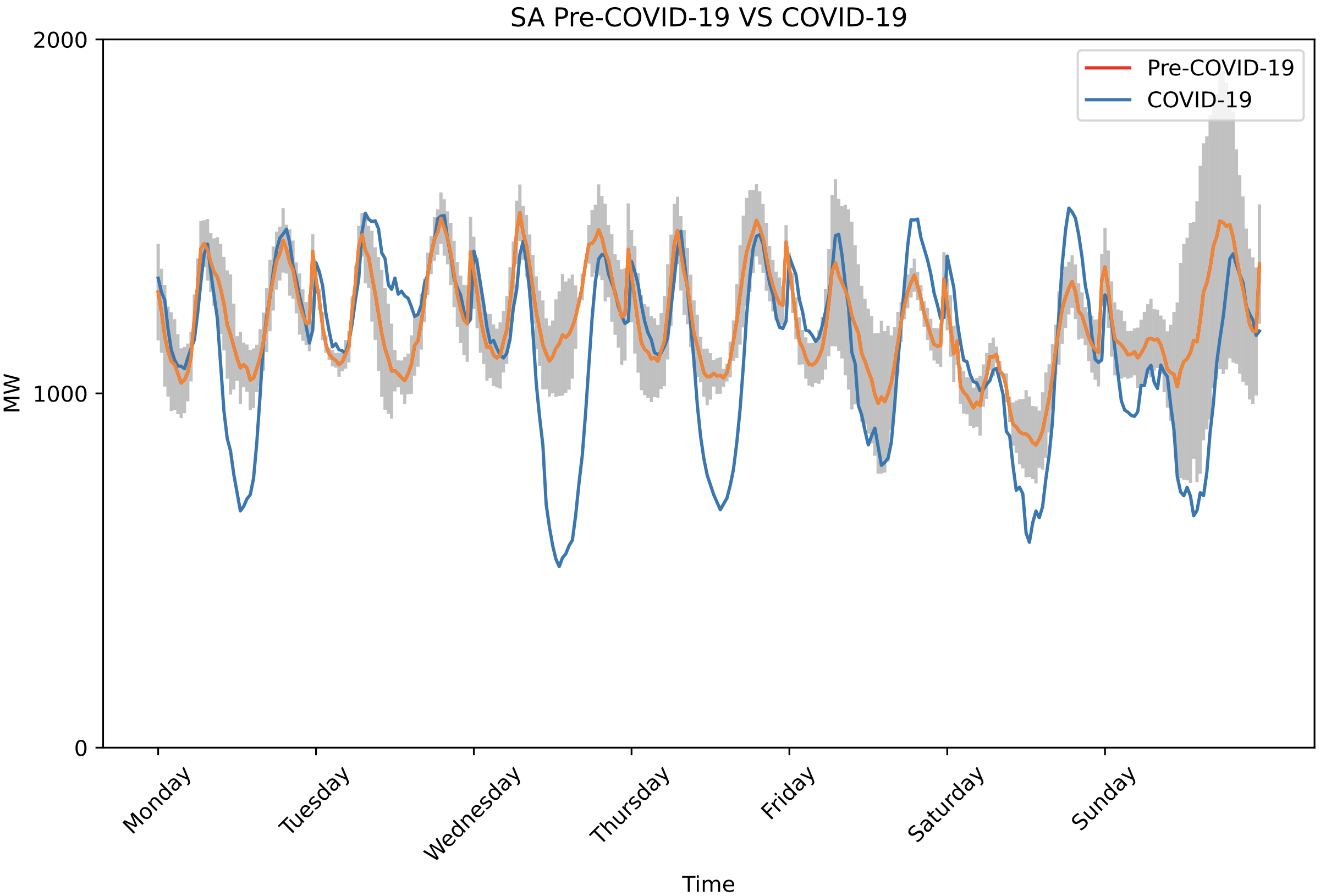}\quad
            \includegraphics[width=.15\textwidth]{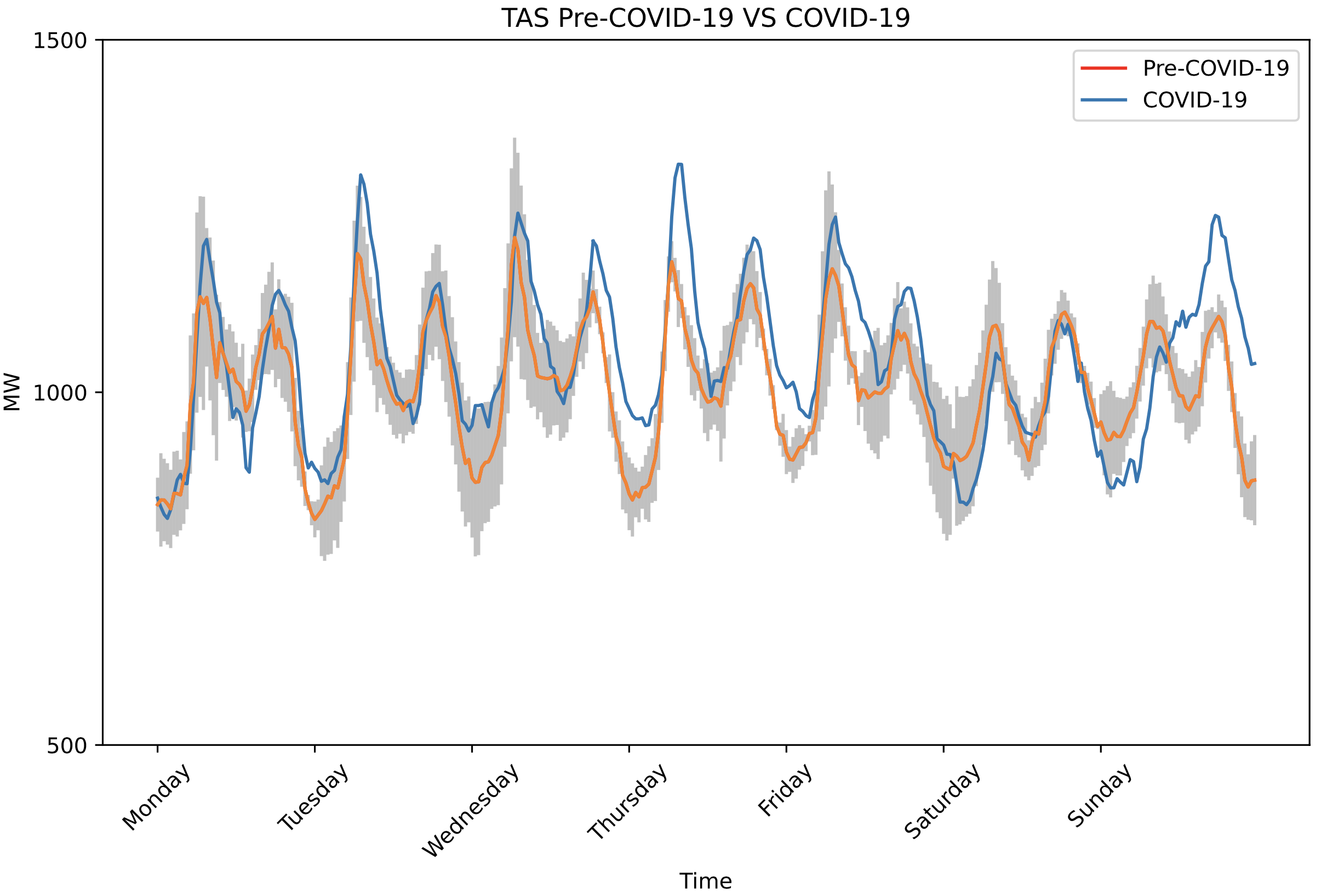}
\caption{The impact of Covid-19 on average weekly energy demand in the Australian states NSW, QLD, VIC, SA, and TAS.}
\label{fig:fig2}
\end{figure}

\subsection{Context and Data}

We obtained and analysed data from AEMO dataset \cite{cite_key1} on electricity demand and the ENTSO-E Transparency Platform dataset \cite{cite_key2} on European countries' total electricity load. The AEMO dataset holds aggregated price and energy demand data across 5 of the 6 Australian states in 30-minute intervals. The entire data series are taken from January 2017 to April 2020. The observations from March 2020 to April 2020 are used as a test set to coincide with the intervention period. The intervention period is determined by the presence of Covid-19 lockdowns to prevent the spread of the virus. The pre-intervention period is set in the absence of these measures, from January 2017 to February 2020. The ENTSO-E dataset contains the total electricity load of 37 European countries in 1-hour intervals. Our data series are obtained from January 2016 to March 2020 in Italy, Poland, Spain, Switzerland and four bidding zones of Sweden. The test data is from the 16th of March 2020 to the 30th of March 2020, when all countries except Sweden imposed Covid-19 lockdowns to inhibit the virus propagation. 

\subsection{Exploratory Data Analysis}
\label{summary}

\begin{figure}[!t]
\includegraphics[width=.15\textwidth]{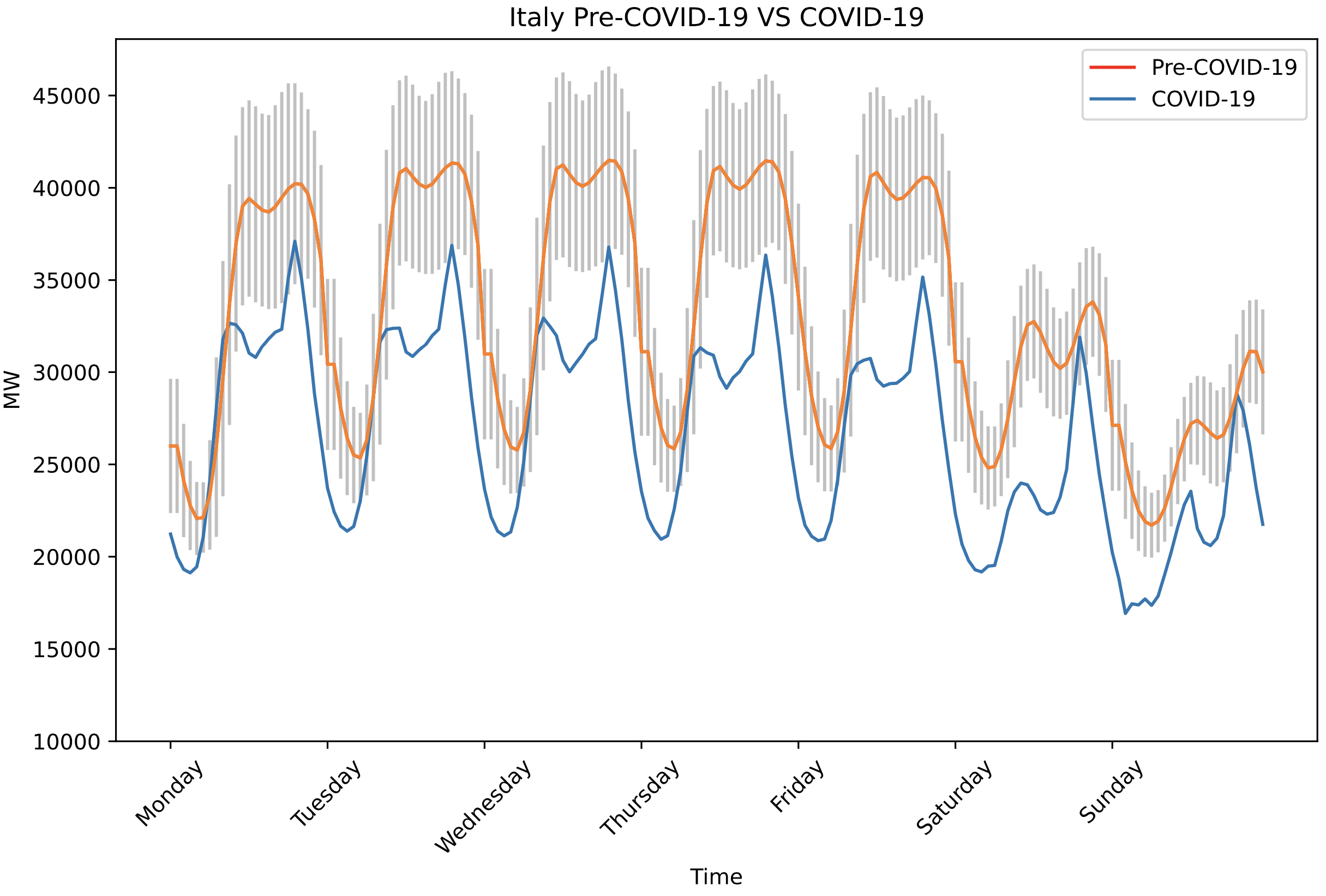}\quad
            \includegraphics[width=.15\textwidth]{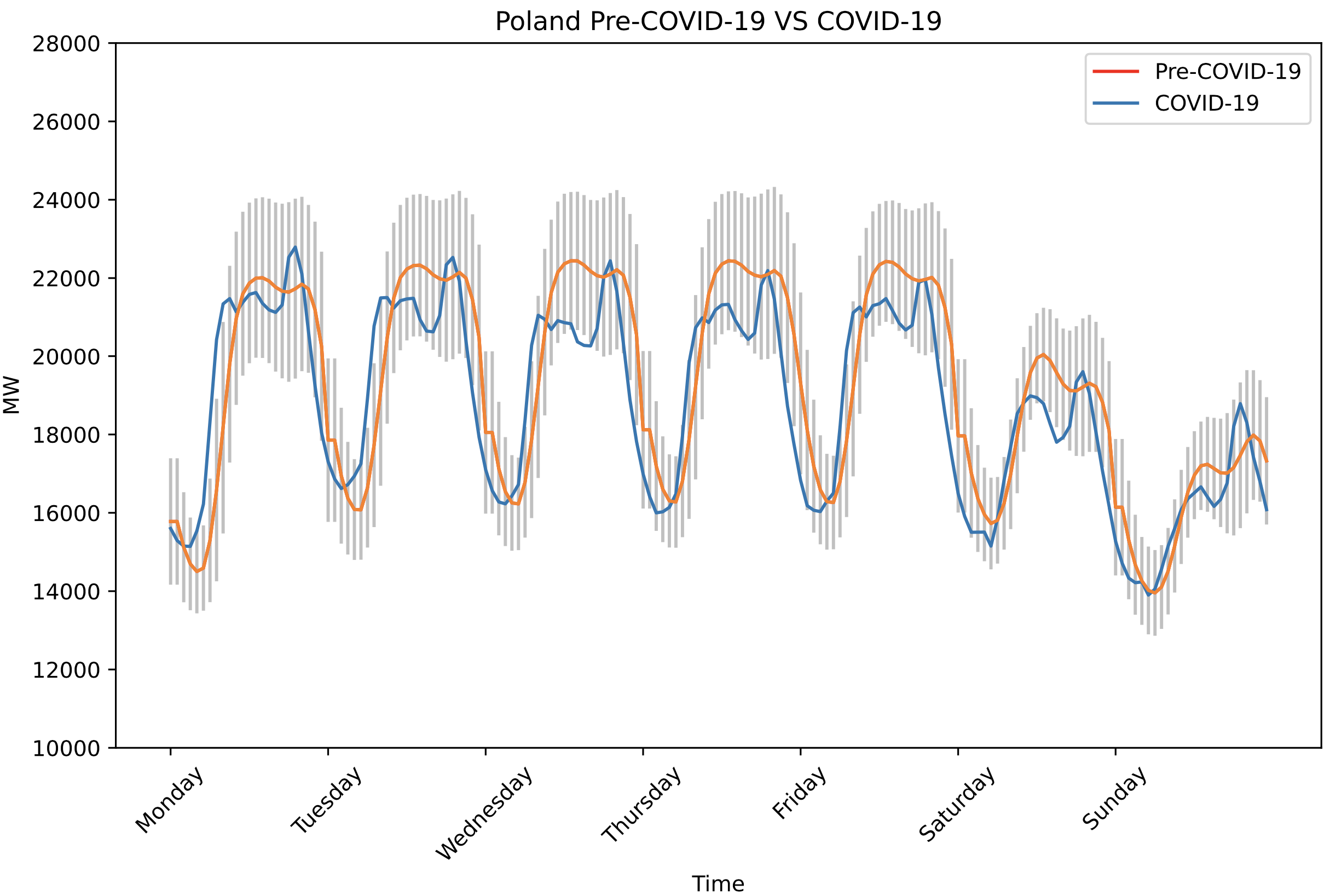}
            \includegraphics[width=.15\textwidth]{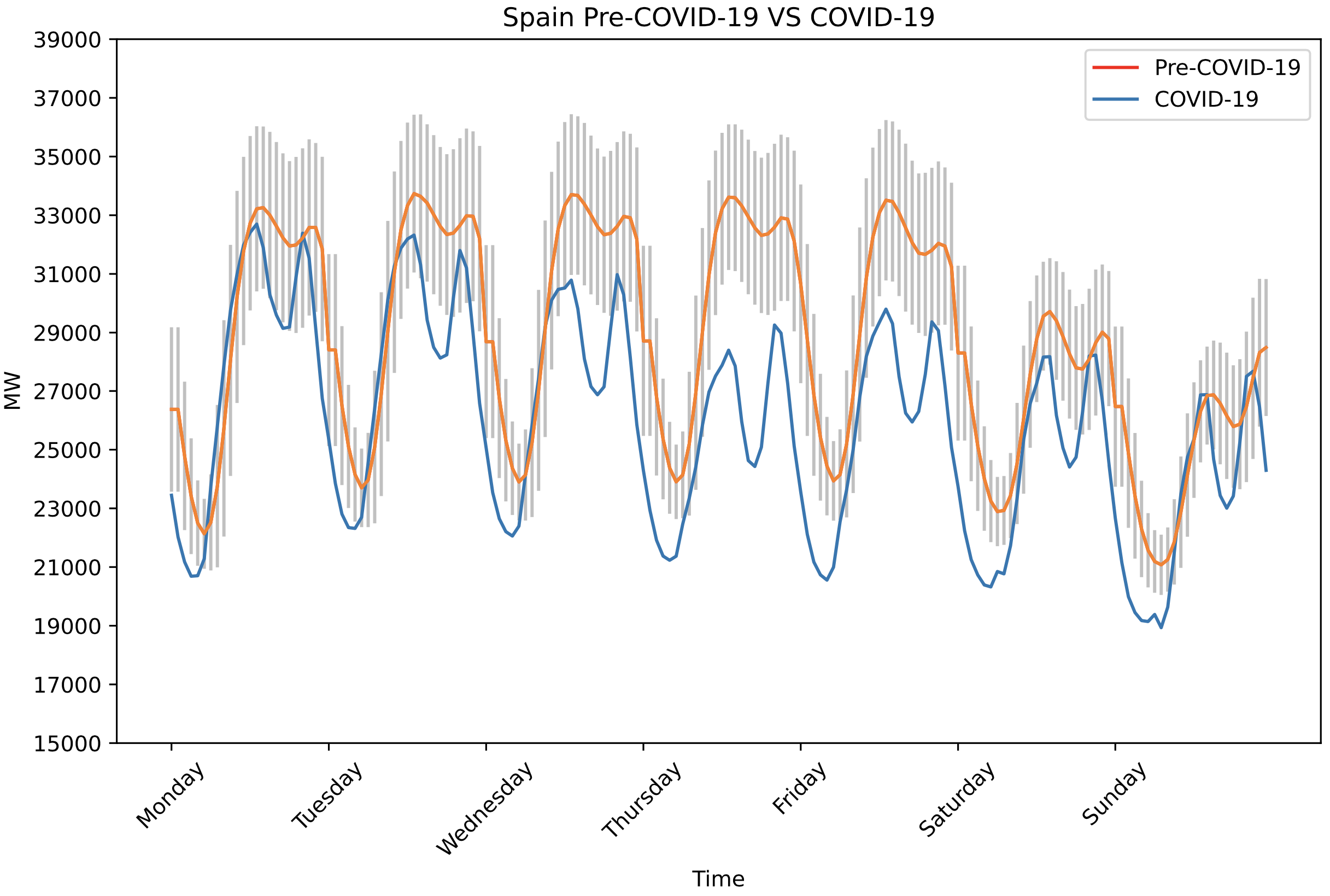}
            \qquad
            \includegraphics[width=.15\textwidth]{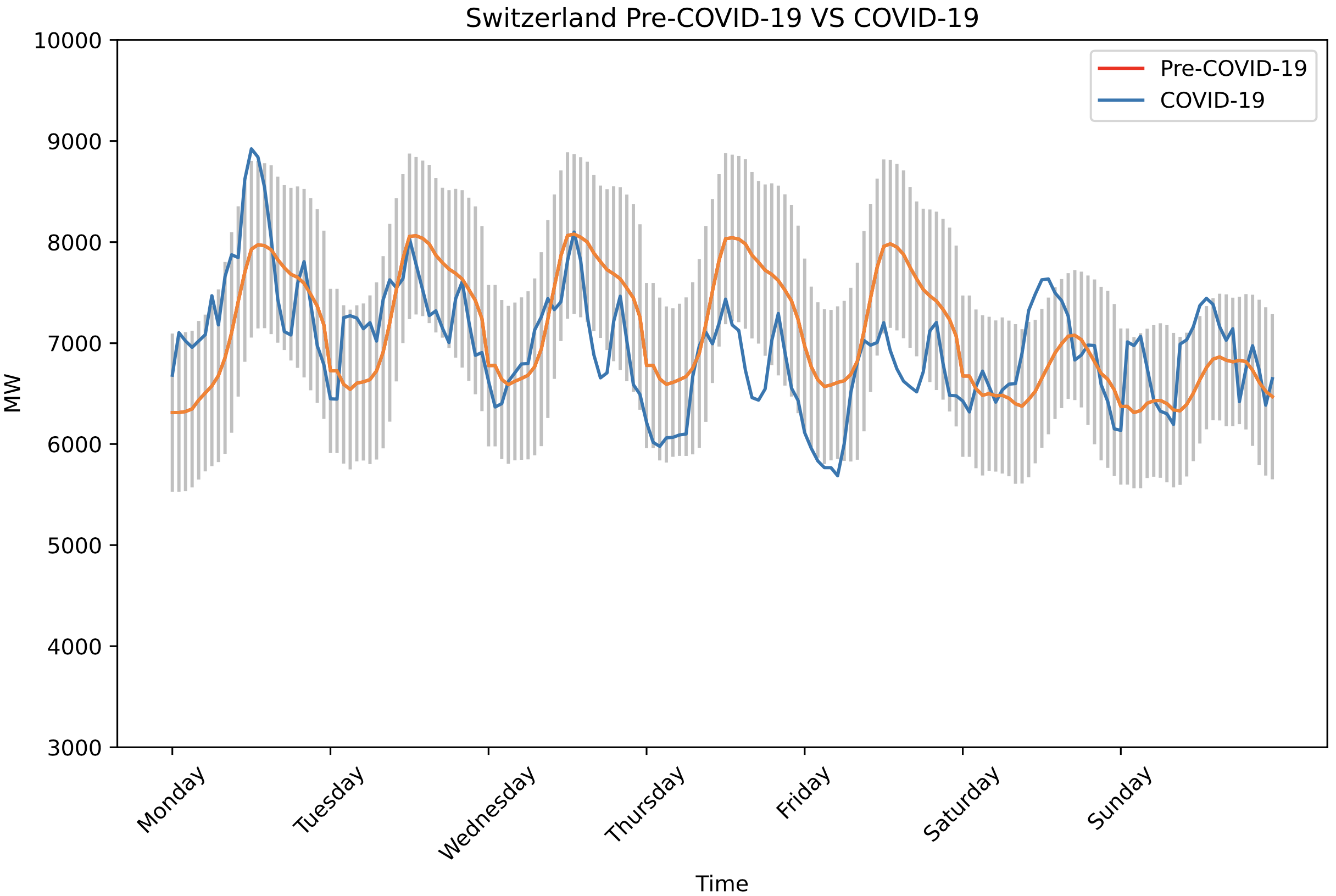}\quad
            \includegraphics[width=.15\textwidth]{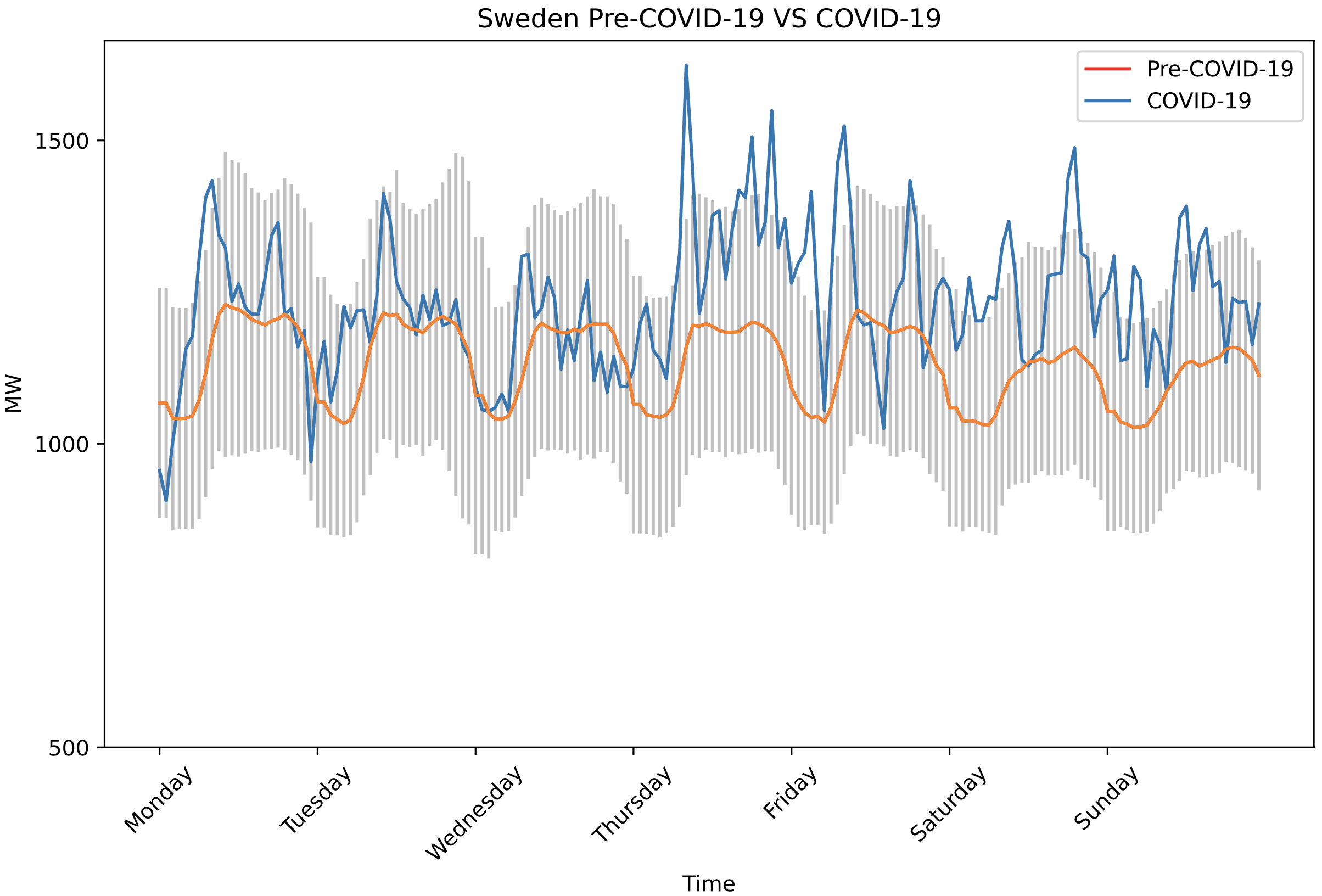}
\caption{The impact of Covid-19 on average weekly energy demand in Italy, Poland, Spain, Switzerland, and Sweden.}
\label{fig:fig3}
\end{figure}

Figure \ref{fig:fig2} displays the differences between Australian states' mean weekly energy consumption pre-Covid-19 and during the pandemic. Our primary variable of interest is electricity consumption
. The pre-treatment weekly consumption was averaged from January 2017 to February 2020, whilst the post-intervention data was from March 2020. The standard deviation was measured for each data point and illustrated in grey. From Figure \ref{fig:fig2},  we observe an apparent reduction in energy demand in New South Wales (NSW) after the Covid-19 advent. This is most evident during working hours on weekdays, where there is a clear vertical downward shift of the morning and evening peaks. The exogenous shock likely causes this shift as the energy demand curve for post-Covid-19 is not within the standard deviation of pre-Covid-19 data. 
Greater fluctuations are evident for the weekend, where the figure showed a significant decrease in energy demand on Sunday, likely due to mandatory lockdowns. The post-Covid-19 energy demand trend in Queensland (QLD) is atypical, with significantly reduced morning peaks. We speculate that solar energy may have contributed to this phenomenon. The shape of Friday's energy demand is similar to Saturday's, which may be due to a behavioural change in energy usage approaching the weekend. The figure thus validates that Tasmania (TAS) should be used as a control unit for the Australian data set as there is greater variability in energy demand for Victoria (VIC) and South Australia (SA), with no clear differences in Tasmania's energy demand. 

Figure \ref{fig:fig3} illustrates the differences between mean weekly energy consumption pre-Covid-19 and post-Covid-19 in 5 European countries. 
For Italy, the post-Covid-19 energy demand curve was vertically shifted downward, surpassing the lower boundaries of the standard deviation. The decrease in demand is most evident during maxima and minima. The evening peak is relatively higher than the morning peak during the pandemic likely due to the relatively lower energy usage in the morning, thus, magnifying the evening peak. Similarly, Spain showcases a significant decrease in energy demand. 
Both post-Covid-19 data for Switzerland and Poland are within the standard deviation. 
There is some reduction in energy usage during daytime in Poland and a relatively more apparent decrease on Thursdays and Fridays in Switzerland. Behavioural changes during Covid-19 may be a possible justification. The relatively small reduction during daytime prompts to analyse quantiles.

Four bidding zone data was available for Sweden; however, considering the population density and initial quantile analysis, Sweden bidding zone 1 (Sweden1) was selected for the control unit as no lockdowns were imposed in Sweden. 
The post-Covid-19 data for Sweden exhibited the most fluctuation; however, this could be justified due to the relatively small population. The variability is primarily within the standard deviation, validating Sweden as the control unit. 

\subsection{Baseline Models and Performance Measuring}
We provide comparisons of the DeepAR against three baseline \emph{local} time series forecasting models such as Seasonal Na\"ive, Exponential Smoothing (ETS), and the Trigonometric Box-Cox ARMA Trend Seasonal (TBATS) models; and two baseline \emph{global} NN models such as a simple Feedforward Neural Network (FFNN) and a canonical RNN. More details about them and their implementation are found in the Supplementary Material, Section C.

Time series forecast evaluation is complex and not straightforward, and no generally accepted consensus exists in the field \cite{hewamalage2022forecast}. Common error metrics in the field, such as Mean Absolute Scaled Error (MASE) and symmetric Mean Absolute Percentage Error (sMAPE), have problems. For example, the sMAPE is not symmetric, and the MASE, though in general a good measure, has the problem that in our application we have a relatively large test set that we predict from a fixed origin. As such, if a one-step-ahead na\"ive forecast is used in the denominator in the MASE, the MASE is not easily interpretable in its typical way of values larger than 1 being bad forecasts. Instead, as our test sets are large, with their sums amounting to large values, we opt to use the Weighted Absolute Percentage Error (WAPE) and the Weighted Root Mean Square Percentage Error (WRMSPE) as two scale-independent forecast metrics that can be used to compare forecast performance between different time series \cite{hewamalage2022forecast}. They are easily interpretable as percentages. Furthermore, we use the Mean Scaled Interval Score (MSIS) and the Continuous Ranked Probability Score (CRPS), which are probabilistic error metrics to evaluate probabilistic forecasting. The detailed definition of each one these metrics are provided in the Supplementary Material, Section D.

\subsection{Placebo and Hypothesis Testing}

The performance of the aforementioned forecasting algorithms is determined by using the point estimation considering only the control units. This permits us to confidently evaluate the treatment effect from the knowledge obtained by the baseline forecast errors under the absence of treatment effect and without focusing on the counterfactual outcomes. The prediction errors of control units should be considerably less and statistically significantly dissimilar from treated units (placebo test). In contrast, the error for treated units needs to be statistically significantly dissimilar from zero, implying a presence of the treatment effect. Therefore, we conduct hypothesis testing using both non-parametric statistical Wilcoxon signed-rank and rank-sum tests, which were implemented 
using the \texttt{scipy.stats} library in Python. That means we are first testing if the distribution of the difference between forecasts and true values is symmetric at about zero, and second, whether the values in one sample (treated units) are likely to be larger than the values in the other sample (control units). Hence, for our study, the hypothesis of interest are

\begin{equation}
\label{eq:15}
\begin{cases}
 H_0^{(1)} :\widehat{\Delta}_T = 0 \; \mbox{and} \;H_0^{(2)} :\widehat{\Delta}_{T}^i = \widehat{\Delta}_{T}^c \\
 H_A^{(1)} :\widehat{\Delta}_T \neq 0 \; \mbox{and} \;H_A^{(2)} : \widehat{\Delta}_{T}^i \neq \widehat{\Delta}_{T}^c,
\end{cases}
\end{equation}

\noindent where $\hat{\Delta}$ is the ATE as defined in Equation (\ref{eq:12}), and $c$ and $i$ denote the control and treated units, respectively.

With respect to the treated units, the rejection of $H_0^{(1)}$ offers substantial evidence for the presence of non-empty policy effects. Acceptance of $H_0^{(1)}$, on the other hand, is the placebo test affirming the null treatment effect over the control units. In addition, in $H_0^{(2)}$, we also test, by using the Wilcoxon rank-sum test, the statistical difference between the errors for treated and control units.

\section{Results}

In the following we present the results for the two use cases of Europe and Australia.

\subsection{European Data}

As we face the non-existence of true underlying impacts in the synthetic counterfactual approach, we first check the errors only of the instances not affected by the intervention, i.e., the control units, to determine the prediction errors for baseline methods. They are reported in Table~\ref{tab:table1} for the Sweden1 time series forecasting as the control group. As explained earlier in Section \ref{summary} and depicted in Figure~\ref{fig:fig3}, Covid-19 should shallowly impact Sweden1's energy demand, in contrast to Italy where the impact of Covid-19 is likely spread throughout the demand distribution.

Table~\ref{tab:table1} shows that the TBATS model, which is a local model that uses exponential smoothing to forecast time series with complex seasonal patterns, is able to outperform all other models, in terms of point prediction accuracy. However, it struggles to retrieve the true quantile distribution due to its non-probabilistic approach, which is depicted by the very low accuracy of the quantile predictions. When forecasts are probabilistic and observations are deterministic, CRPS and MSIS are critical metrics for comparing forecasting model accuracy. In terms of the forecast distribution, both the probabilistic DeepAR and FFNN global models performed better in obtaining prediction intervals with lower MSIS and CRPS values, resulting in a superior probabilistic forecast for control series.

\begin{table}[htb]
\caption{Error Metric results for the control units Sweden1 for European data and Tasmania for Australian data.}
\label{table:x}
\centering
\resizebox{0.48\textwidth}{!}{
\begin{tabular}{llrrrr}
\hline
    & METHOD       &             WAPE & WRMSPE & MSIS & CRPS  \\
    \hline
   \multirow{6}{*}{\textbf{European data}} &DeepAR        &           0.123 &    0.148 &  15.009 & \bf{0.090}   \\
  & RNN        &           0.133 &    0.164&  18.734 & 0.099       \\
  & FFNN         &           0.136&    0.172&  \bf{14.967} & 0.095\\
  & Seasonal Na\"ive   &        0.098&   0.131& 56.588 & 0.098\\
  & ETS     &            0.092 & 0.134&   41.824 & 0.168\\
   &TBATS       &            \bf{0.088}&   \bf{0.117}&   44.026 & 0.179\\
\hline
\hline
  \multirow{6}{*}{\textbf{Australian data}} & DeepAR        &          \bf{ 0.050}&\bf{0.061}&\bf{6.765}&\bf{0.045}\\
   &RNN       & 0.084&0.108&9.634&0.060\\
   &FFNN     &    0.765&1.355&186.305&0.694\\
   &Seasonal Naïve   &        0.113&0.130&61.771&0.096\\
   &ETS     &            0.204&0.241&100.856&0.381\\
   &TBATS     &  0.088&0.103&26.494&0.098\\
  \hline
\end{tabular}}
\label{tab:table1}
\end{table}


\begin{figure}
    \centering
    \subfloat[DeepAR model]{{\includegraphics[width=0.24\textwidth,height=2.5cm]{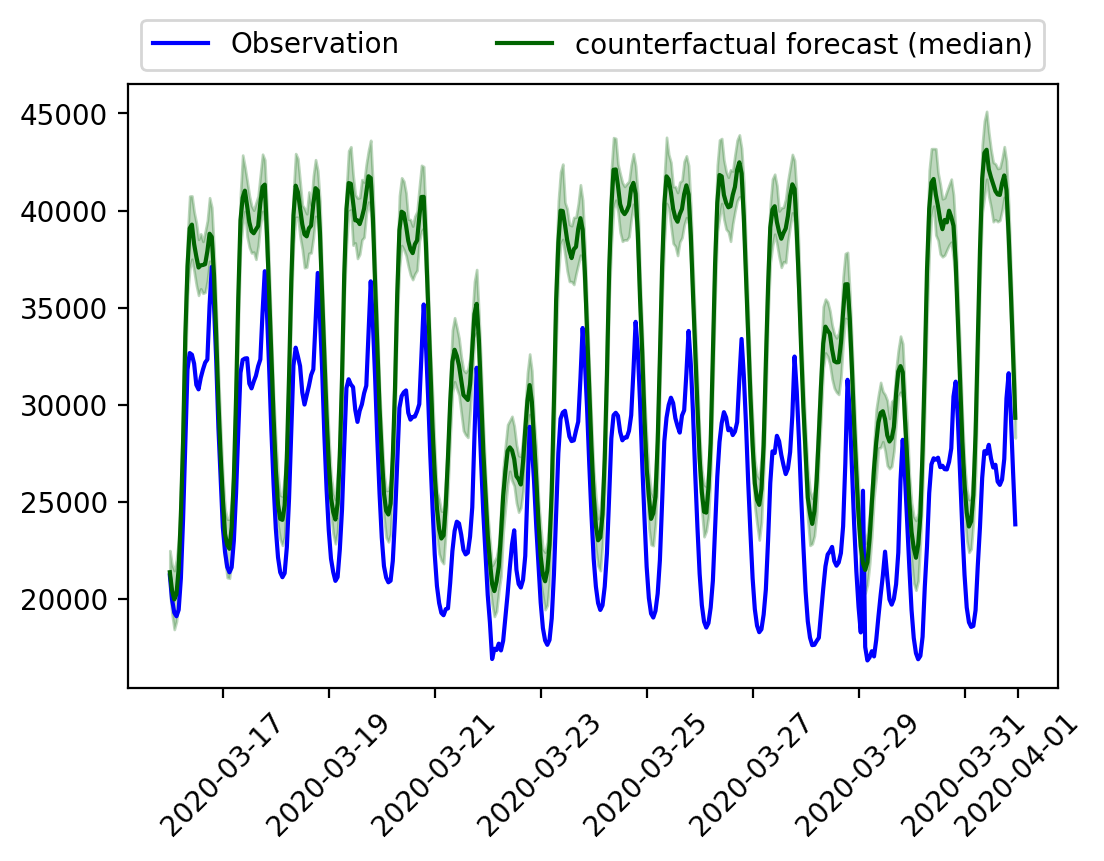} }
    {\includegraphics[width=0.24\textwidth,height=2.5cm]{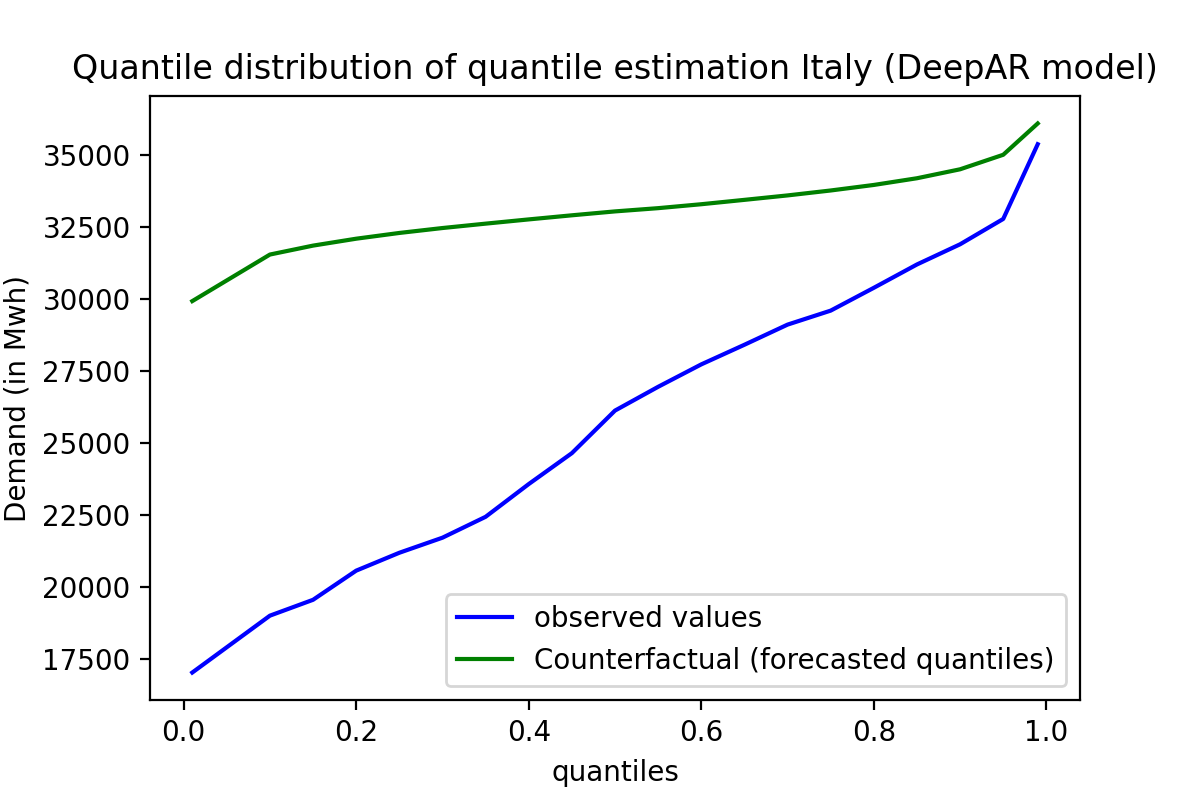} }}
    \qquad
    \subfloat[TBATS model]{{\includegraphics[width=0.24\textwidth,height=2.5cm]{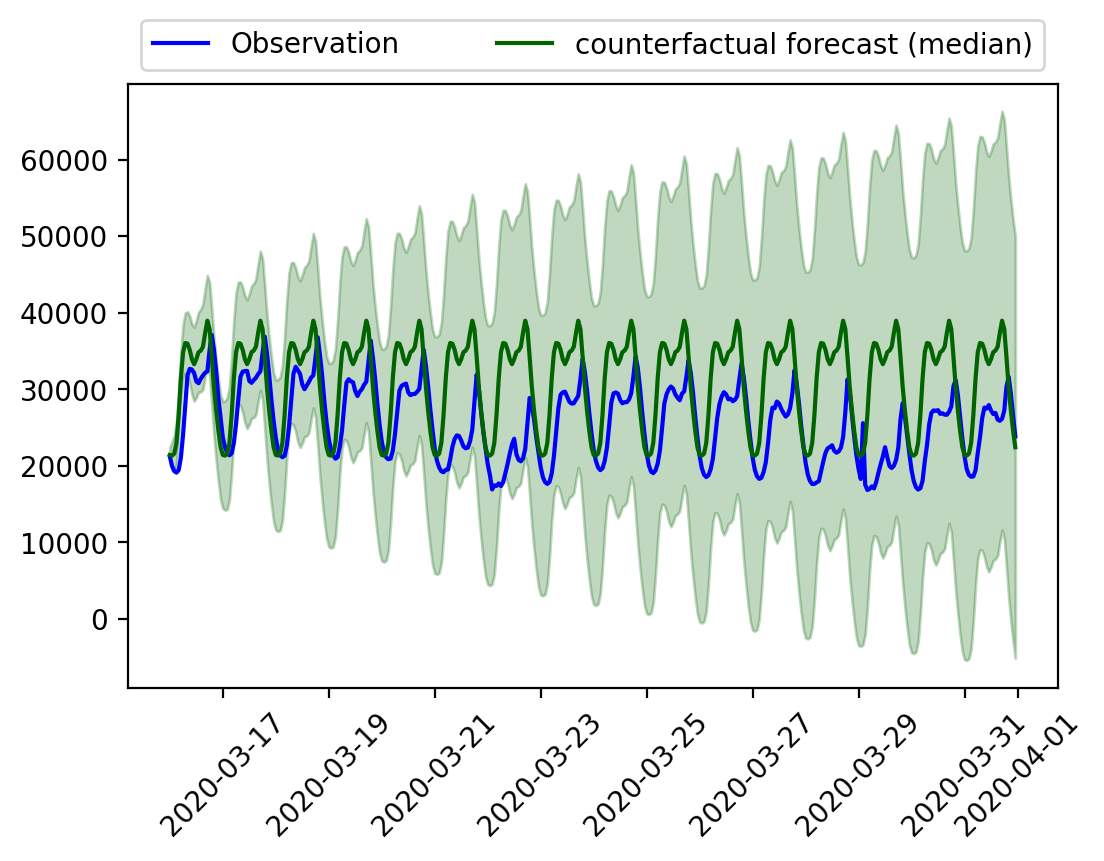} }
    {\includegraphics[width=0.24\textwidth,height=2.5cm ]{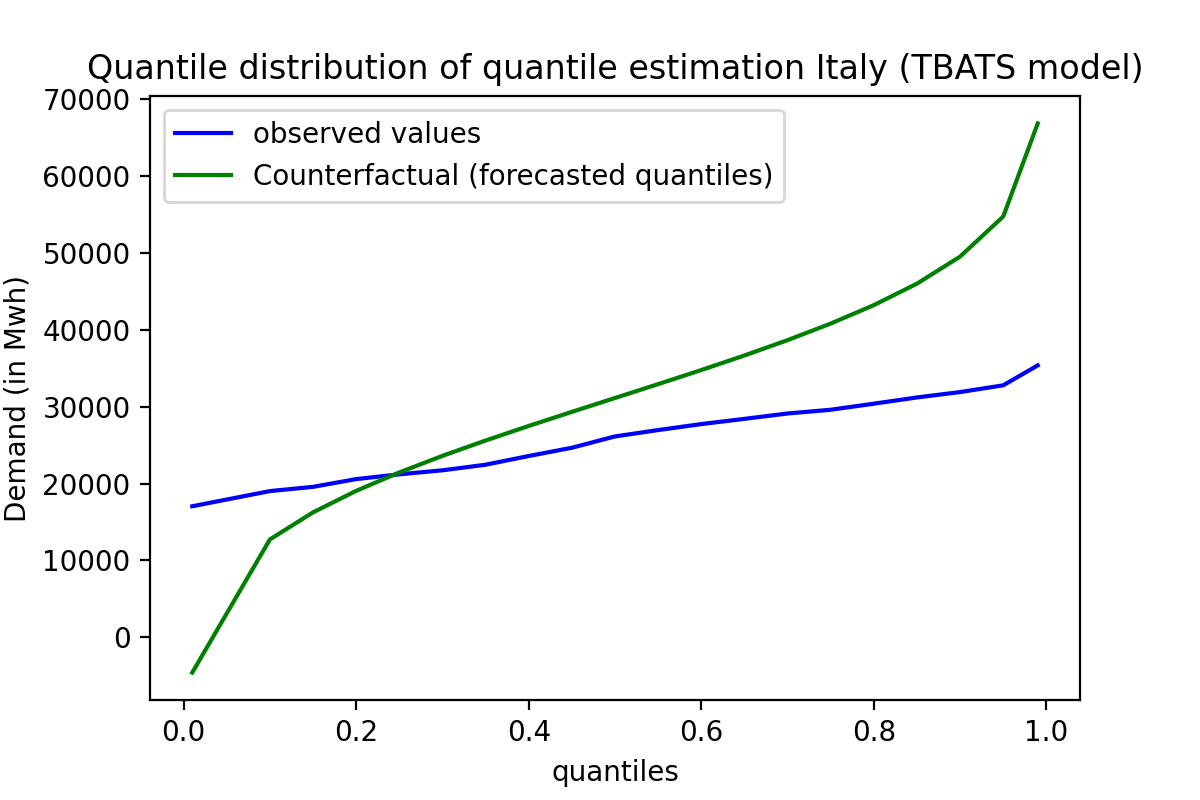} }}
    \caption{Point counterfactual forecast of the DeepAR and TBATS model with 10\% and 90\% prediction interval (left) and the quantile distribution of the counterfactual forecast (right) for Italy.}
    \label{fig:fig7}
\end{figure}

Further, we analyse the counterfactual forecasting results for treated units, focusing only on the comparison between DeepAR and TBATS, as the best probabilistic and non-probabilistic methods from the previous analysis (Table~\ref{tab:table1}). Figure~\ref{fig:fig7}a (right) demonstrates how well (taking Italy as example) our global probabilistic baseline DeepAR model captures the distribution patterns from the unaffected observed data. We observe that the counterfactual distribution is more compatible with the pre-intervention data (without the effects of the lockdowns) but not the observed post-intervention data, confirming, therefore, the impact of lockdown measures on reducing the electricity demand. Point and distributional counterfactual estimations (left and right of Figure~\ref{fig:fig7}a, respectively) show this reduction, but the distributional forecast makes clearer the non-uniform aspect of this impact. Despite the counterfactual quantile function showing a decrease in demand spread across the entire distribution (right of Figure~\ref{fig:fig7}a), it evidences the stronger average effect in the troughs and the less substantial impact over the peaks of demand. The fact that DeepAR's median forecast has remained in the centre of the model's prediction confidence intervals (left of Figure~\ref{fig:fig7}a), ensures the quality of its probabilistic estimation. The DeepAR model also encapsulates the seasonal pattern for the post-intervention data and recovers the forecast distribution better than the TBATS (Figure~\ref{fig:fig7}b) model, which justifies the values reported in probabilistic error metrics of the models in Table~\ref{tab:table1}.

The evidence of DeepAR's good performance on capturing seasonality is demonstrated in Figure~\ref{fig:fig7}a, where the weekend distributions of March 21st and 22nd 2020, are forecasted better by the model. The TBATS model, on the other hand, failed to capture these details, and the forecast with a wider prediction interval rendered the model insignificant (left of Figure~\ref{fig:fig7}b). 


\begin{table*}[htbp]
\centering
\caption{ Average Treatment Effect (in \%) on energy consumption using DeepAR model.}
\centering \label{table:Table II}
\begin{tabularx}{\textwidth}{L|R|R|R|R|R|R|R|R|R|R}
\hline
\multirow{2}{*}{\textbf{Quantiles}} &   \multicolumn{5}{>{\bf}c|}{European data} &            \multicolumn{5}{>{\bf}c}{Australian data}\\
\cline{2-11}
 &\bf{Italy} &\bf{Poland} &\bf{Spain} &\bf{Switzerland} &\bf{Sweden1} & \bf{NSW} & \bf{QLD} & \bf{VIC} & \bf{SA} &\bf{TAS}\\
 \hline
0.1&-65.84&-29.90&-36.59&-19.97&-13.40&-13.35&-14.30&-9.21&-42.63&-5.12\\
0.25&-52.31&-24.38&-28.72&-18.27&-12.98&-8.71&-13.30&-4.57&-13.87&-2.05\\
0.3&-49.43&-23.22&-25.52&-17.90&-12.86&-7.18&-11.75&-2.36&-7.41&-1.64\\
0.4&-38.92&-17.65&-20.06&-15.50&-12.25&-3.50&-8.40&-0.82&-1.92&-0.83\\
0.5&-26.43&-10.51&-15.01&-15.42&-11.26&-2.17&-6.67&0.09&2.04&0.77\\
0.6&-20.03&-5.70&-11.51&-14.21&-10.85&-1.44&-4.89&1.72&4.91&2.45\\
0.75&-14.08&-3.68&-7.31&-12.44&-9.54&-0.74&-1.28&5.22&8.42&3.97\\
0.8&-11.74&-3.41&-5.85&-12.46&-7.98&-0.64&0.29&6.53&9.96&4.40\\
0.95&-6.79&-2.54&-2.38&-9.53&-3.24&1.01&4.23&8.25&12.85&5.88\\
0.99&-2.03&-2.60&-0.22&-2.91&15.07&-1.33&3.64&6.10&11.27&6.54\\
\hline
\multirow{2}{*}{\textbf{\emph{p-values}}}&4.7e-07/  & 4.7e-07/  &4.7e-07/  & 4.7e-07/  & 0.005 & 6.7e-06/  & 0.001/ & 0.87/ & 0.87/  & 0.06 \\
&4.7e-07  & 4.7e-07  &4.7e-07  & 4.7e-07  & - & 0.00025  & 0.00025 & 0.66 & 0.78  & - \\
\hline
\end{tabularx}
\label{tab:table2}
\end{table*}

Table~\ref{tab:table2} shows the Average Treatment Effect (ATE) of the impact of Covid-19 on electricity demand in European countries using the DeepAR model, in terms of their demand change in percentage (a positive value indicates an increase in demand, while a negative value indicates a decrease in demand). Counterfactual probabilistic distribution estimation allows for the analysis of effects by quantile, rather than just the median effects as in point forecast estimation. There is a greater impact showing a significant decrease in demand in the troughs of the distribution in Spain and Poland, whereas there is a massive reduction in consumption in Italy from the troughs to the 75th percentile of the distribution. Whereas the impact of Covid-19 on Sweden1 is low in comparison to other countries (visual evidences by boxplot graphs analysis are provided in the Supplementary Material, Section F).

Finally, hypothesis testing was performed to ensure the treatment's relevant effect and the null effect on the control unit. We can confirm significant effects for all treated units and that these effects are statistically different from those of the control unit once both tests returned very low p-values of 4.7e-07, as shown in the last line of Table~\ref{tab:table2}. Although the control unit presented a not-so-high p-value of 0.00527, we can conclude an almost null effect for Sweden1.

\subsection{Australian Data}

 As discussed previously in Section \ref{summary}, given the degree of state restriction, the intervention effect in Tasmania is low and less than in the other states. Hence, such as in the European case, we start by verifying the results for the control units (Tasmania) in the Australian energy use case which are shown in Table~\ref{tab:table1}. We see that here, the DeepAR model outperformed all benchmark models. Along with the better distribution recovery ability, the model surpassed all models in analysing the median point forecasting errors. After the DeepAR model, the TBATS model is adequate for point forecasting, whereas the probabilistic error metric of the RNN model is comparable to DeepAR in predicting probabilistic forecasts.

\begin{figure}
    \centering
    \subfloat[NSW]{{\includegraphics[width=0.24\textwidth,height=2.5cm]{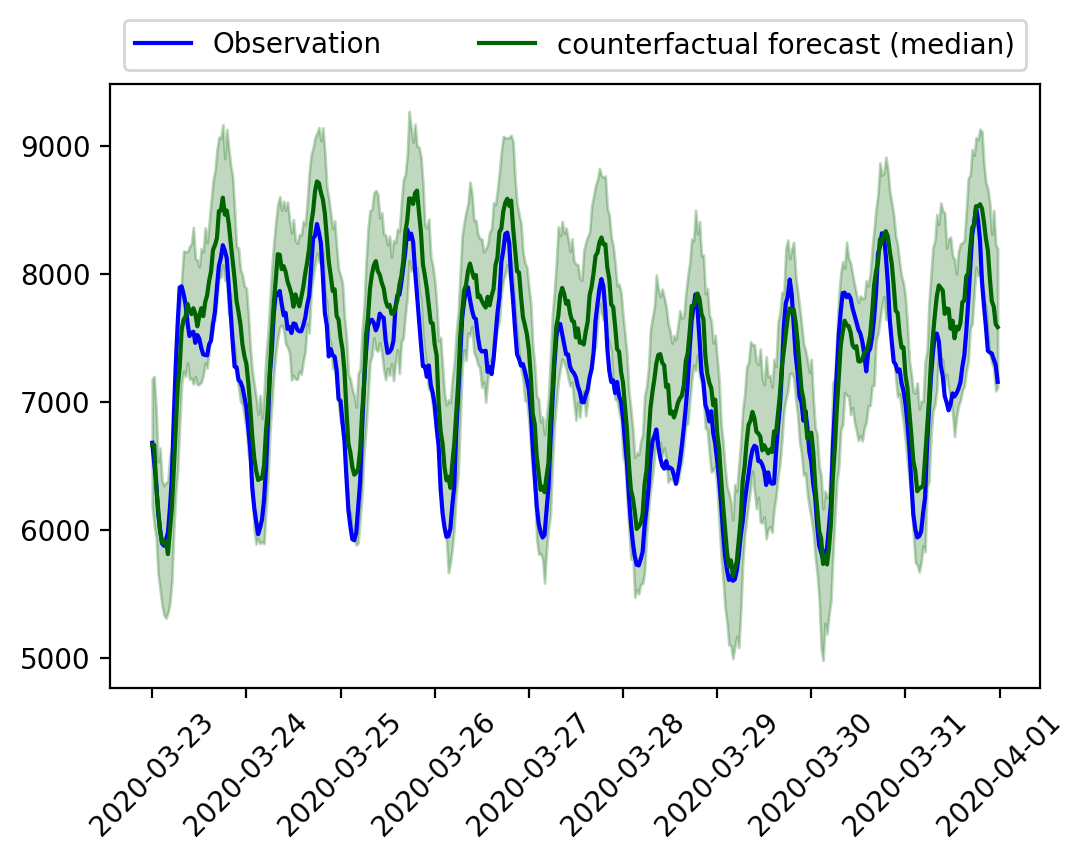} }
    {\includegraphics[width=0.24\textwidth,height=2.5cm]{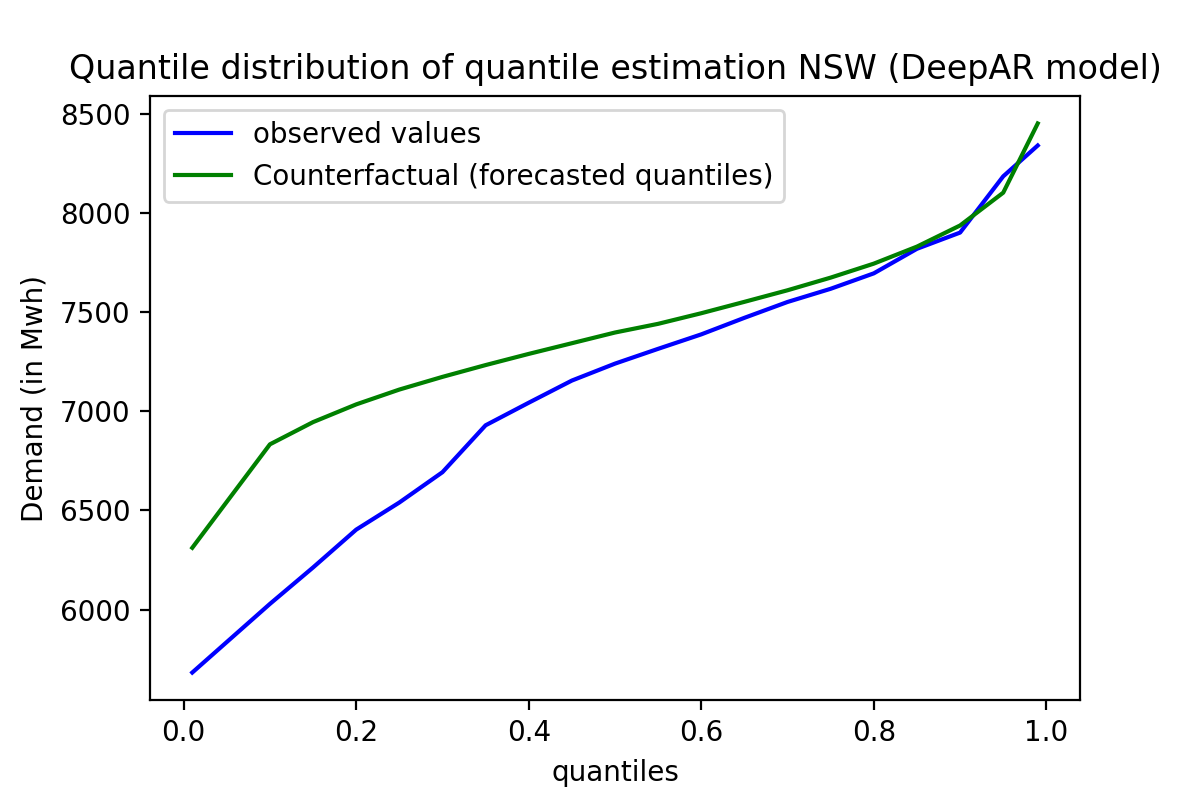} }}
    \qquad
    \subfloat[TAS]{{\includegraphics[width=0.24\textwidth,height=2.5cm]{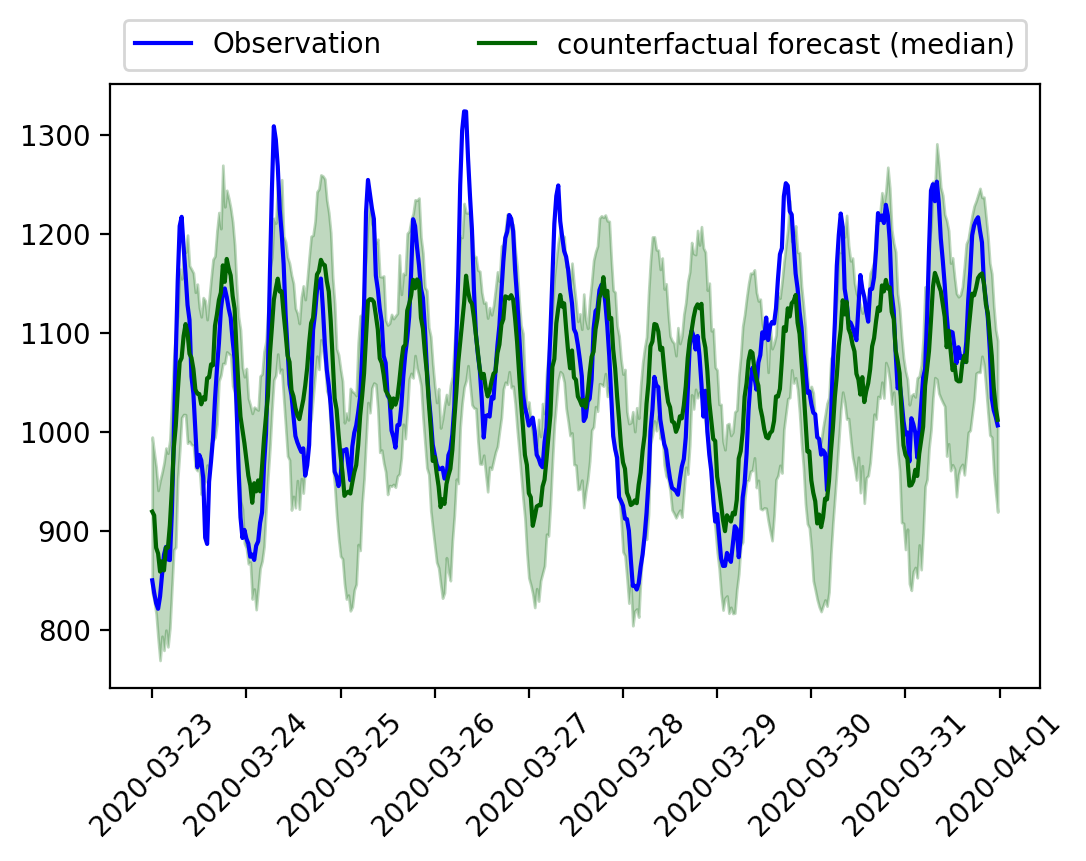} }
    {\includegraphics[width=0.24\textwidth,height=2.5cm ]{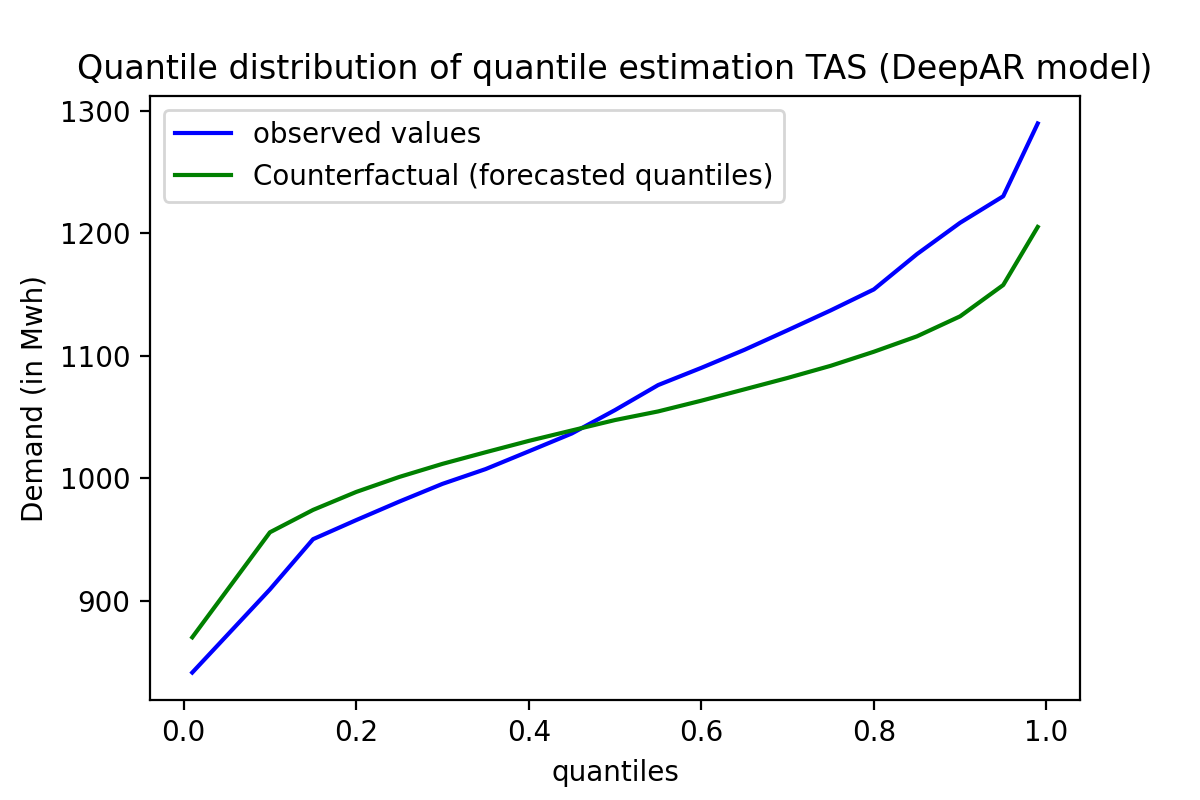} }}
    \caption{Point counterfactual forecast with 10\% and 90\% prediction confidence interval (left) and the quantile distribution of the counterfactual forecast (right) of the DeepAR model.}
    \label{fig:fig11}
\end{figure}

Moving on to the analysis of the Australian treated states, we focus our analysis on the DeepAR results, considering it is the best performing method in the previous baseline control error discussion. The DeepAR point forecasting for NSW (left of Figure~\ref{fig:fig11}a) shows that the peak observed in the daily distribution of the intervention period falls within the model's prediction confidence interval, demonstrating a small impact during the maximum periods of consumption. On the other hand, the model's result shows considerable reduction in the lower quantiles. Both effects are demonstrated by the right plot of Figure~\ref{fig:fig11}a, which displays the estimated counterfactual quantile distribution. In the case of Tasmania (control unit), the observed data peak is outside the prediction confidence interval (left of Figure~\ref{fig:fig11}b), and the control unit demonstrates a slight positive effect over the demand peaks, which might be due to warm weather period as Tasmania did not impose lockdowns, and had no Covid-19 cases at the time, so that large behaviour changes in the population seem unlikely.

Table~\ref{tab:table2} shows the ATE of the DeepAR model for each quantile of the forecasted counterfactual distribution and the percentage change in demand with respect to the observed data. It can be seen that there is a reduction in demand for NSW throughout the entire forecasted probability distribution, whereas for the other states, the demand reduction is only observed in the troughs, and the demand in peak hours contrarily increases. Interestingly, we can see an almost null effect in the median of the electricity consumption distribution in Victoria and South Australia after the lockdown (0.09\% and 2.04\% of change, respectively); however, the distribution changes completely, with higher peaks of consumption and lower troughs after the restriction measures.

The p-values of NSW and QLD (Table~\ref{tab:table2}) confirm relevant ATE and that these effects are different in terms of statistical significance from those observed in the control unit, respectively. However, the results for VIC and SA do not have a significance to confidently classify them as treated or confirm that the effects come from the lockdown restrictions. They are also not statistically different from TAS. These results allow us to conclude that the lower changes observed in VIC and SA could also be related to other factors (such as weather) more than the lockdown. On the other hand, the test confirms the null effect over TAS with a p-value of 0.06.

This information can be used by an energy control center and generator/distributor to identify the actual and precise uncertainty in electricity demand. During policy interventions, this approach allows policymakers to make critical decisions in the power system environment, such as resource planning, voltage regulation, and accurate load/price forecasting once the analysis provides us with the effects of each quantile. The impact of policy implementation can be determined at the quantile level of the distribution, and as a result, policy decisions or policy amendments can be made accordingly. 

\section{Conclusion}

Load forecasting under uncertainty is critical for intelligent power control systems. While ensuring power grid balance, one of the major issues in demand response programs is responding to external interventions that cause power system instability. 
The proposed framework models the conditional quantile functions to build counterfactual distribution outcomes and estimate the distributional effect caused by non-uniform interventions. To the best of our knowledge, no work in the electricity literature has used a probabilistic forecasting model with GFMs to evaluate the uncertainty in demand/price affected by any intervention. Many previous works have, however, estimated the overall percentage change in demand due to Covid-19, but without taking a probabilistic approach to analyse the effect of Covid-19 on the electricity industry. 

Causal impact analysis becomes more difficult in the presence of a non-uniform pattern of impact or data with skewness and heavy tails. Our framework's causal effect estimation of demand provides energy sector managers with insights for power system planning, operations, and load forecasting to make the right decision during the treatment uncertainty. One limitation of this current work is the absence of employing a hyperparameter optimisation process, which we should address in future work. Other external predictors, known to highly influence the energy demand (e.g., the weather), can be inputted into the model to boost model accuracy. Our work can be expanded to analyse different types of electricity markets like residential, non-residential, commercial, and others. The proposed probabilistic GFM-RNN based approach for counterfactual estimation has shown that training a model globally from non-linear, unassociated time series of different scales and mapping complicated patterns can be done effectively.

\section*{Acknowledgments}

This research was supported by the Australian Research Council under grant DE190100045, Monash University Graduate Research funding and MASSIVE - High performance computing facility, Australia.

\bibliography{IEEE_TPWRS_reviewed.bib}{}
\bibliographystyle{IEEEtranN}
\end{document}


%
\title{Supplementary material for \\ $\ $ \\Causal Effect Estimation with Global Probabilistic Forecasting: A Case Study of the Impact of Covid-19 Lockdowns on Energy Demand}

\author{
  \IEEEauthorblockN{
Ankitha~Nandipura~Prasanna\IEEEauthorrefmark{1},
Priscila~Grecov\IEEEauthorrefmark{1},
Angela~Dieyu~Weng\IEEEauthorrefmark{2},
and Christoph Bergmeir\IEEEauthorrefmark{1}
  }
  \vspace{3mm}

\IEEEauthorblockA{\IEEEauthorrefmark{1}Department of Data Science and Artificial Intelligence, Monash University, Melbourne, Australia.}

\IEEEauthorblockA{\IEEEauthorrefmark{2}Lauriston Girls' School, Melbourne, Australia
}}

\maketitle

\begin{abstract}
\noindent This supplementary material includes additional information to support the findings presented in the paper ``Causal Effect Estimation with Global Probabilistic Forecasting: A Case Study of the Impact of Covid-19 Lockdowns on Energy Demand," co-authored by Ankitha N. Prasanna, Priscila Grecov, Angela D. Weng, and Christoph Bergmeir. The supporting material includes a more detailed explanation of the baseline models that were used for comparison with the DeepAR model, as well as evaluation metrics for measuring the performance of the models. This material also includes detailed results from both datasets as well as additional empirical results.
\\
\end{abstract}

\IEEEpeerreviewmaketitle

\section{Introduction}
\label{intro}
This is the supplementary material for the paper ``Causal Effect Estimation with Global Probabilistic Forecasting: A Case Study of the Impact of Covid-19 Lockdowns on Energy Demand," co-authored by Ankitha N. Prasanna, Priscila Grecov, Angela D. Weng, and Christoph Bergmeir. The material includes a brief explanation of the baseline models and their functions, as well as the packages used to implement them. The evaluation metrics used to measure the accuracy of these models are also described in detail. This material also includes additional figures that support our findings in the paper.

The supplementary material is organized as follows. 
Section B contains the additional information on related works and details on baseline models are described in section C. Section D contains the details of performance measuring and accuracy scores. The additional figures of exploratory data analysis are presented in Section E. Finally, the additional figures and full tables for the result section of our paper are presented in Section F. 

\section{Additional information on Related work (section II)}

\citet{kwekha2021coronavirus} conducted an experimental study to analyse the machine learning applications used in Covid-19 disease, with Logistic Regression being the most commonly used machine learning algorithm. \citet{liu2021aged} used Gaussian mixture modelling to analyse the demand data to identify the demand profile based on the typical magnitude of energy use, the timing of peak demands and shapes of energy use patterns. In addition, the impact of the Covid-19 pandemic on gaseous and solid air pollutants is estimated using ARIMA model by \citet{rybak2021impact}. 

Several studies have been conducted to analyse the energy demand data of many countries around the world, considering the behaviour of the level of restrictions imposed. \citet{beck2021australia} conducted a survey in the middle of March 2020 to measure the people's attitude on changes in travel activity due to lockdown. The comparison of Melbourne and Sydney CityMapper Mobility Index curves begins to fall precipitously during the lockdown, with weekly household trips dropping by half. The aviation industry took a significant hit due to the border closure in Australia, with 47\% of passengers cancelling their flights. \citet{percy2020covid} conducted a descriptive analysis of the percentage change in average weekly electricity demand in selected Australian states, which revealed the greatest demand reduction in New South Wales, almost no change in Queensland and Victoria, and a slight increase in Tasmania and South Australia.

Furthermore, the EU's energy consumption fell by around -12.6\%. Italy was the first country to be hit by the pandemic, and by the end of February, it had imposed lockdown measures. Italy is a highly informative case study due to the short-term economic impact of gradually increasing the severity of restrictions and the possible path for economic recovery. According to \citet{bompard2020immediate}, the demand reductions in Italy, Spain, Poland, Switzerland, and Sweden are -20.9\%, -16.9\%, -6.5\%, +0.3\%, and -0.3\%, respectively. Sweden's average daily load profile (ADLP) is relatively close to 2019 and 2020 and has never imposed/milder confinement restrictions. The economic impact of the Swedish measures is less severe than seen elsewhere. Sweden is consensus-driven \cite{orlowski2020four}, where there are cooperations between the state and its citizenry, and it is also referred to as a society of state individualism. However, Poland has been grappling with the issue of rising energy demand \cite{czosnyka2020electrical}, necessitating electricity imports. Reduced energy consumption did not result in lower energy imports in Poland, causing the country's economy to slow. Whereas Switzerland has 26 cantons, each canton experienced a wide range of reductions ranging from extreme reduction to mild reduction \cite{janzen2020electricity}.

Finally, the experiments to analyse the Energy demand change in Australia and European countries due to the Covid-19 crisis were limited in visualising and analysing the change in demand using simple statistical tools. The Causal effect of the Covid-19 intervention were not studied in detail by examining the quantiles of the energy demand distribution.

\section{Detail Description of the Baseline Models}
\noindent Here we provide more information about the baseline models used to challenge our proposed framework. All these models, as such as DeepAR, were implemented using the GluonTS\footnote{https://gluon-ts.mxnet.io} library, a Python library for deep-learning-based time series modelling \cite{alexandrov2020gluonts}, as follows:

\begin{itemize}
\item{\emph{Seasonal Na\"ive (local)}: a seasonal na\"ive forecaster where each forecast equals the last observed value from the same season. It was implemented using the \texttt{gluonts.model.seasonal\_naive} package. The Seasonal Na\"ive model gives a na\"ive lower bound for forecasting performance that can be achieved.}
\item{\emph{ETS (local)}: the Exponential Smoothing (ETS) model (\citet{hyndman2008forecasting}) from the \texttt{forecast} R package. To forecast, ETS uses weighted averages of past observations, with the weights decaying exponentially as the observations go further back in time. We implement ETS using the wrapper \texttt{RForecastPredictor} from GluonTS which calls the R \texttt{forecast} package.}
\item{\emph{TBATS (local)}: the Trigonometric Box-Cox ARMA Trend Seasonal (TBATS) model, also from the \texttt{forecast} R package, uses an automated combination of Fourier terms with an exponential smoothing state space model and a Box-Cox transformation (\citet{de2011forecasting}). Identically to ETS, it was implemented by using \texttt{RForecastPredictor} from GluonTS.}
\item{\emph{FFNN (global)}: a global FFNN uses a simple multi-layer perceptron (MLP) model predicting the next target time steps given the previous ones. It was implemented using the \texttt{gluonts.model.simple\_feedforward} package.}
\item{\emph{RNN (global)}: a canonical RNN model with LSTM as cell-type, which was implemented using the \newline \texttt{model.canonical.CanonicalRNNEstimator} class in GluonTS.}
\end{itemize}

\section{Details on Performance Measuring and Accuracy Scores}
The metrics used to measure the performance accuracy of all models are defined as follows, and they are evaluated in the prediction range $h=T-T_{0}$, i.e., post-intervention:

\begin{subequations}
\label{eq:14}
\begin{align}
    \textit{WAPE} = & \frac{\sum_{t=n+1}^{n+h}\left|Y_t-\hat{Y}_t\right|}{\sum_{t=n+1}^{n+h} Y_t},\label{eq:14a}\\
    \textit{WRMSPE} = & \frac{\sqrt{\frac{1}{h}\sum_{t=n+1}^{n+h}\left(Y_t-\hat{Y}_t\right)^2}}{\frac{1}{h}\sum_{t=n+1}^{n+h} Y_t}, \label{eq:14b}\\
    \textit{MSIS} =& \frac{1}{h}\,
    \dfrac{\left(
       \splitdfrac{\splitdfrac{\sum_{t=n+1}^{n+h} (U_{t}-L_{t})}
        {{} + \frac{2}{\alpha} (L_{t}-Y_{t}) \mathds{1}\{Y_{t}<L_{t}\}}}{+ \frac{2}{\alpha} (Y_{t}-U_{t}) \mathds{1}\{Y_{t}>U_{t}\}}\right)}
        {\frac{1}{n-S}\sum_{t=S+1}^{n}|Y_t-Y_{t-S}|},\label{eq:14c}\\
    \textit{CRPS} =& \int_{0}^{1} \! 2\mathcal{L}_{\tau}(\widehat{CDF}^{-1}(\tau),z) \, \mathrm{d}\tau. \label{eq:14d}
\end{align}
\end{subequations}

 Here, $Y_t$ and $\hat{Y}_t$ are actuals and forecasts, respectively, $n$ the in-sample size, $L_t$ and $U_t$ in (\ref{eq:14c}) are the lower and upper bounds of the corresponding prediction interval for period $t$, $S$ the seasonal frequency, $\alpha$ defines the prediction interval (i.e., $\alpha=0.05$ for a 95\% interval), and $\mathds{1}$ is the indicator function (being 1 if $Y_t$ is within the postulated interval and 0 otherwise). Finally, in (\ref{eq:14d}) we have $\widehat{CDF}^{-1}$ representing the quantile function (indicated by the inverse of the cumulative distribution function), and $\mathcal{L}_{\tau}$ the quantile loss which is expressed by $\mathcal{L}_{\tau}(Y_t,\hat{Y_t}) = max[\tau(\hat{Y}_t-Y_t),(\tau - 1)(\hat{Y}_t-Y_t)]$, where $\tau$ is a certain quantile of the predictive distribution. Lower values of MSIS and CRPS indicate better prediction intervals.

\section{Additional Figures of Introduction and Exploratory Data Analysis (Sections I and IV.B)}
\label{figs1}

\begin{figure*}[ht!]
\centering
\includegraphics[width=\textwidth]{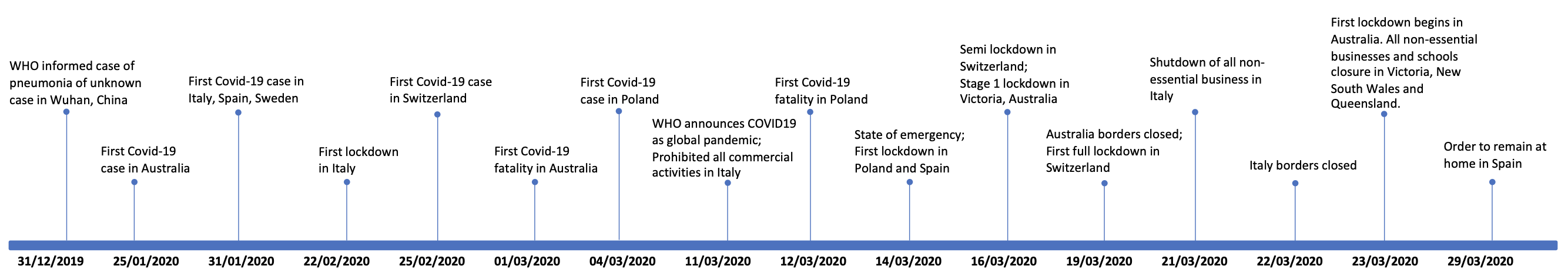}
\caption{Timeline of first lockdown in Australia and European countries.}
\label{fig:fig1}
\end{figure*}

The timeline of the initial Covid-19 lockdowns in European countries and in Australia are represented in Figure \ref{fig:fig1}. The heat map graphs for Sweden1, comparing 2019 monthly energy demand to 2020 monthly energy demand in Figure \ref{fig:fig5}, illustrates the choice of it being the control unit. The graphs show minimal colour changes, thus energy demand across the consecutive years in contrast to other countries such as Spain. 

\smallskip

\FloatBarrier
\FloatBarrier

\bigskip
\bigskip

\FloatBarrier
\begin{figure*}[h!]
            \includegraphics[width=.33\textwidth]{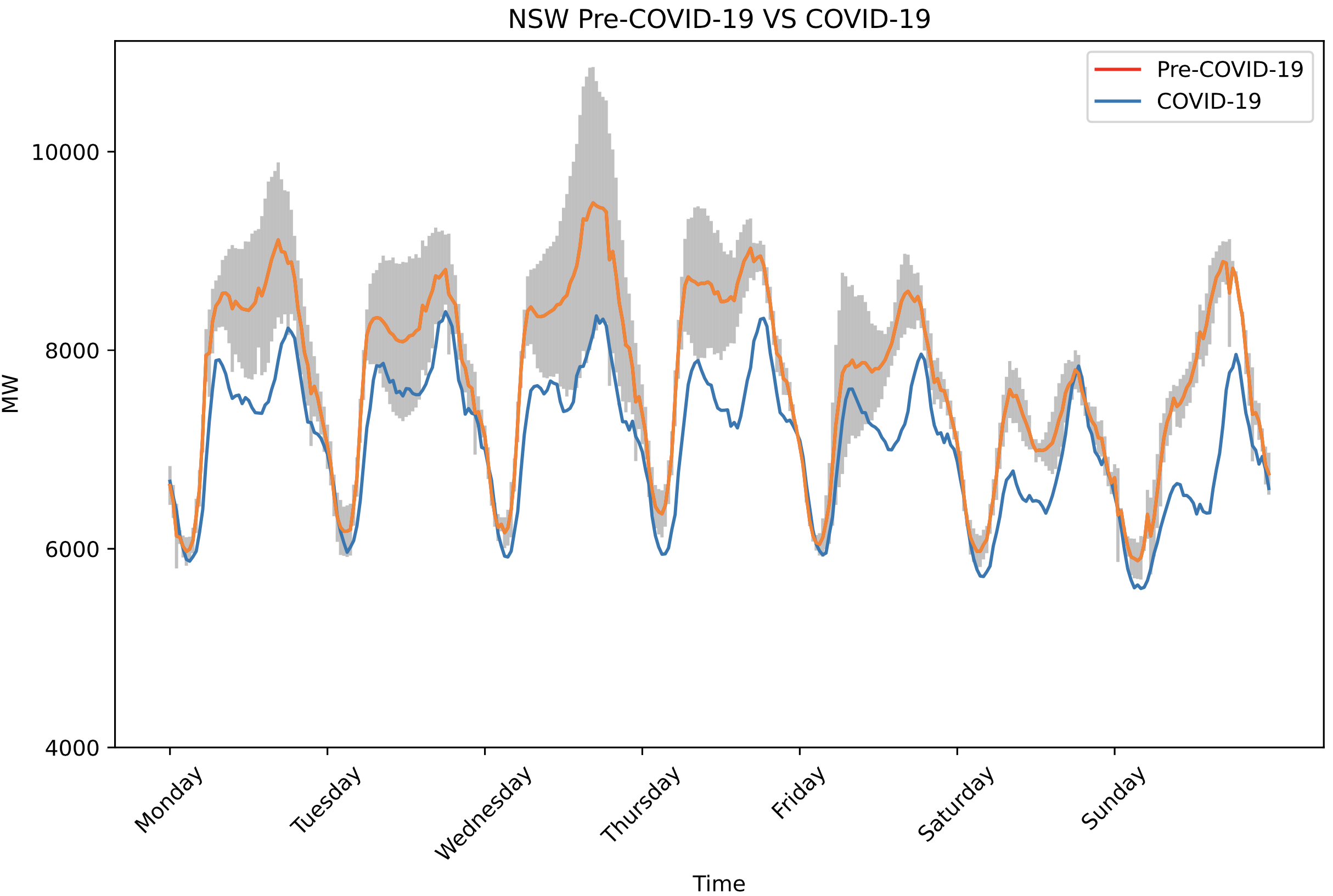}\quad
            \includegraphics[width=.33\textwidth]{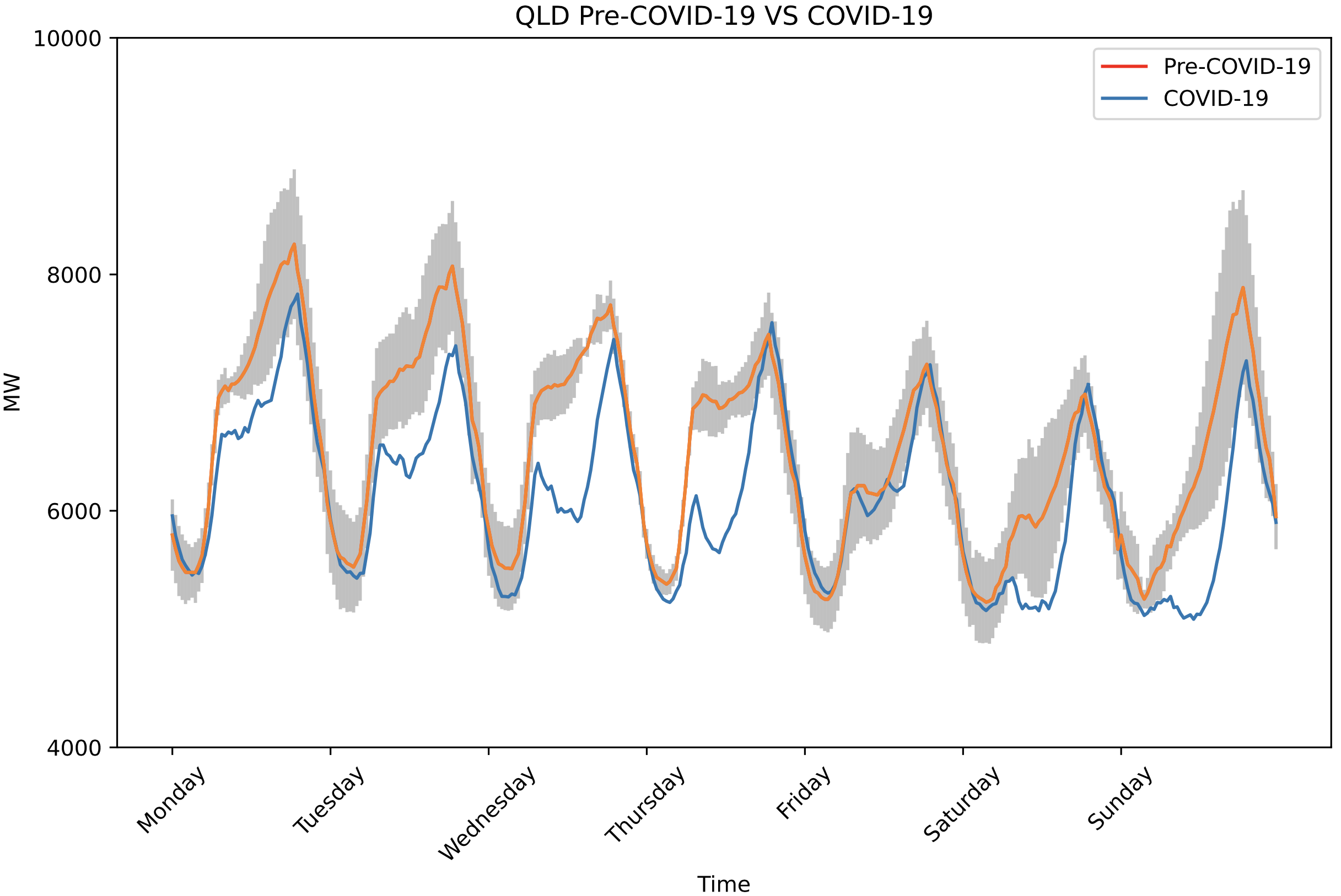}\quad
            \includegraphics[width=.33\textwidth]{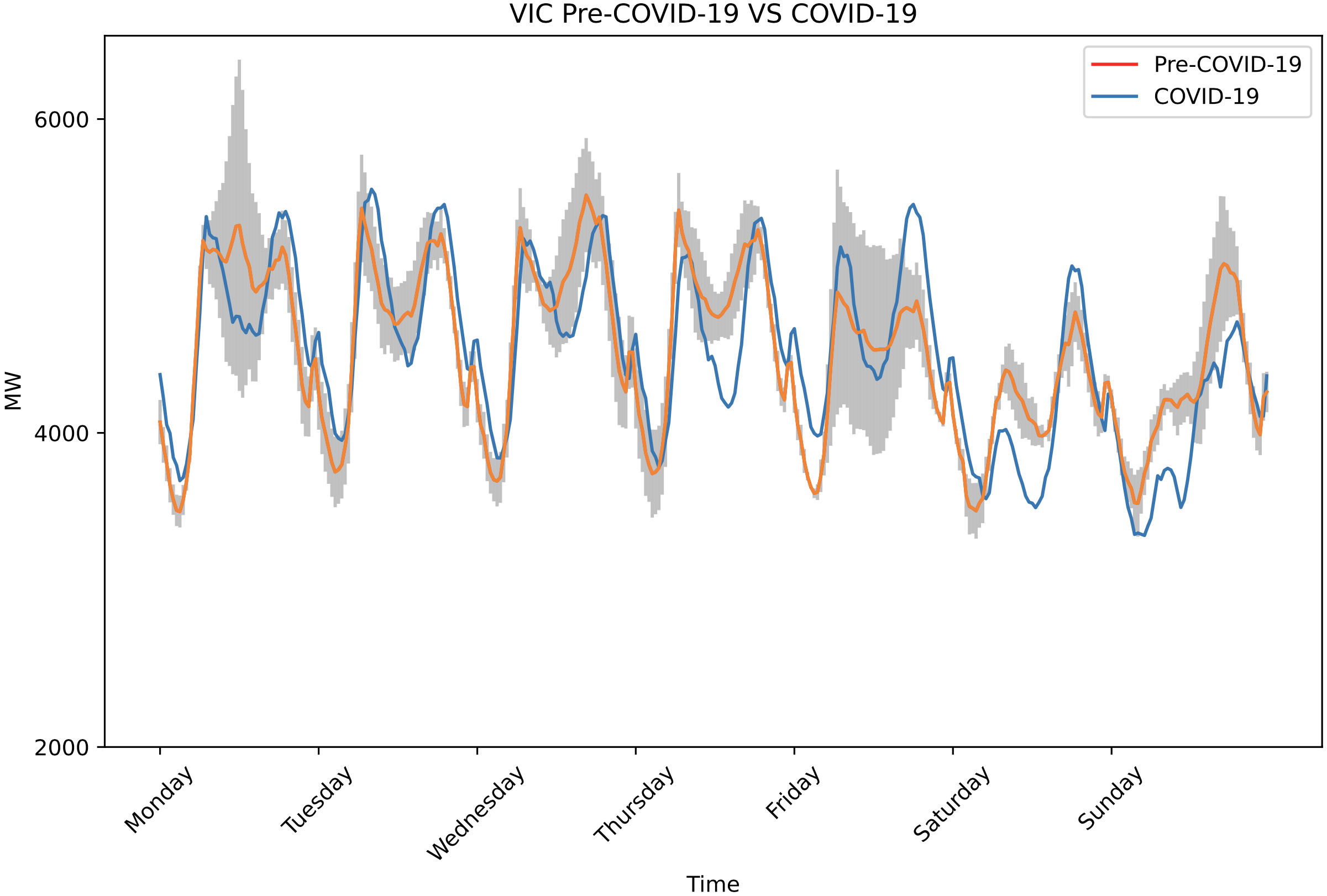}
            \medskip
            \includegraphics[width=.33\textwidth]{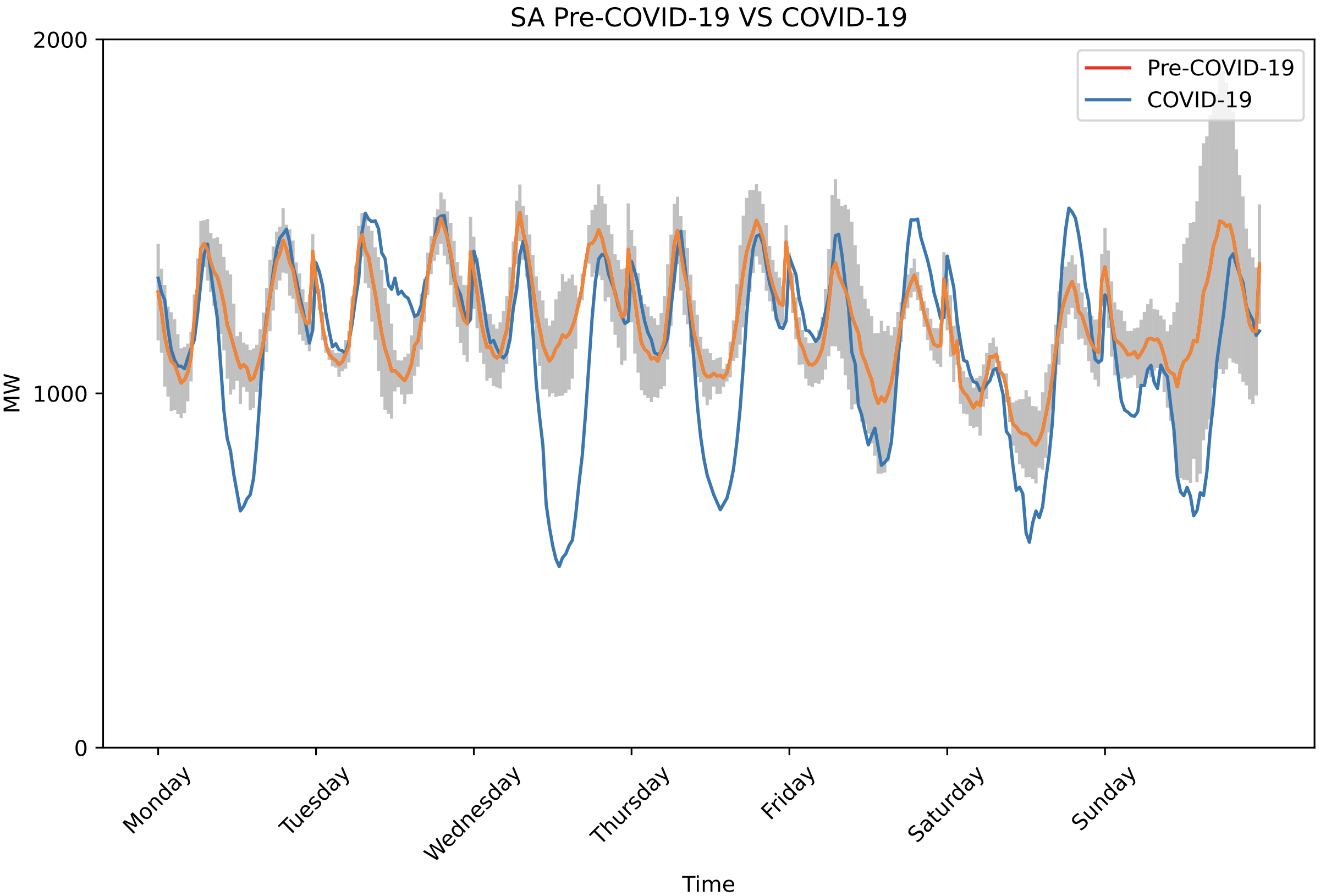}\quad
            \includegraphics[width=.33\textwidth]{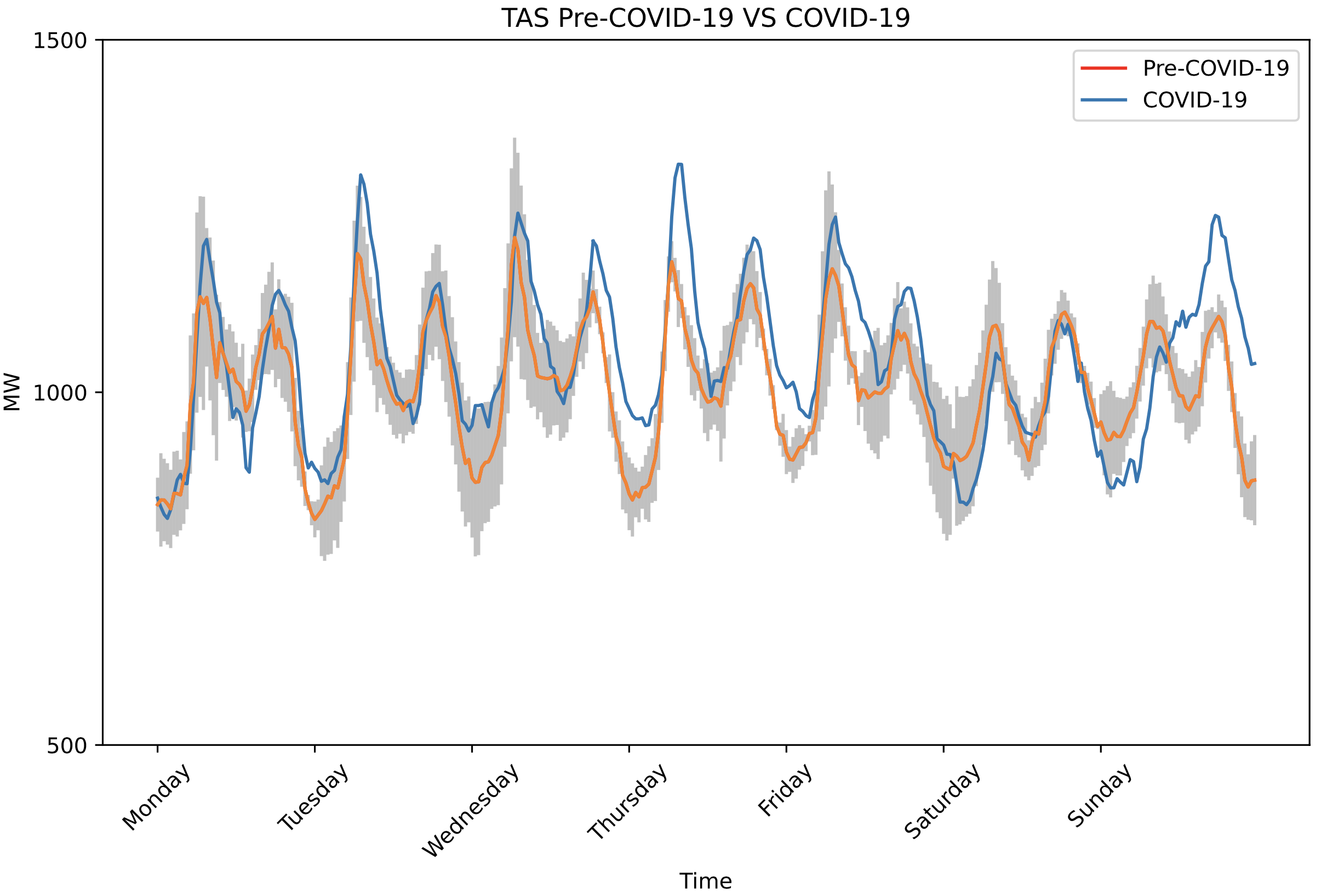}
    \caption{The impact of Covid-19 on average weekly energy demand in the Australian states NSW, QLD, VIC, SA, and TAS.}
\label{fig:fig2}
\end{figure*}
\FloatBarrier

\FloatBarrier
\begin{figure*}[ht!]
            \includegraphics[width=.33\textwidth]{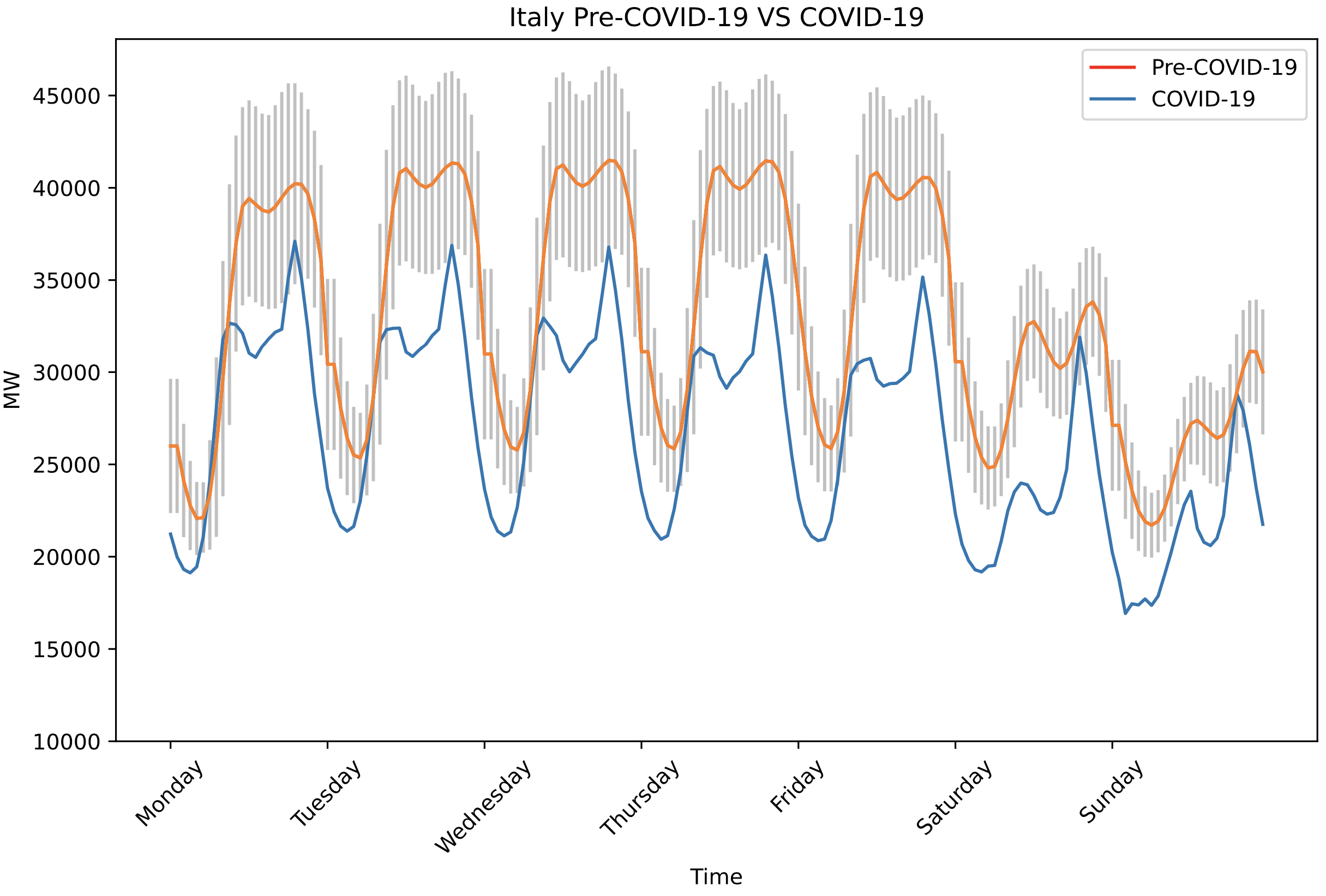}\quad
            \includegraphics[width=.33\textwidth]{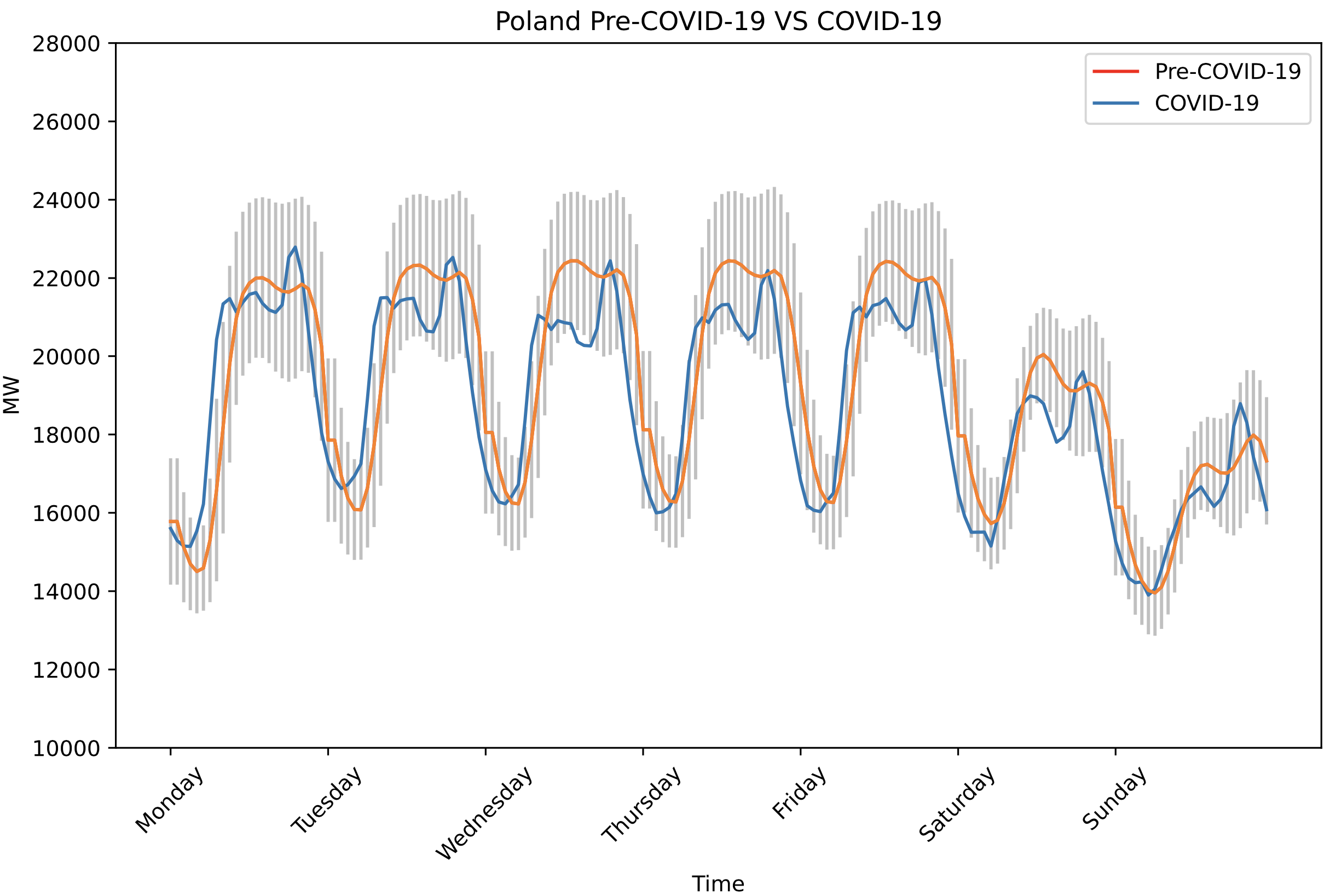}\quad
            \includegraphics[width=.33\textwidth]{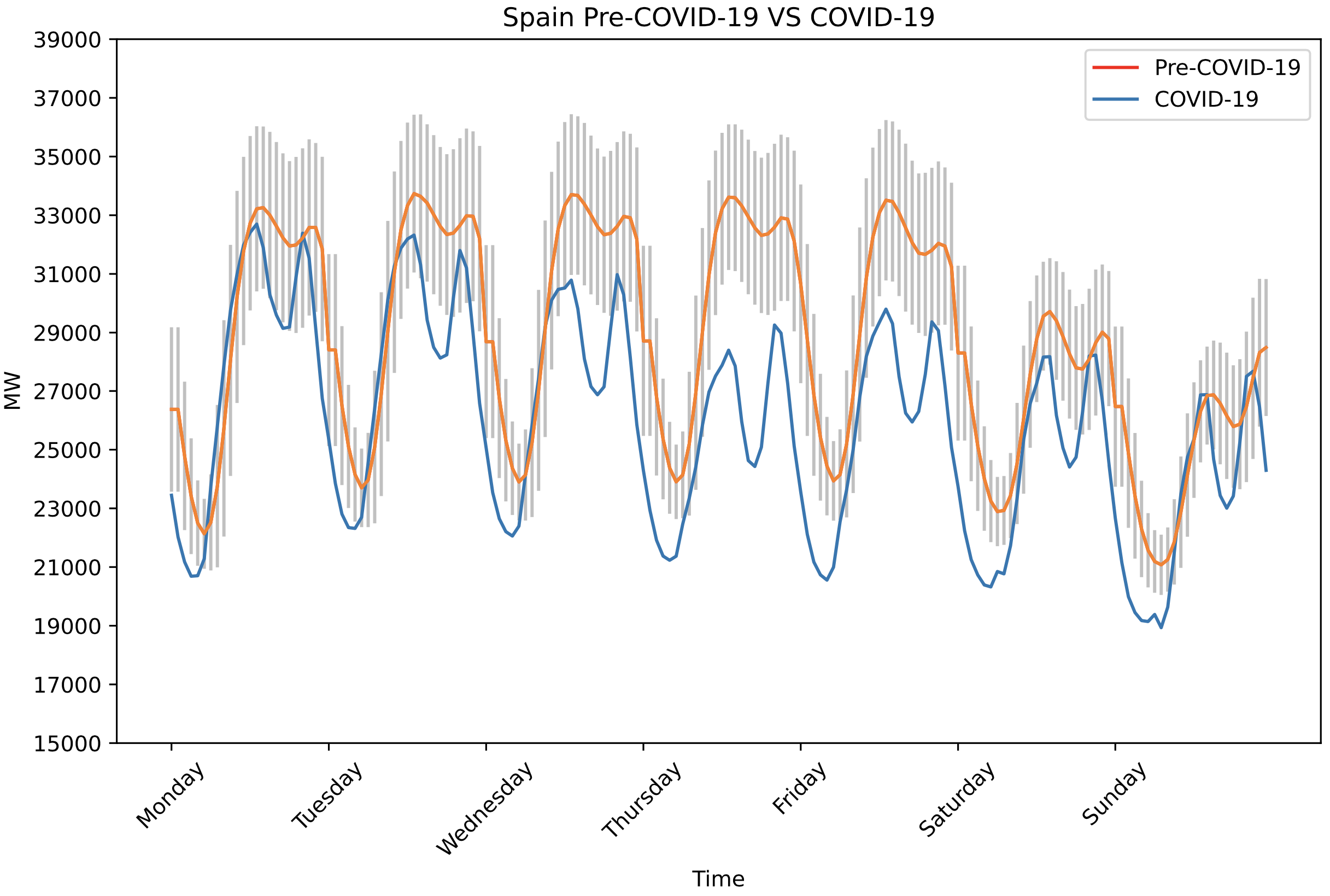}
            \medskip
            \includegraphics[width=.33\textwidth]{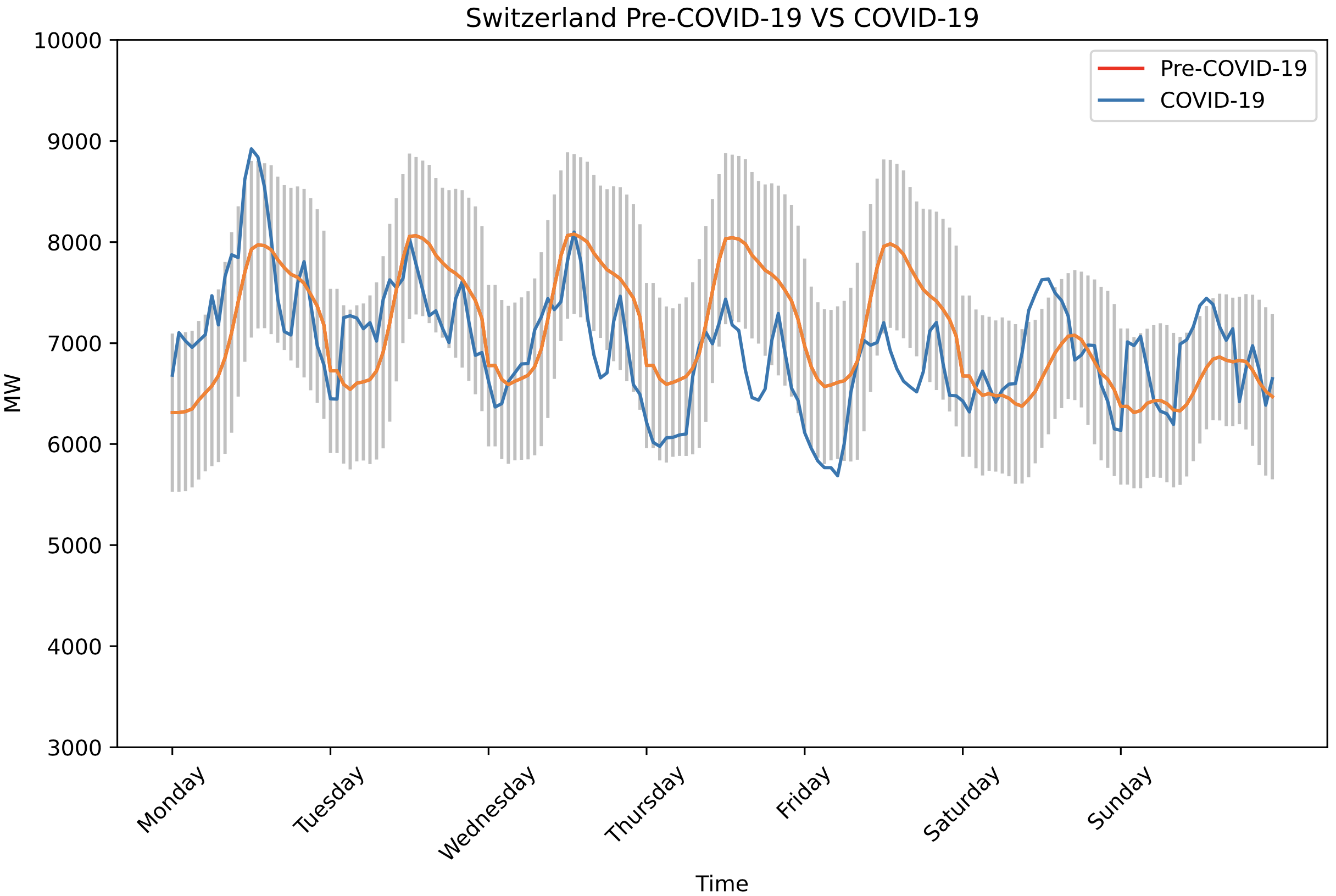}\quad
            \includegraphics[width=.33\textwidth]{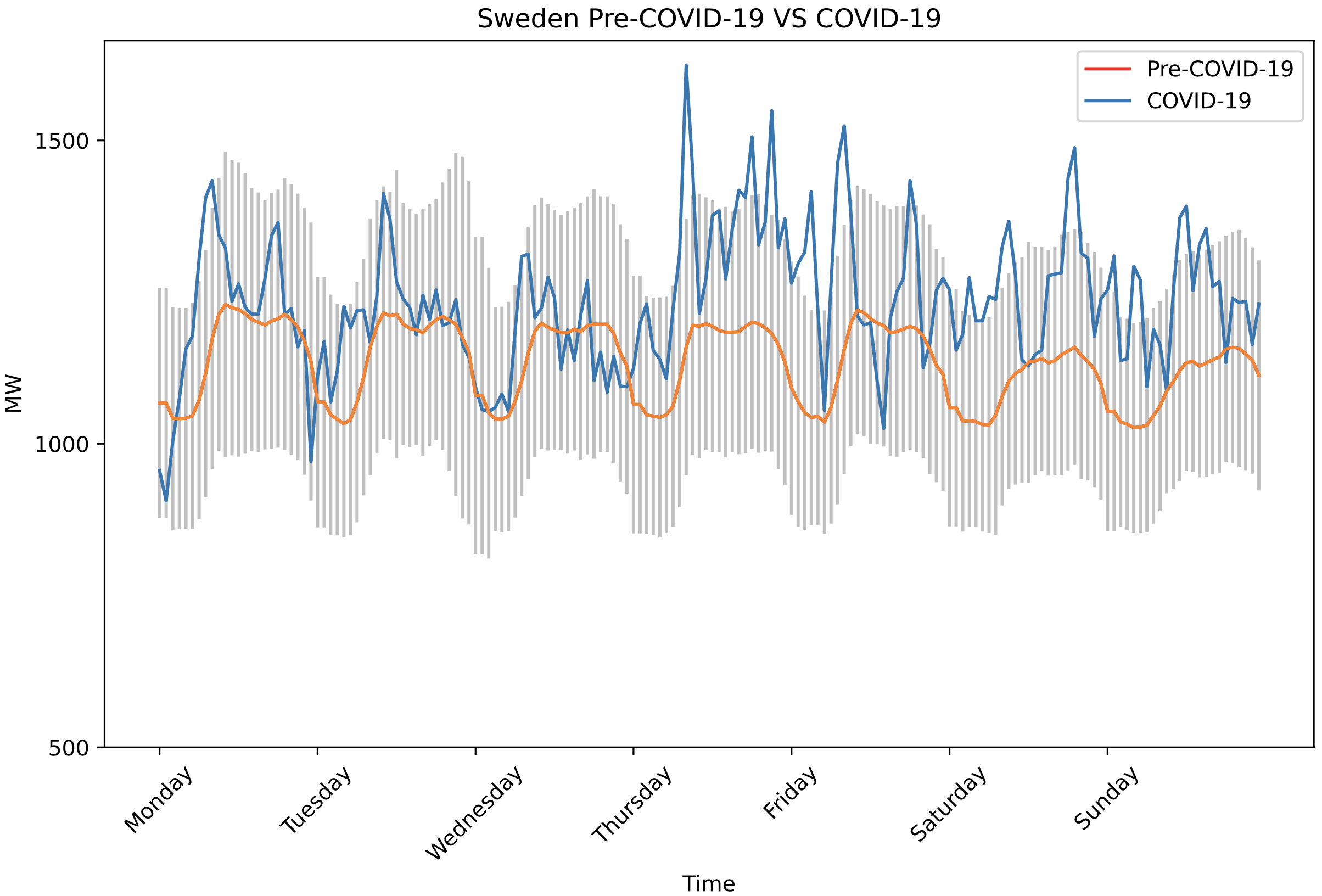}
    \caption{The impact of Covid-19 on average weekly energy demand in Italy, Poland, Spain, Switzerland, and Sweden.}
\label{fig:fig3}
\end{figure*}
\FloatBarrier

The heat map graphs for Spain comparing 2019 monthly energy demand to 2020 monthly energy demand exhibit a significant reduction in energy usage from April to June, where the colour changes from light green and aqua to dark blue, indicative of a reduction of around 10,000MW. It is speculated that the exogenous shock of Covid-19 likely contributed to this decrease. The heat map graphs support the figure in exploratory data analysis.

\FloatBarrier
\begin{figure}[h!]
\centering
\subfloat{
    \includegraphics[width=0.4\textwidth]{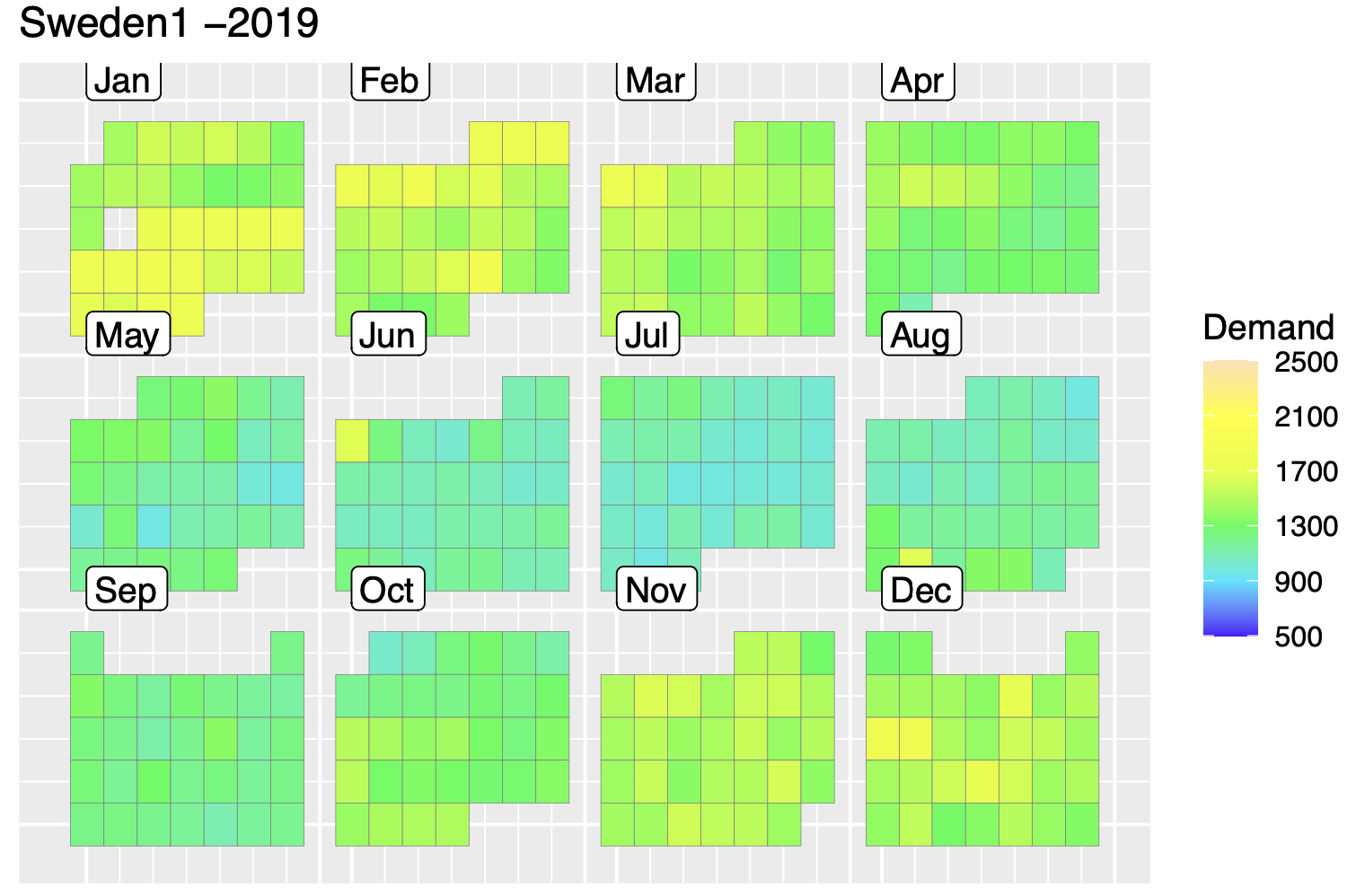}
    \includegraphics[width=0.4\textwidth]{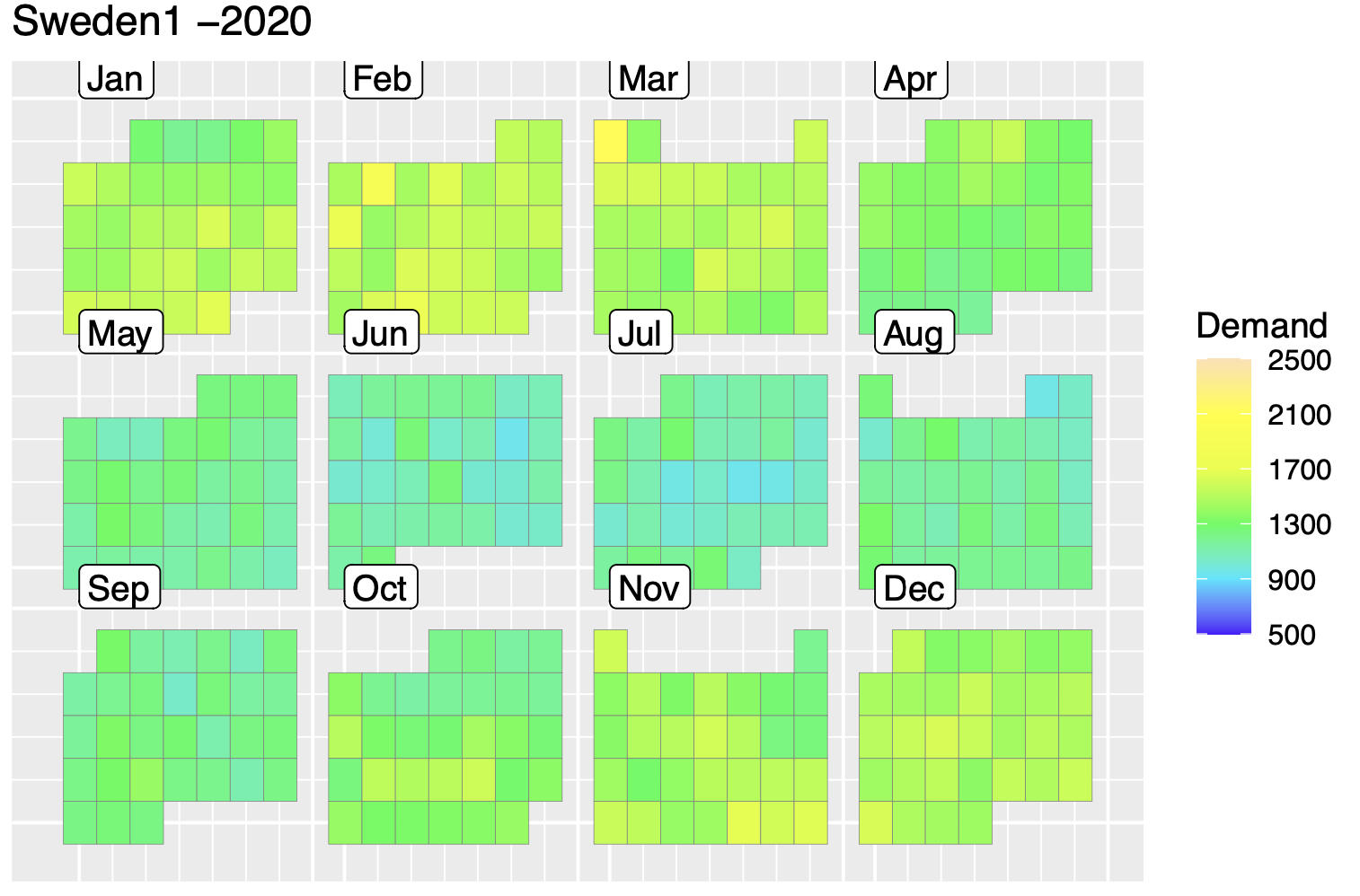}
    }
    \qquad
\subfloat{
    \includegraphics[width=0.4\textwidth]{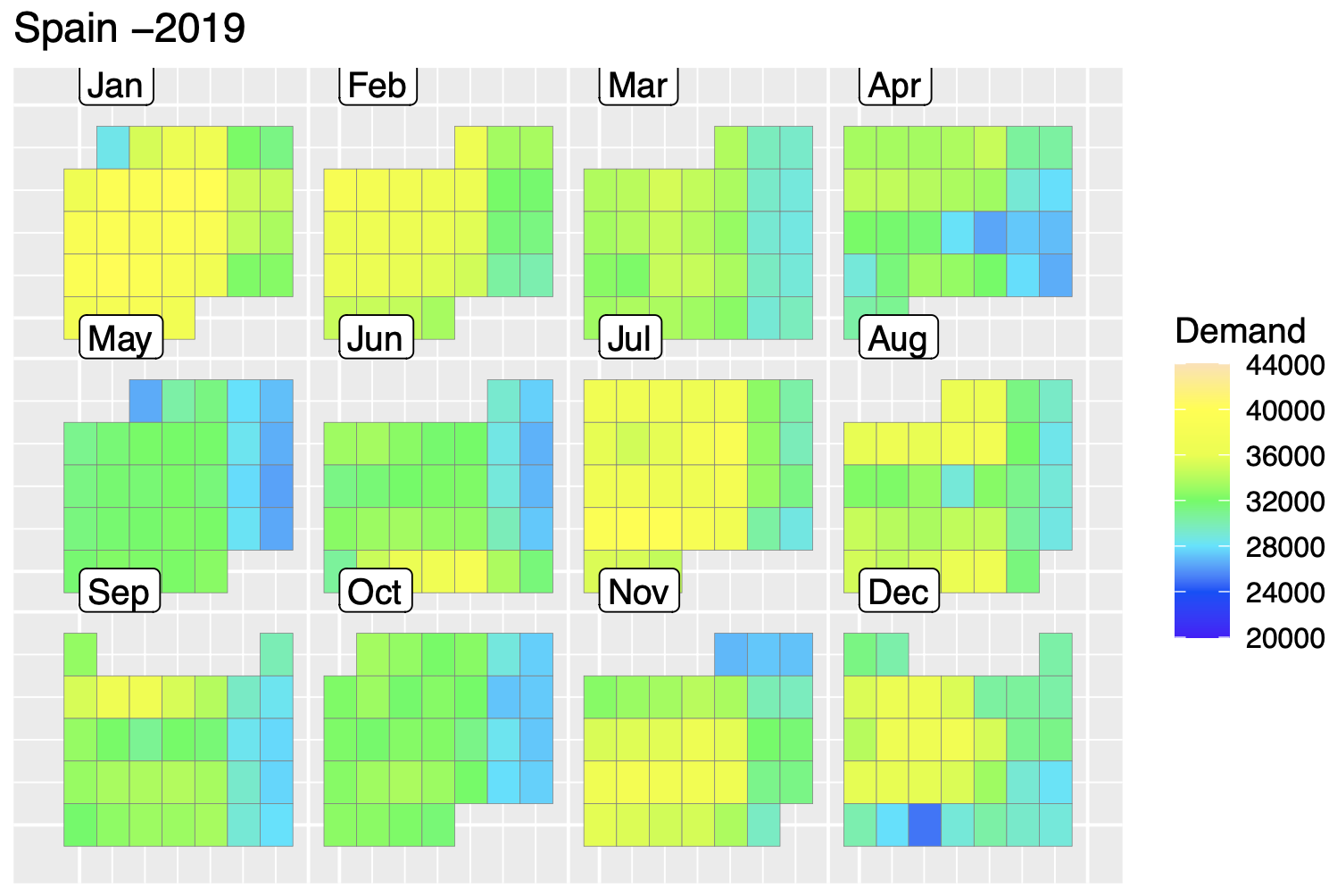}
    \includegraphics[width=0.4\textwidth]{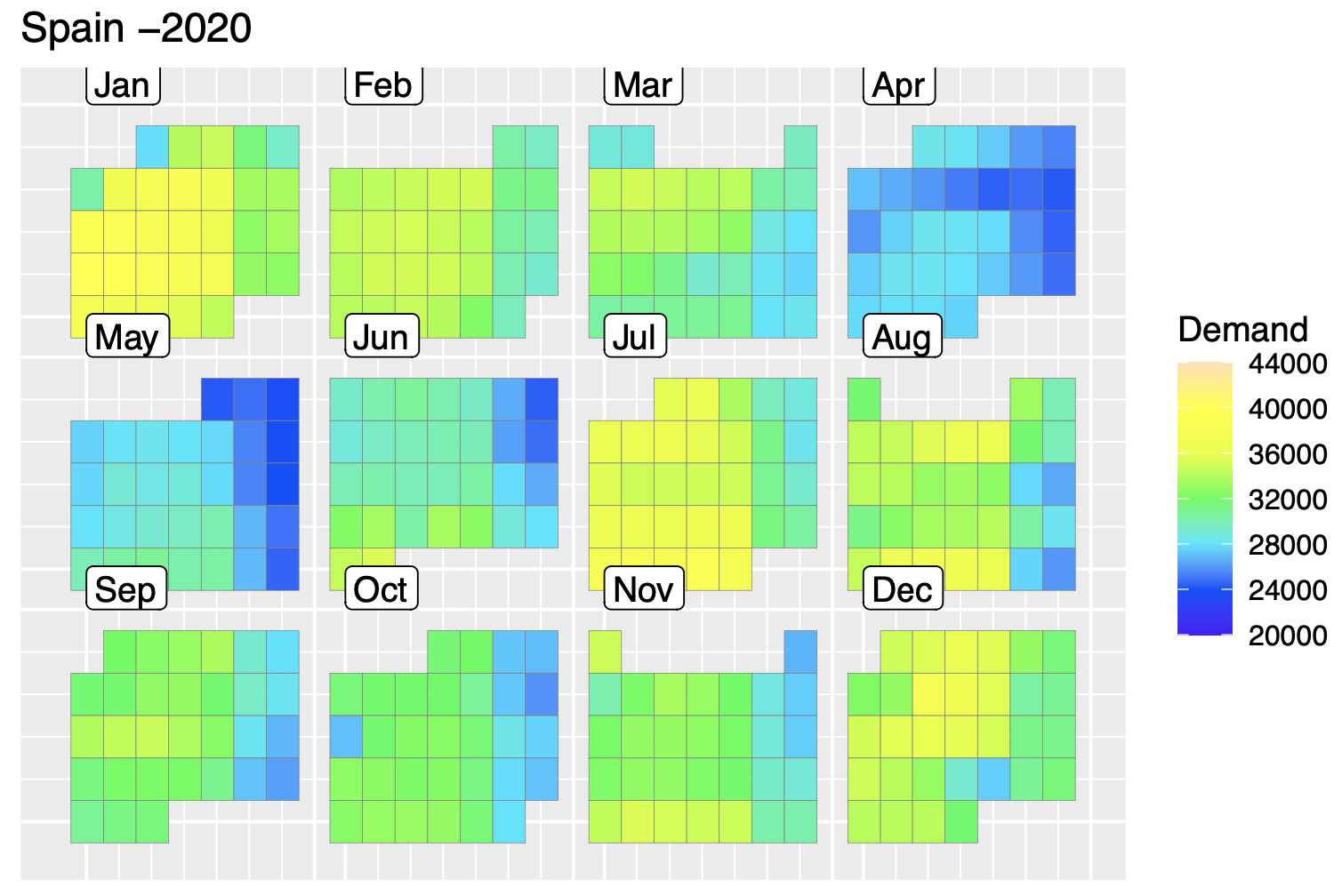}
    }
\caption{Heat map of energy demand in Sweden1 and Spain for 2019 vs 2020}
\label{fig:fig5}
\end{figure}
\FloatBarrier

\newpage

\section{Additional Figures and Tables of Results Analysis (Section V)}
\label{figs2}

\FloatBarrier
\begin{table*}[!ht]
\centering
\caption{ Average Treatment Effect in energy consumption  (absolute and percentage values) in European data using the DeepAR model.} 
\centering \label{table:Table II}
\begin{tabularx}{\textwidth}{L|RR|RR|RR|RR|RR}
\hline
\multirow{4}{*}{\textbf{Quantiles}}      &            \multicolumn{2}{>{\bf}c|}{Italy} & \multicolumn{2}{>{\bf}c|}{Poland} & \multicolumn{2}{>{\bf}c|}{Spain} & \multicolumn{2}{>{\bf}c|}{Switzerland}  & \multicolumn{2}{>{\bf}c}{Sweden1}\\
    \cline{2-11}
      & Avg. Treat. Effect(in Mwh)& Change in Demand (\%) & Avg. Treat. Effect(in Mwh)& Change in Demand (\%) & Avg. Treat. Effect(in Mwh)& Change in Demand (\%) & Avg. Treat. Effect(in Mwh)& Change in Demand (\%) & Avg. Treat. Effect(in Mwh)& Change in Demand (\%) \\
    \hline
0.1&-12522.03&-65.84&-4574.77&-29.90&-7570.25&-36.59&-1273.93&-19.97&-145.62&-13.40\\
0.25&-11089.92&-52.31&-3993.76&-24.38&-6449.63&-28.72&-1218.97&-18.27&-148.15&-12.98\\
0.3&-10738.12&-49.43&-3860.16&-23.22&-5908.88&-25.52&-1206.36&-17.90&-148.51&-12.86\\
0.4&-9177.60&-38.92&-3104.13&-17.65&-4902.37&-20.06&-1079.43&-15.50&-144.71&-12.25\\
0.5&-6906.85&-26.43&-1985.74&-10.51&-3861.84&-15.01&-1087.14&-15.42&-136.31&-11.26\\
0.6&-5553.66&-20.03&-1134.97&-5.70&-3077.51&-11.51&-1022.09&-14.21&-133.74&-10.85\\
0.75&-4166.02&-14.08&-758.78&-3.68&-2059.89&-7.31&-926.34&-12.44&-122.12&-9.54\\
0.8&-3567.10&-11.74&-709.43&-3.41&-1680.12&-5.85&-934.04&-12.46&-104.76&-7.98\\
0.9&-2600.64&-8.15&-546.67&-2.56&-930.79&-3.11&-855&-11.03&-68.09&-4.89\\
0.95&-2225.37&-6.79&-550.12&-2.54&-727.32&-2.38&-762.67&-9.53&-47.13&-3.24\\
0.99&-719.58&-2.03&-580.75&-2.60&-70.58&-0.22&-257.90&-2.91&282.12&15.07\\
\hline
\textbf{\emph{p-values}}&\multicolumn{2}{>{\em}c|}{4.7e-07 / 4.7e-07}&\multicolumn{2}{>{\em}c|}{4.7e-07 / 4.7e-07}&\multicolumn{2}{>{\em}c|}{4.7e-07 / 4.7e-07}&\multicolumn{2}{>{\em}c|}{4.7e-07 / 4.7e-07}&\multicolumn{2}{>{\em}c}{0.005}\\
\hline
\end{tabularx}
\label{tab:table2}
\end{table*}
\FloatBarrier

\FloatBarrier
\begin{table*}[!ht]
\centering
\caption{Average Treatment Effect in energy consumption (absolute and percentage values) for the Australian dataset using the DeepAR model.}
\label{table:x}
\centering
\begin{tabularx}{\linewidth}{L|RR|RR|RR|RR|RR}
\hline
\multirow{4}{*}{\textbf{Quantiles}}      &            \multicolumn{2}{>{\bf}c|}{New South Wales} & \multicolumn{2}{>{\bf}c|}{Queensland} & \multicolumn{2}{>{\bf}c|}{Victoria} & \multicolumn{2}{>{\bf}c|}{South Australia}  & \multicolumn{2}{>{\bf}c}{Tasmania}\\
    \cline{2-11}
      & Avg. Treat. Effect(in Mwh)& Change in Demand (\%) & Avg. Treat. Effect(in Mwh)& Change in Demand (\%) & Avg. Treat. Effect(in Mwh)& Change in Demand (\%) & Avg. Treat. Effect(in Mwh)& Change in Demand (\%) & Avg. Treat. Effect(in Mwh)& Change in Demand (\%) \\
    \hline
0.1&-804.57&-13.35&-749.85&-14.30&-348.77&-9.21&-320.17&-42.63&-46.57&-5.12\\
0.25&-569.54&-8.71&-732.57&-13.30&-188.36&-4.57&-136.65&-13.87&-20.09&-2.05\\
0.3&-480.59&-7.18&-662.14&-11.75&-100.36&-2.36&-78.18&-7.41&-16.31&-1.64\\
0.4&-246.17&-3.50&-495.75&-8.40&-36.17&-0.82&-21.69&-1.92&-8.52&-0.83\\
0.5&-156.86&-2.17&-406.43&-6.67&3.84&0.09&24.41&2.04&8.12&0.77\\
0.6&-106.60&-1.44&-306.64&-4.89&79.60&1.72&61.37&4.91&26.76&2.45\\
0.75&-56.36&-0.78&-85.50&-1.28&256.59&5.22&112.10&8.42&45.18&3.97\\
0.8&-48.95&-0.64&19.74&0.29&328.63&6.53&136.33&9.96&50.81&4.40\\
0.9&-35.91&-0.45&204.19&2.85&407.01&7.79&174.50&12.11&76.28&6.31\\
0.95&82.41&1.01&314.47&4.23&442.51&8.25&191.02&12.85&72.38&5.88\\
0.99&-111.19&-1.33&280.7&3.64&332.81&6.10&171.77&11.27&84.36&6.54\\
\hline
\textbf{\emph{p-values}}&\multicolumn{2}{>{\em}c|}{6.7e-06 / 0.00025}&\multicolumn{2}{>{\em}c|}{0.001 / 0.00025}&\multicolumn{2}{>{\em}c|}{0.87 / 0.66}&\multicolumn{2}{>{\em}c|}{0.87 / 0.78}&\multicolumn{2}{>{\em}c}{0.06}\\
\hline
\end{tabularx}
\label{tab:table4}
\end{table*}
\FloatBarrier

\FloatBarrier
\begin{figure*}[h!]
    \centering
    \subfloat{
        {\includegraphics[width=0.35\textwidth,height = 4cm]{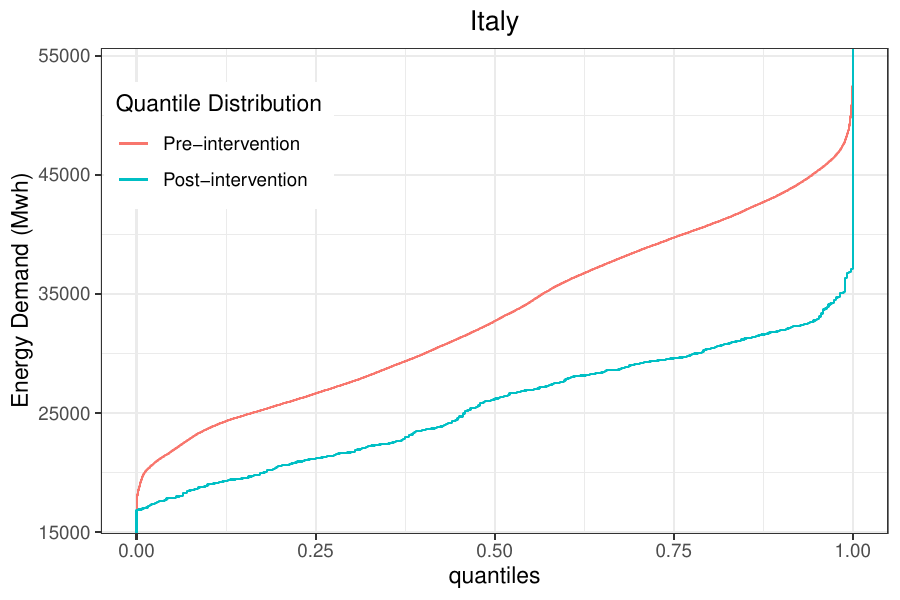}}
        {\includegraphics[width=0.35\textwidth,height = 4cm]{CovidImpact_EnergyDemandStudy_clean/DeepAR_Italy.png}}
        {\includegraphics[width=0.35\textwidth,height = 4cm]{CovidImpact_EnergyDemandStudy_clean/DeepAR_new_QD_Italy.png}}
       
        }
        \qquad
    \subfloat{
        {\includegraphics[width=0.35\textwidth,height = 4cm]{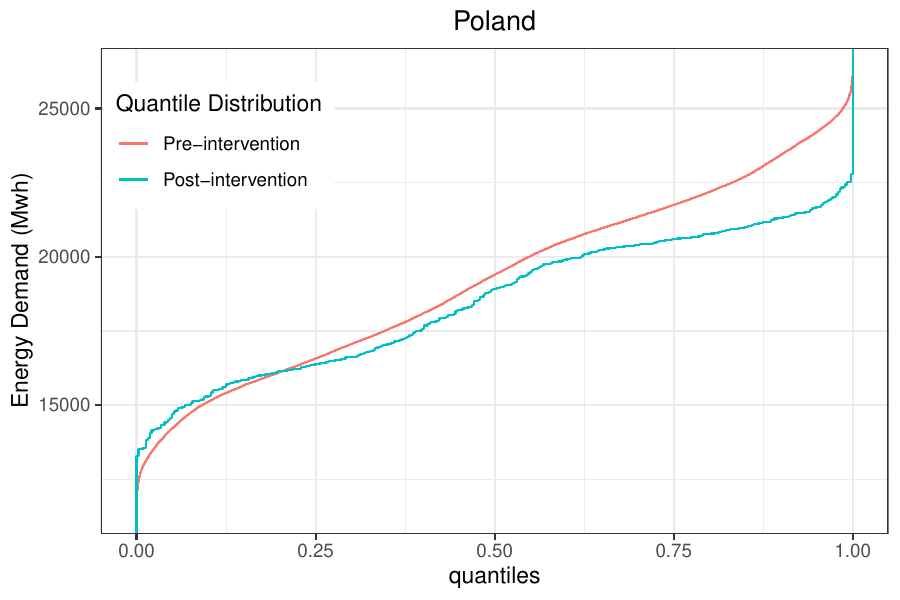}}
        {\includegraphics[width=0.35\textwidth,height = 4cm]{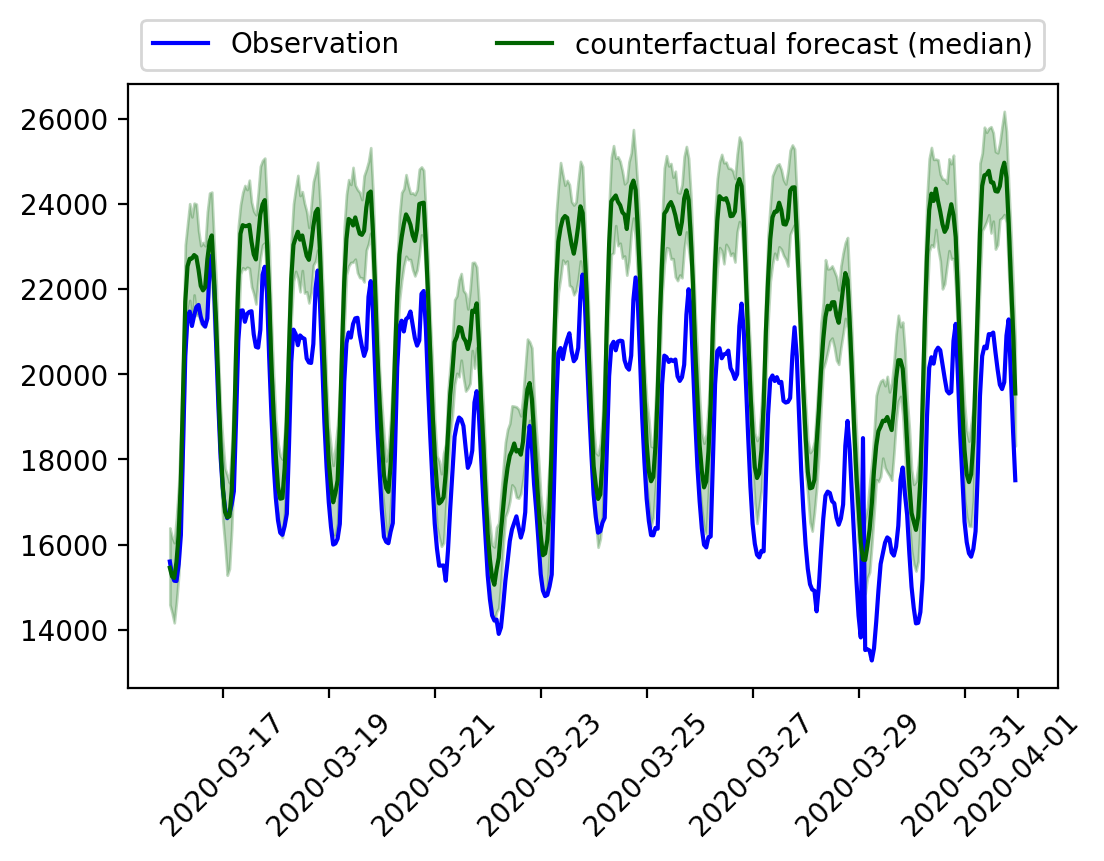}}
        {\includegraphics[width=0.35\textwidth,height = 4cm]{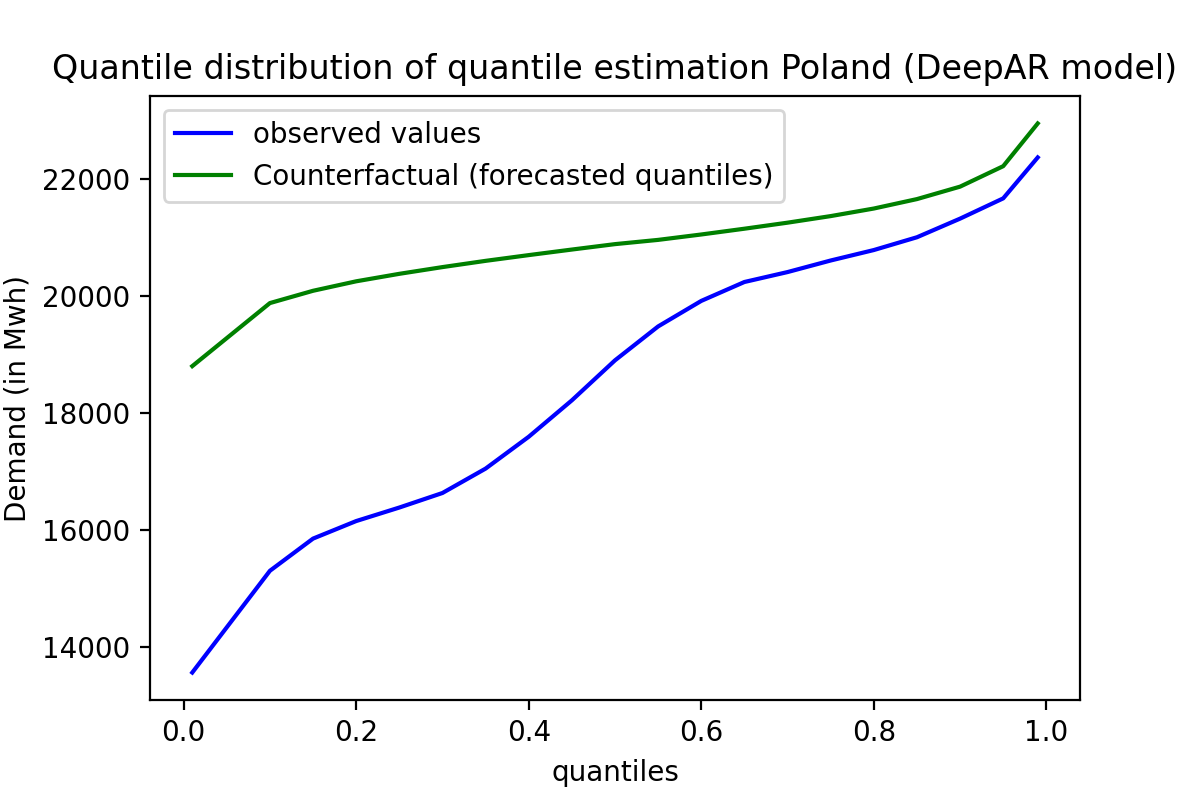}}
        
        }
    \qquad
    \subfloat{
        {\includegraphics[width=0.35\textwidth,height = 4cm]{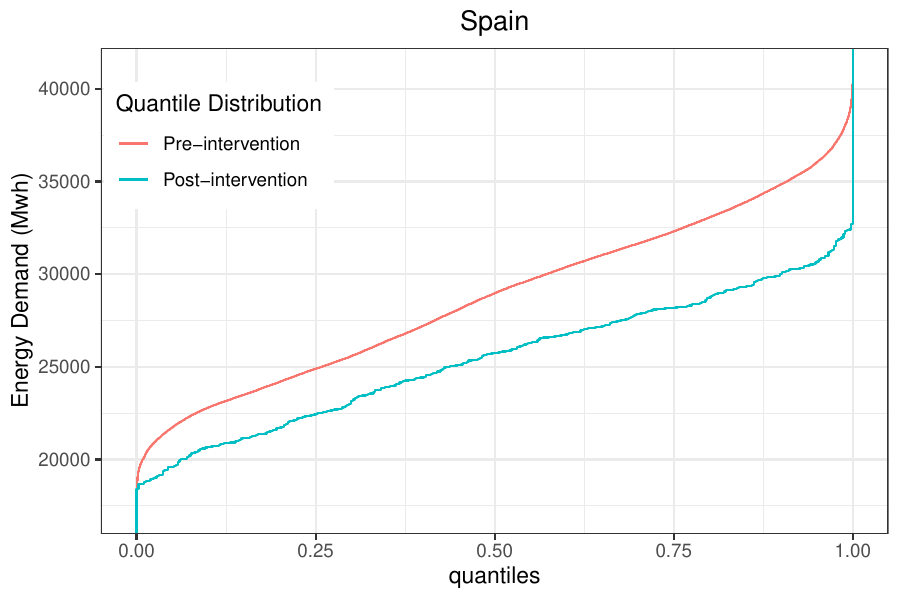}}
        {\includegraphics[width=0.35\textwidth,height = 4cm]{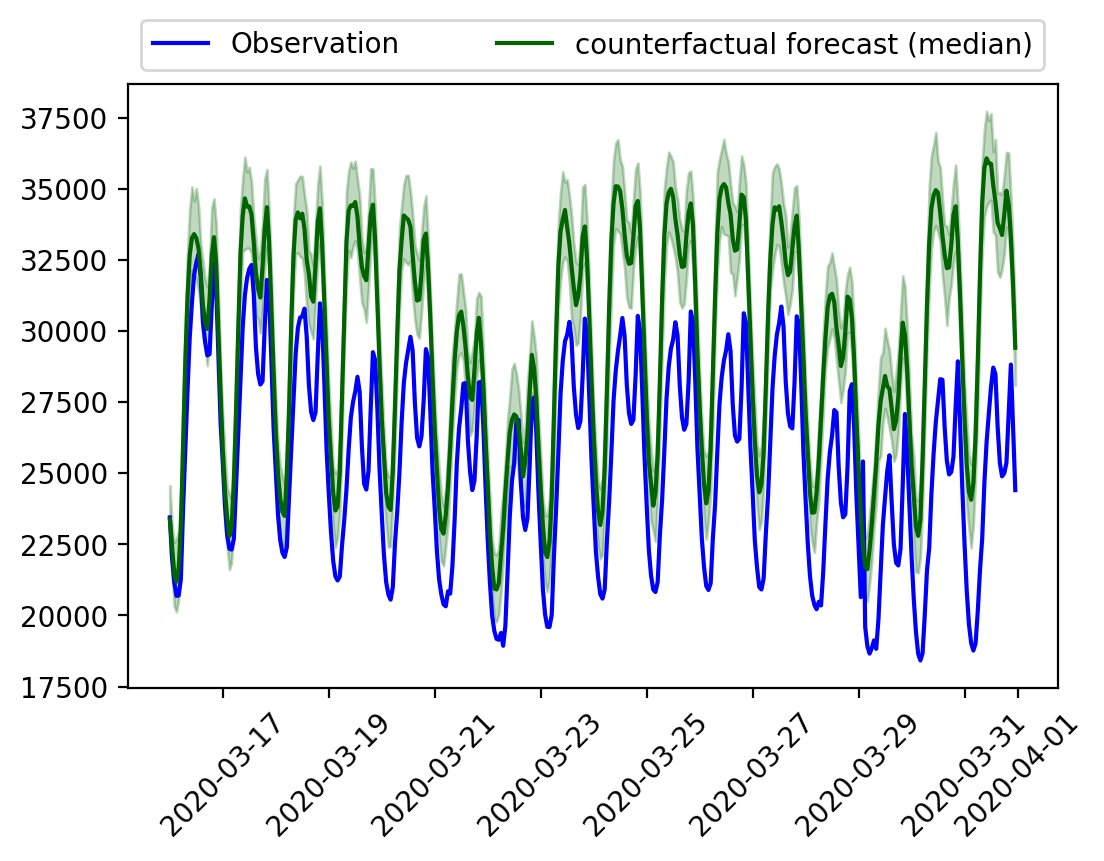}}
        {\includegraphics[width=0.35\textwidth,height = 4cm]{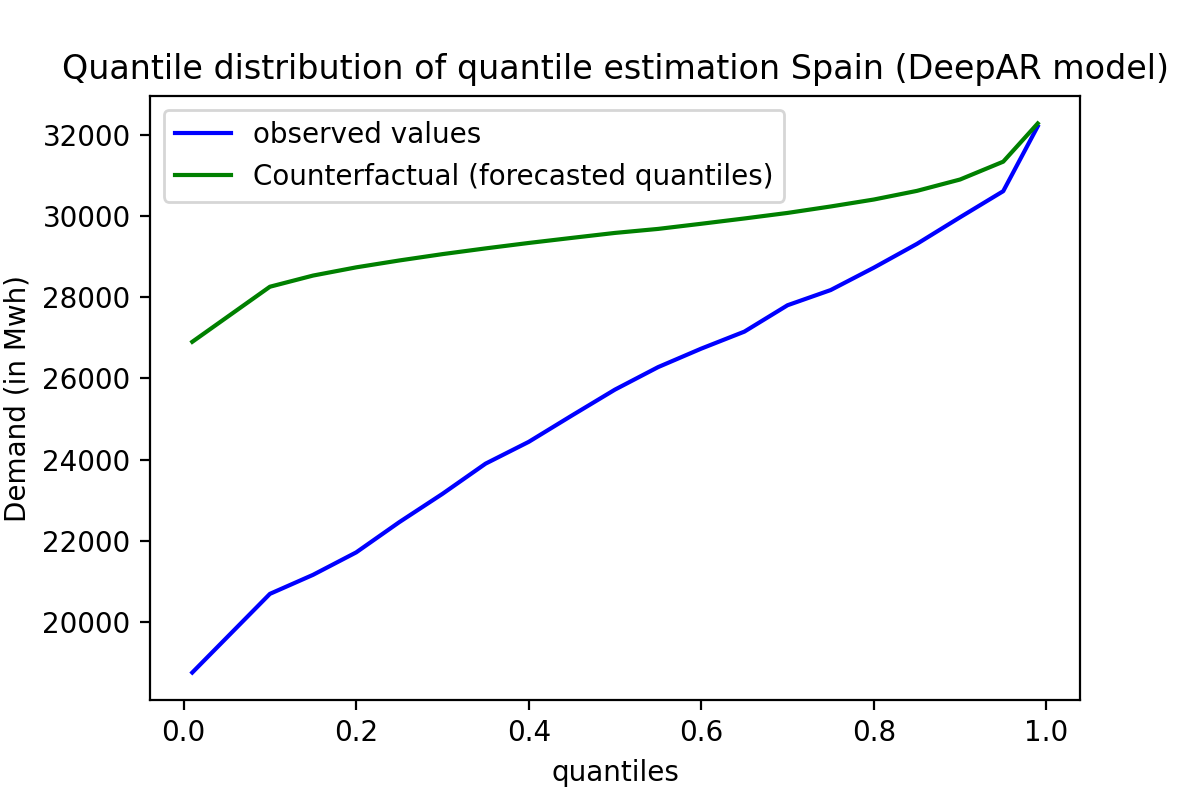}}
        
        }
    \qquad
    \subfloat{
        {\includegraphics[width=0.35\textwidth,height = 4cm]{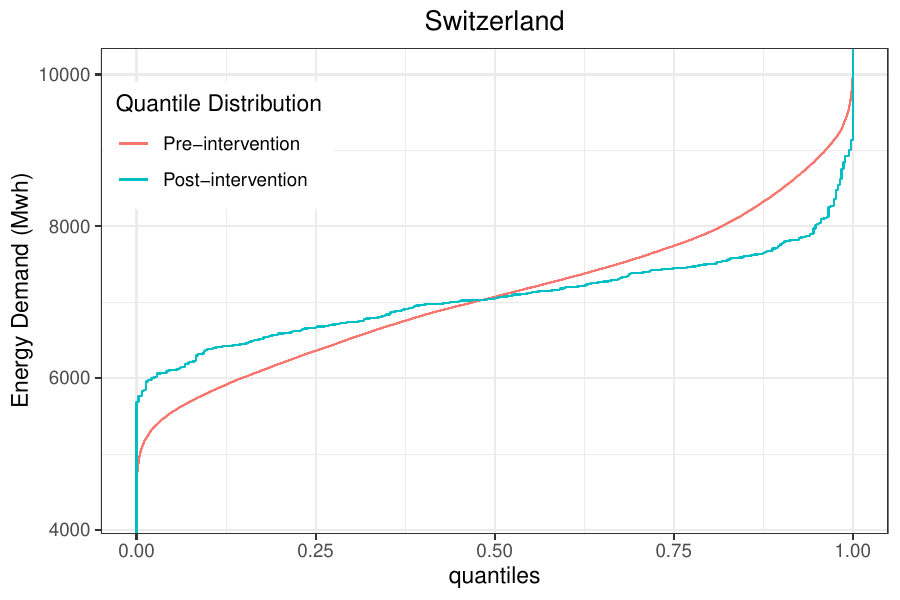}}
        {\includegraphics[width=0.35\textwidth,height = 4cm]{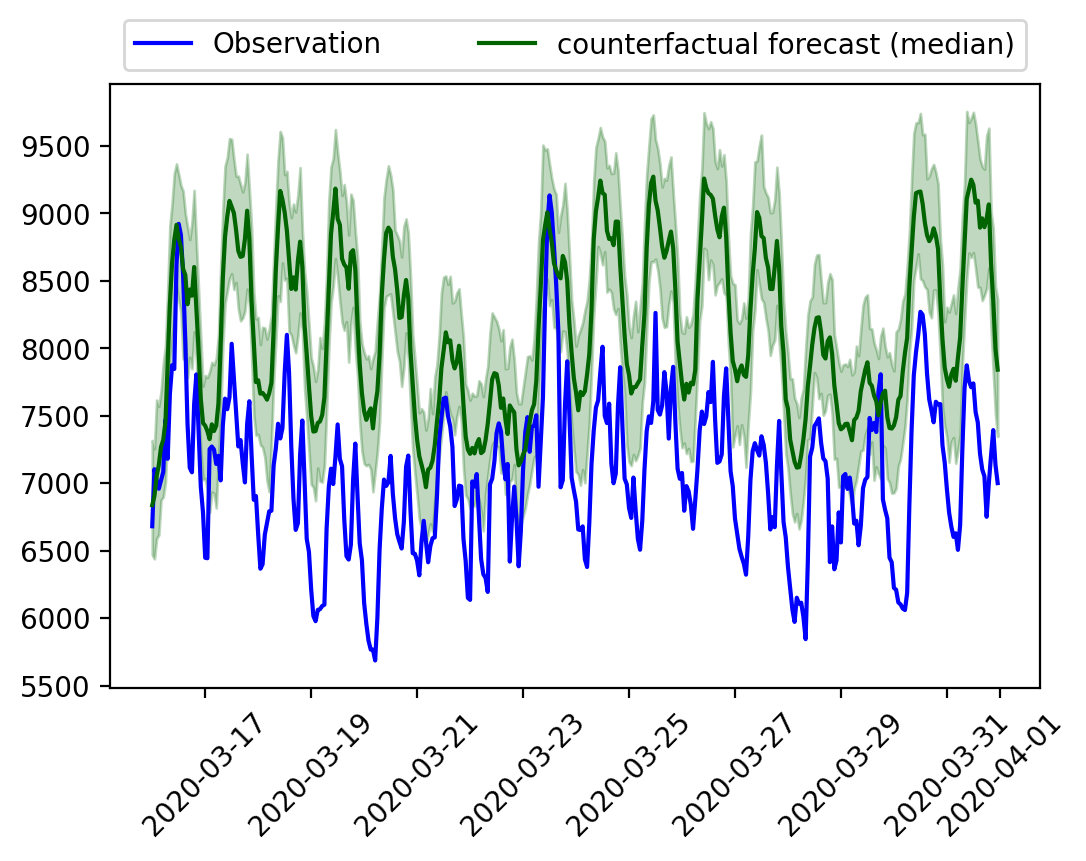}}
        {\includegraphics[width=0.35\textwidth,height = 4cm]{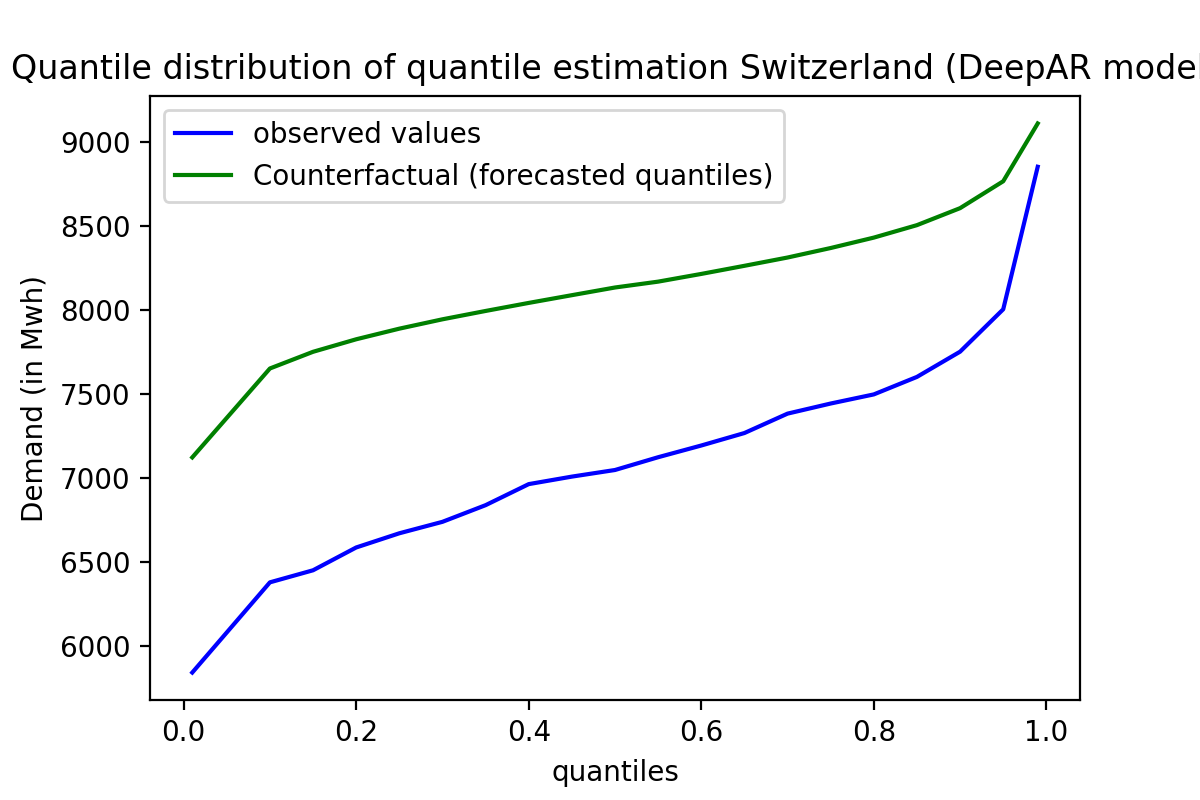}}
      
        }
    \qquad
    \subfloat{
        {\includegraphics[width=0.35\textwidth,height = 4cm]{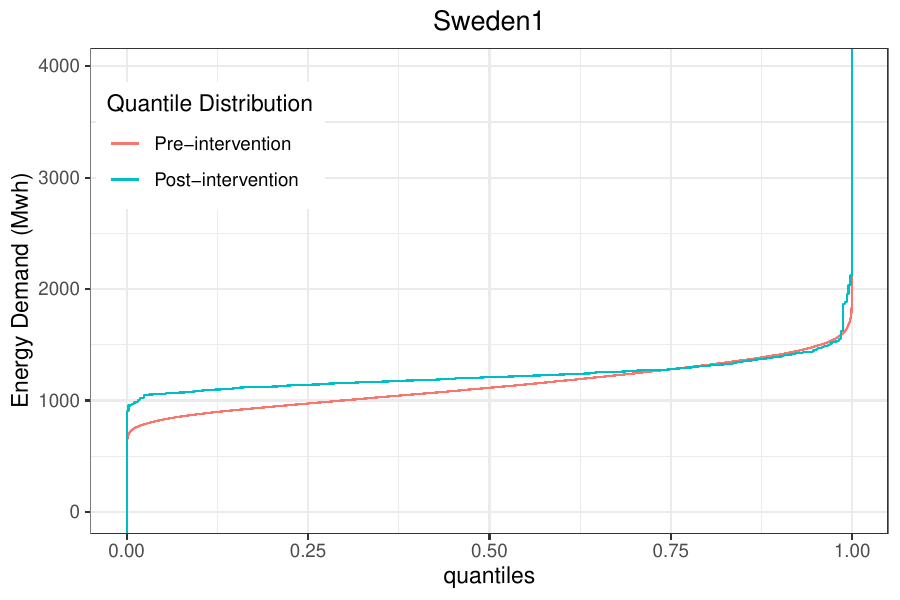}}
        {\includegraphics[width=0.35\textwidth,height = 4cm]{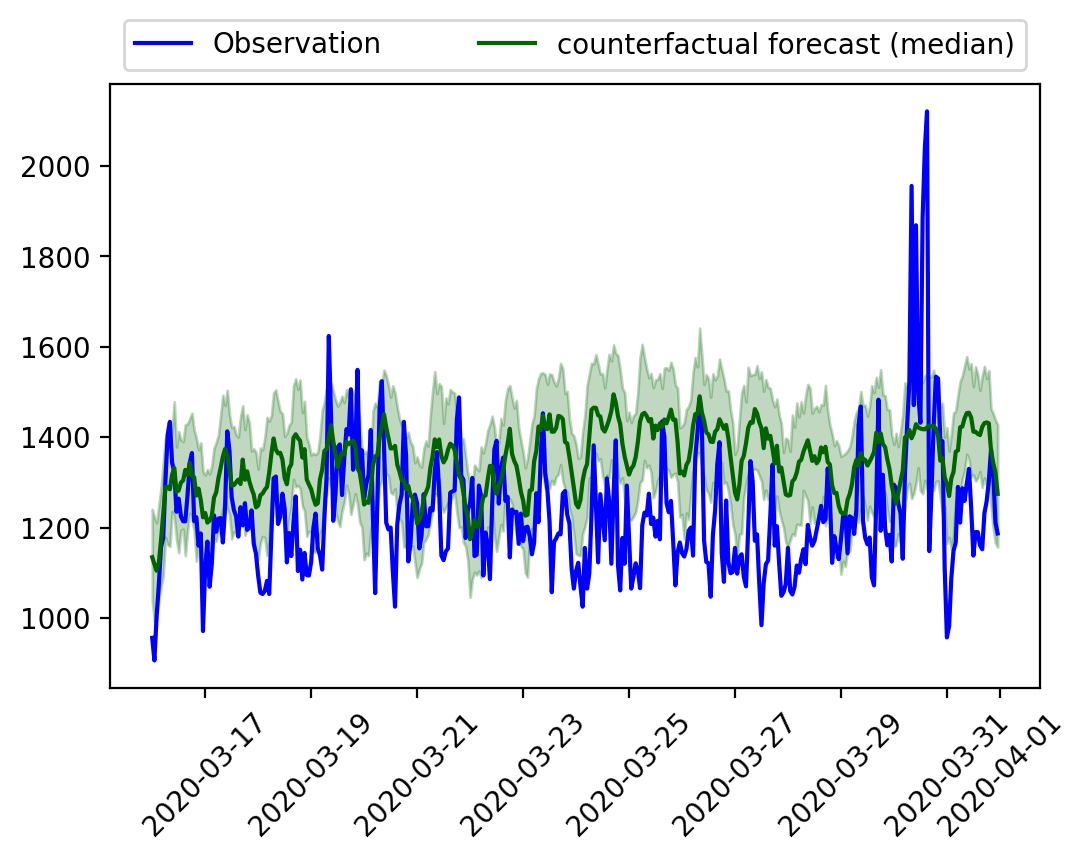}}
        {\includegraphics[width=0.35\textwidth,height = 4cm]{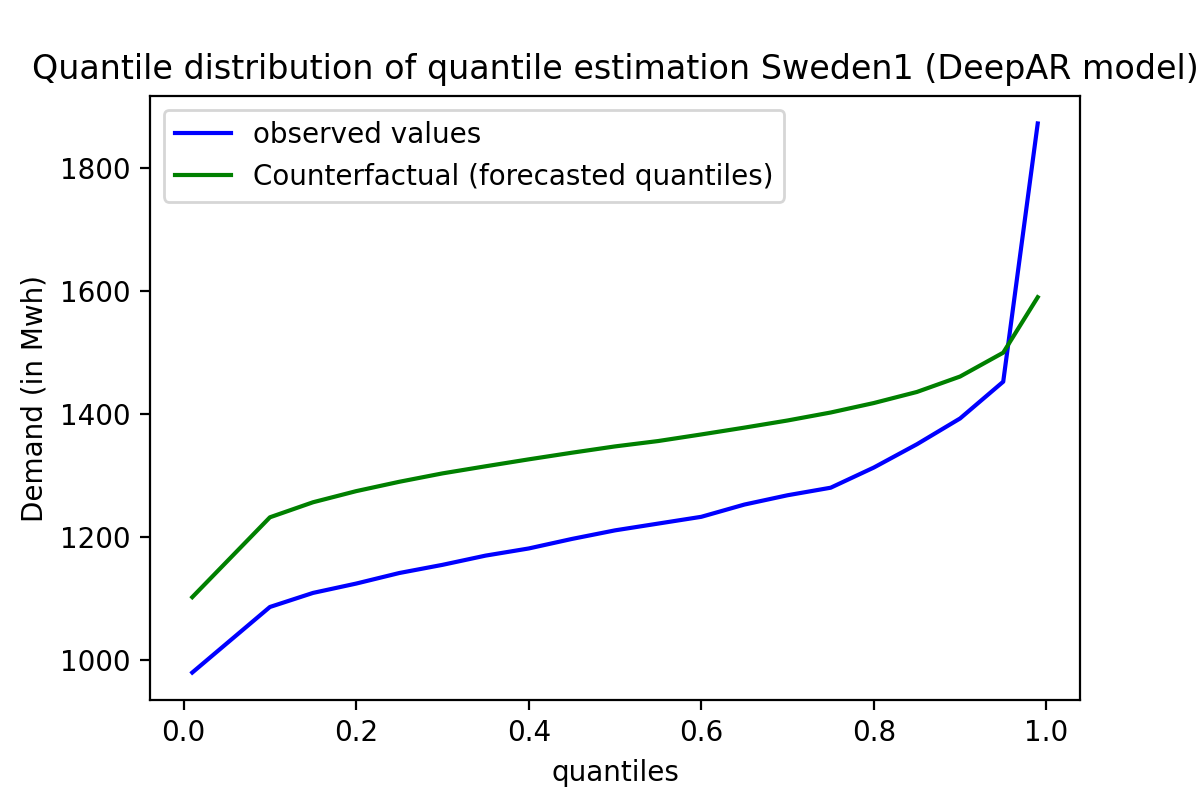}}
        }
    \caption{European data -- Observed quantile distribution of pre- and post-intervention data (left), DeepAR model's point forecast (middle) and its forecast quantile distribution (right).\label{label1}}

\end{figure*}
\FloatBarrier

\FloatBarrier
\begin{figure*}[h!]
   \centering
    \subfloat{
        {\includegraphics[width=0.35\textwidth,height = 4cm]{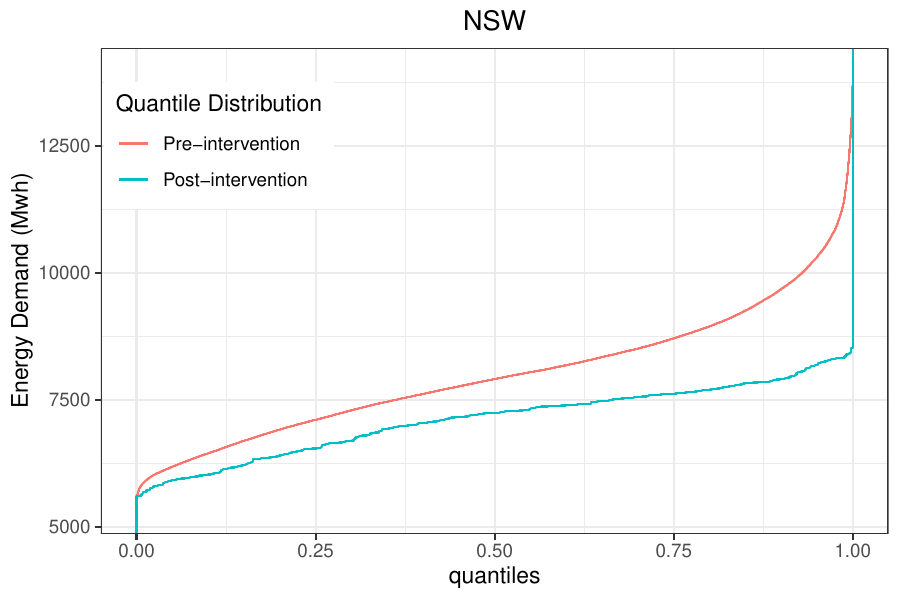}}
        {\includegraphics[width=0.35\textwidth,height = 4cm]{DeepAR_NSW.png}}
        {\includegraphics[width=0.35\textwidth,height = 4cm]{DeepAR_new_QD_NSW.png}}
        }
    \qquad
    \subfloat{
        {\includegraphics[width=0.35\textwidth,height = 4cm]{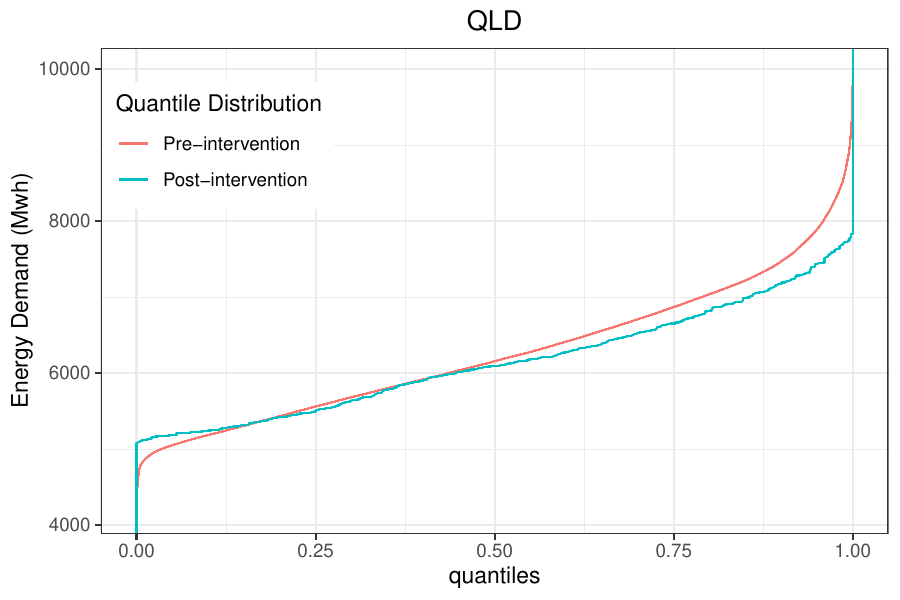}}
        {\includegraphics[width=0.35\textwidth,height = 4cm]{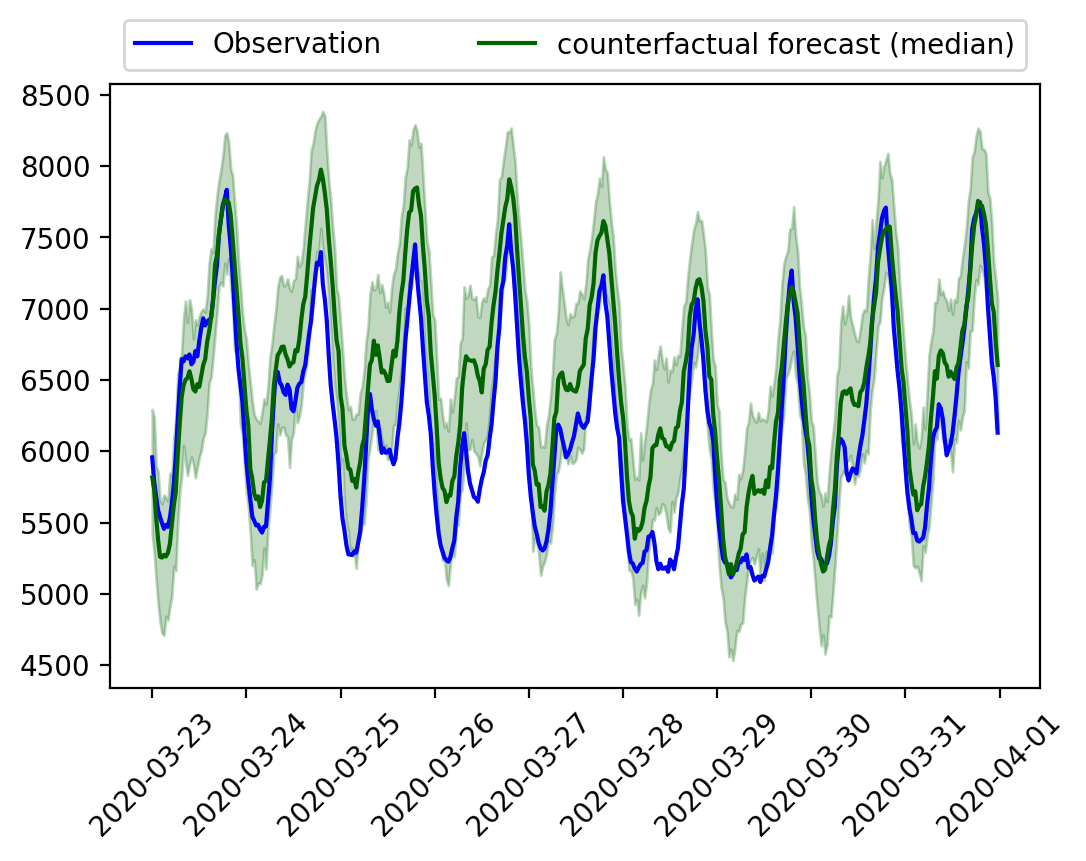}}
        {\includegraphics[width=0.35\textwidth,height = 4cm]{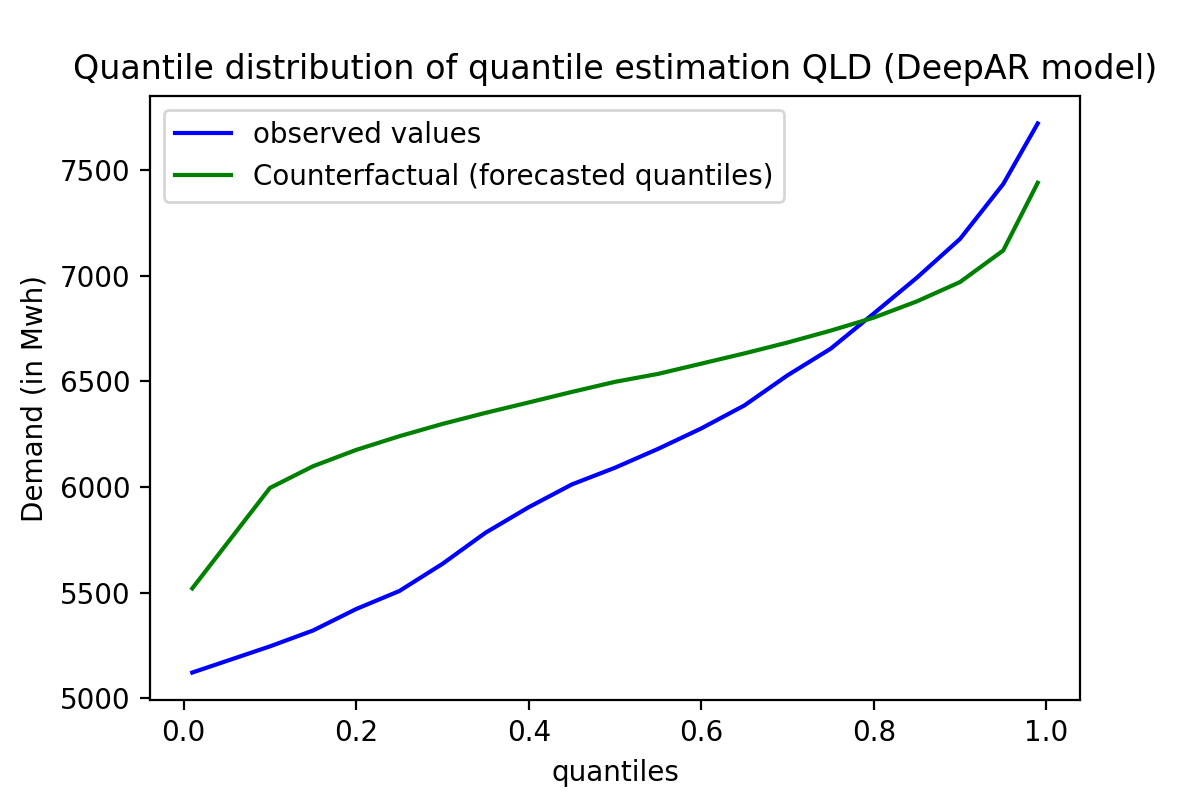}}
        }
    \qquad
    \subfloat{
        {\includegraphics[width=0.35\textwidth,height = 4cm]{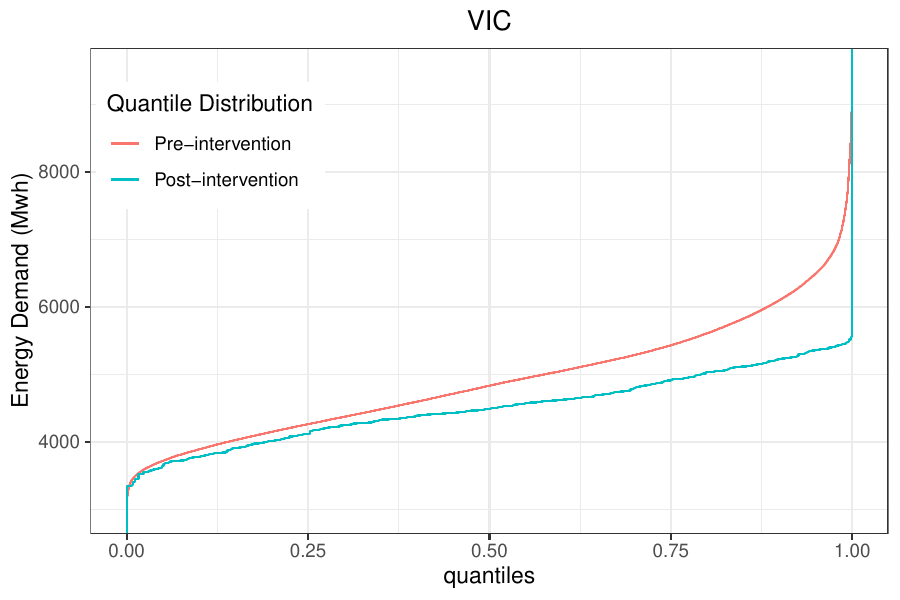}}
        {\includegraphics[width=0.35\textwidth,height = 4cm]{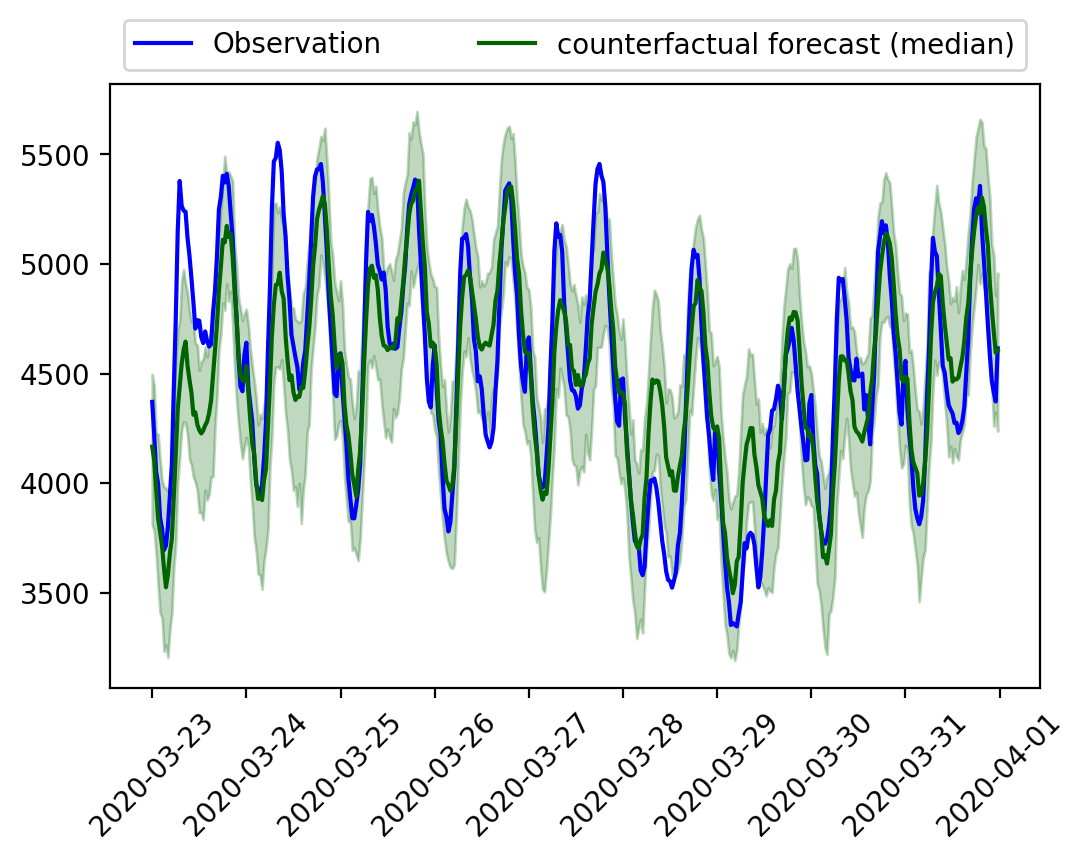}}
        {\includegraphics[width=0.35\textwidth,height = 4cm]{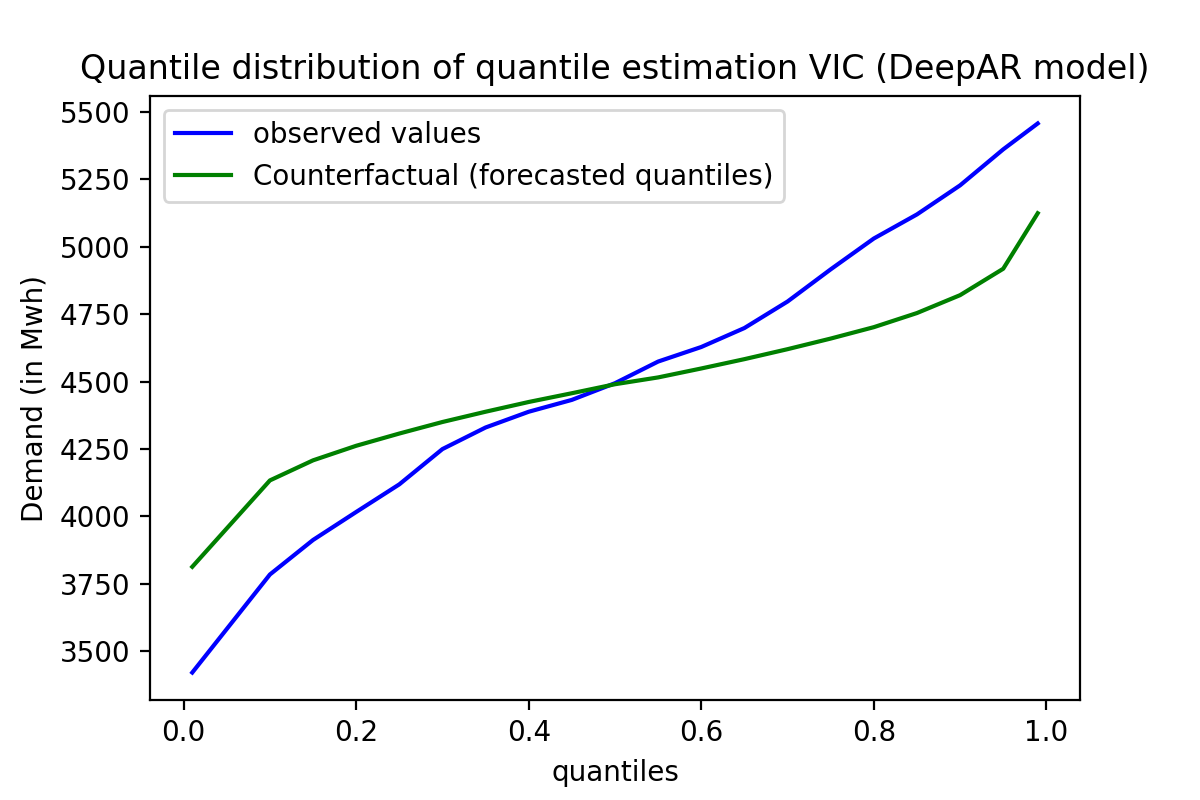}}
        }
    \qquad
    \subfloat{
        {\includegraphics[width=0.35\textwidth,height = 4cm]{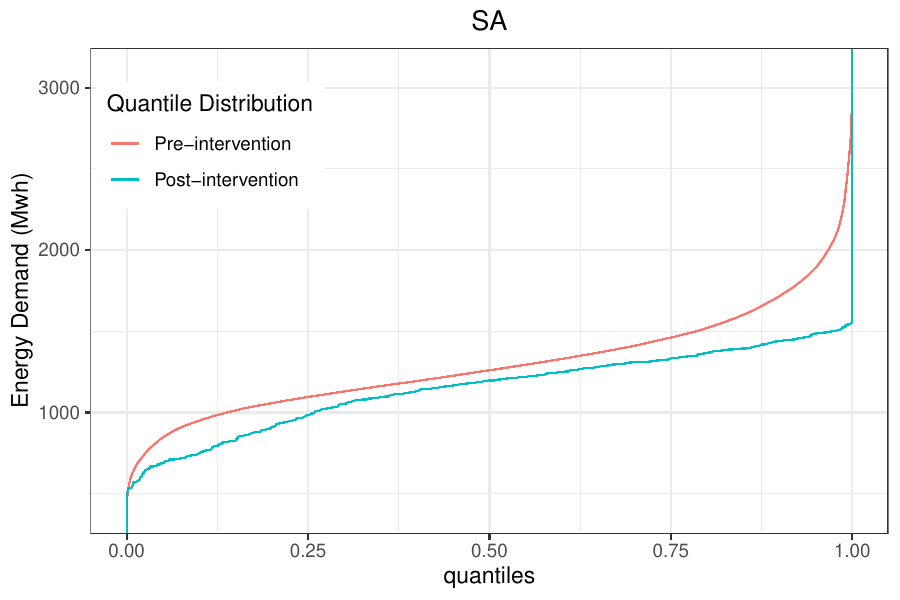}}
        {\includegraphics[width=0.35\textwidth,height = 4cm]{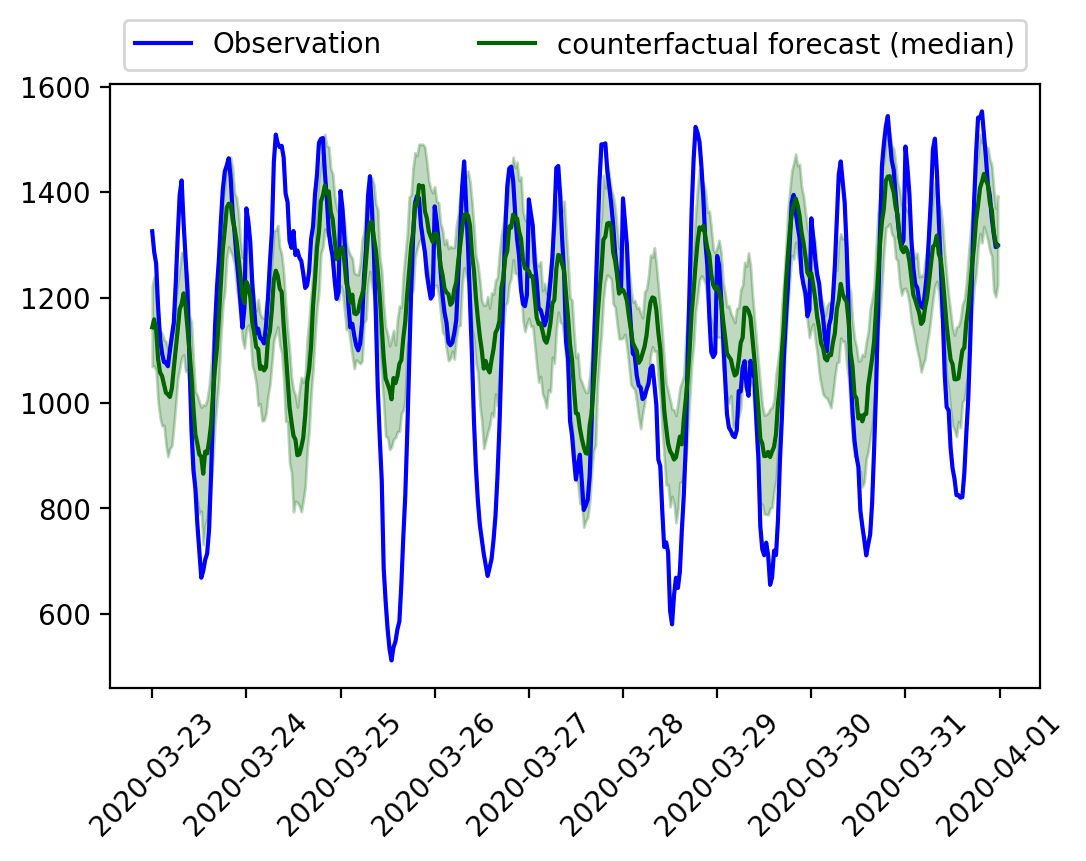}}
        {\includegraphics[width=0.35\textwidth,height = 4cm]{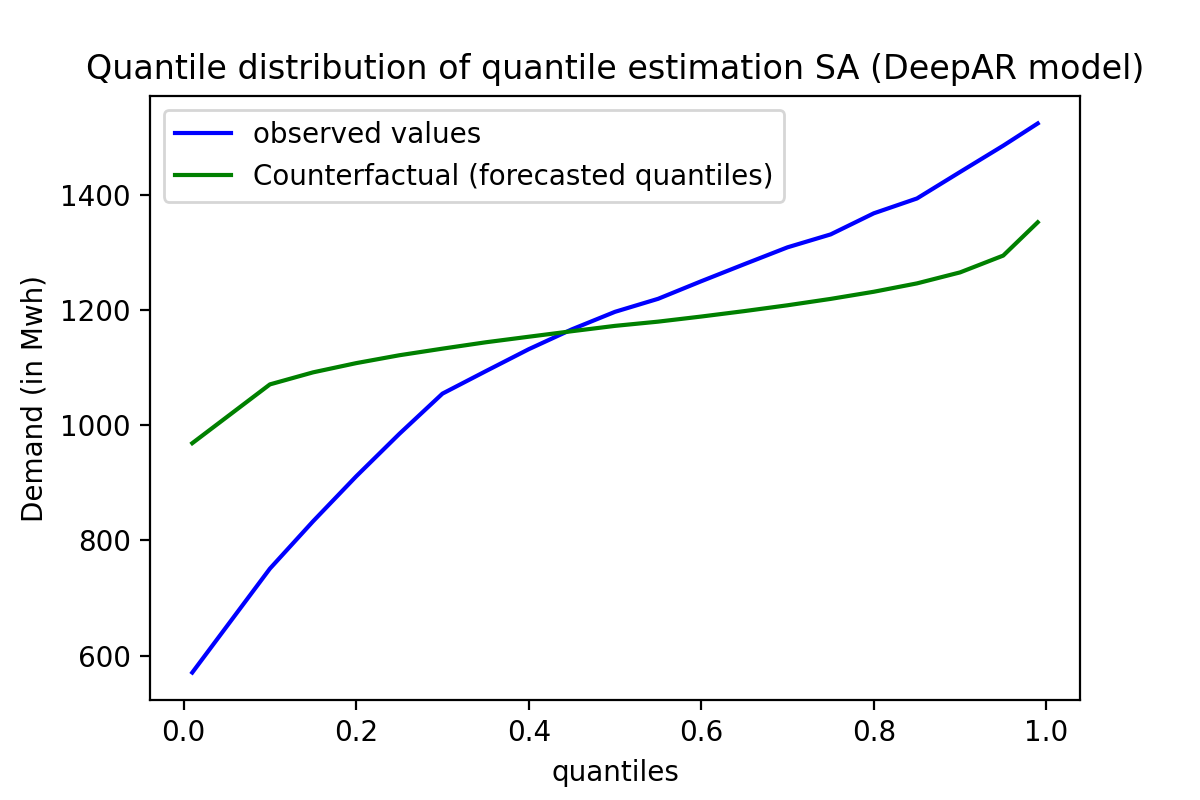}}
        }
    \qquad
    \subfloat{
        {\includegraphics[width=0.35\textwidth,height = 4cm]{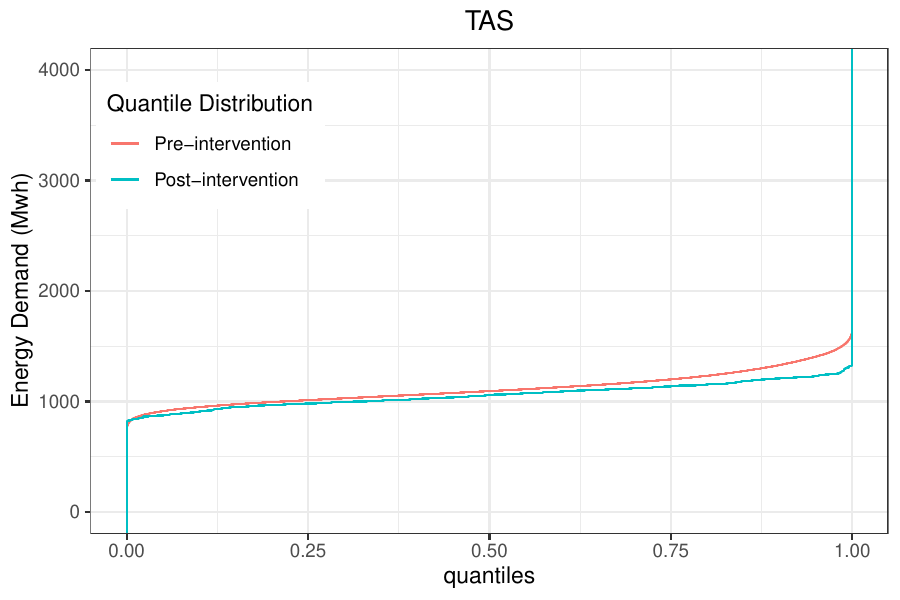}}
        {\includegraphics[width=0.35\textwidth,height = 4cm]{DeepAR_TAS.png}}
        {\includegraphics[width=0.35\textwidth,height = 4cm]{DeepAR_new_QD_TAS.png}}
        }

    \caption{Australian data -- Observed quantile distribution of pre- and post-intervention data (left), DeepAR model's point forecast (middle) and its forecast quantile distribution (right).\label{label2}}

\end{figure*}
\FloatBarrier

\FloatBarrier
\begin{figure}[ht!]
\centering
{\includegraphics[width=0.4\textwidth]{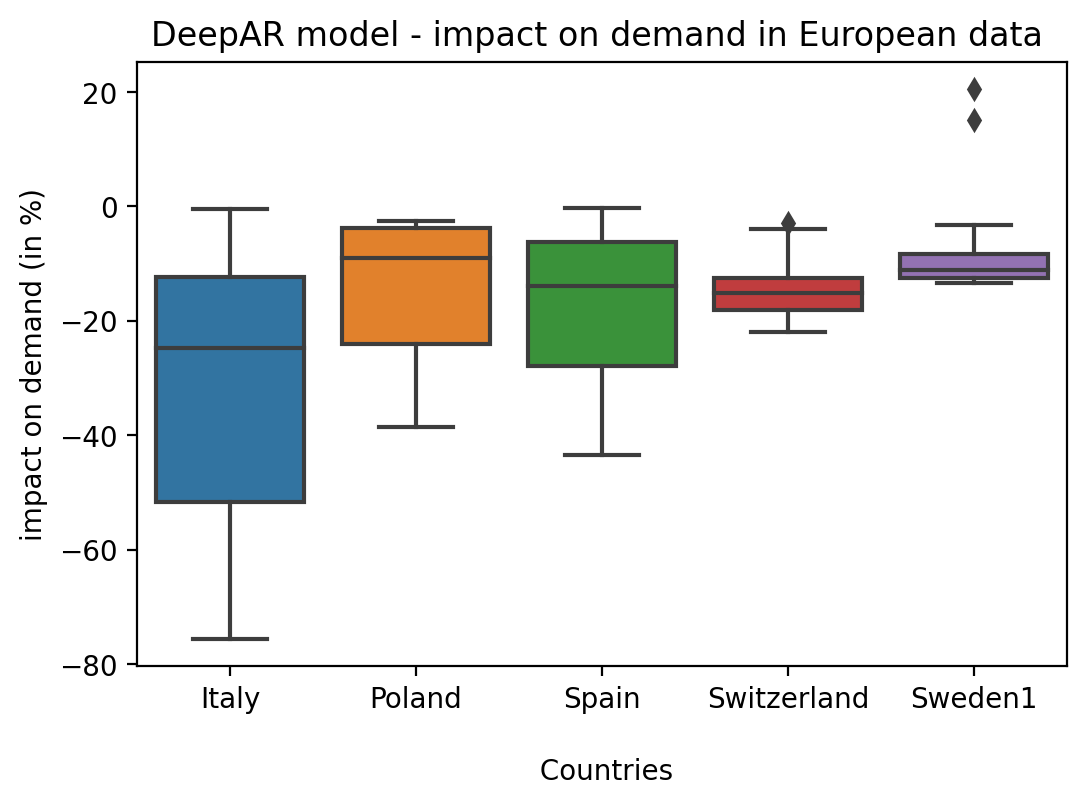}}
    \includegraphics[width=0.4\textwidth]{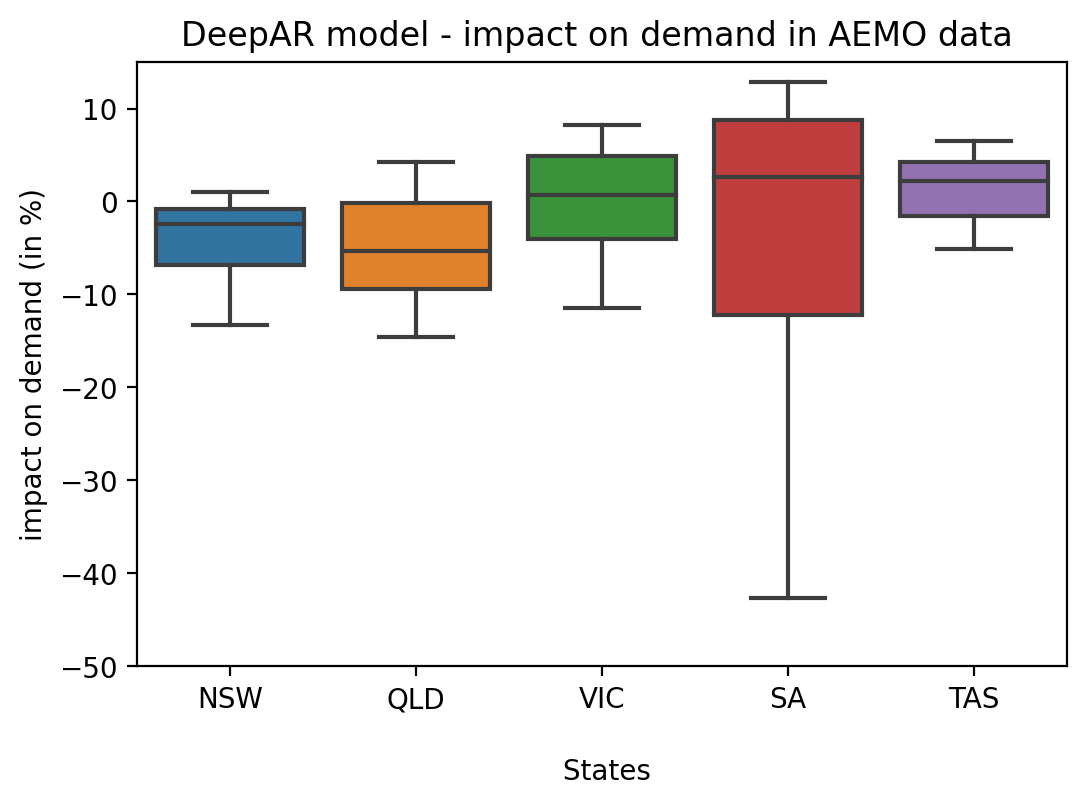}
\caption{ Boxplot representation of impact of Covid-19 on energy demand.}
\label{fig:fig12}
\end{figure}
\FloatBarrier

\bibliography{{IEEE_TPWRS_reviewed.bib}}{}
\bibliographystyle{IEEEtranN}

\section*{Biography}
\vspace{-36pt}
\begin{IEEEbiographynophoto}{Ankitha Nandipura Prasanna}
received the Bachelor of Engineering in Computer Science and Engineering from the Visvesvaraya Technological University, India, in 2014. She finished her Master degree in Artificial Intelligence at the Faculty of Information Technology, Monash University, Melbourne, Australia in 2022. Her research interests includes deep neural networks and time series forecasting.
\end{IEEEbiographynophoto}
\vspace{-36pt}
\begin{IEEEbiographynophoto}{Priscila Grecov}
is a PhD student in Computer Science at the Faculty of Information Technology, Monash University, Melbourne, Australia. Her research interests include Causal Inference, Deep Neural Networks and time series forecasting using Machine Learning methods. She finished her Master degree in Data Science at Monash university in 2020, and also holds an M.Sc. degree in Mathematics from the Institute of Pure and Applied Mathematics, Brazil. Her B.Sc. degree is in Economics obtained from the University of Brasilia, Brazil.
\end{IEEEbiographynophoto}
\vspace{-36pt}
\begin{IEEEbiographynophoto}{Angela Dieyu Weng}
is a Year 12 IB student at Lauriston Girls' School, Melbourne, Australia. Her research interests include applied and computational mathematics and machine learning. Angela has studied university-level mathematics courses at her leisure. She is planning to pursue a degree in the field of computational mathematics at university.
\end{IEEEbiographynophoto}
\vspace{-36pt}
\begin{IEEEbiographynophoto}{Christoph Bergmeir}
is a Senior Research Fellow in Data Science and Artificial Intelligence, and a 2019 ARC DECRA Fellow in the Department of Data Science and Artificial Intelligence at Monash University. His fellowship is on the development of “efficient and effective analytics for real-world time series forecasting”. Christoph holds a PhD in Computer Science from the University of Granada, Spain, and an M.Sc. degree in Computer Science from the University of Ulm, Germany.
\end{IEEEbiographynophoto}
\vfill